\documentclass[letterpaper]{article} 
\usepackage{aaai2027}\nocopyright

\usepackage[hyphens]{url}  
\usepackage{graphicx} 
\usepackage{natbib}  
\usepackage{caption} 
\usepackage{booktabs}
\usepackage{graphicx}
\usepackage[table]{xcolor}
\usepackage{array}
\usepackage{multirow}
\usepackage{placeins} 

\usepackage{amsmath}
\usepackage{amssymb}

\usepackage{newfloat}
\usepackage{listings}

\DeclareCaptionStyle{ruled}{labelfont=normalfont,labelsep=colon,strut=off} 

\definecolor{GroupRow}{HTML}{EFEFEF}
\definecolor{OursRow}{HTML}{F2F7FF}
\definecolor{KeepADGain}{HTML}{138A36}
\definecolor{KeepADDrop}{HTML}{C62828}

\newcommand{\method}{\textbf{KeepAD}}
\newcommand{\best}[1]{\textbf{#1}}
\newcommand{\second}[1]{\underline{#1}}
\newcommand{\bestpm}[2]{\textbf{#1} $\pm$ \textbf{#2}}
\newcommand{\secondpm}[2]{\underline{#1 $\pm$ #2}}

\newcommand{\gain}[1]{\textcolor{KeepADGain}{\scriptsize\ensuremath{\, (+#1)}}}
\newcommand{\drop}[1]{\textcolor{KeepADDrop}{\scriptsize\ensuremath{\, (-#1)}}}
\newcommand{\same}[1]{\textcolor{gray}{\scriptsize\ensuremath{\, (#1)}}}

\newcommand{\added}[1]{#1}
\newcommand{\hladd}[1]{#1}

\newcommand{\addedcap}[1]{#1}

\newenvironment{revision}{}{}
\newcommand{\codefix}[1]{#1}

\newcommand{\speedhead}{%
  \begin{tabular}[c]{@{}c@{}}
    \textbf{Speed}\\[-1pt]
    {\scriptsize img/s}
  \end{tabular}%
}

\title{Keep the Needle, Prune the Haystack: Defect-Preserving Token Pruning for Efficient Zero-Shot Anomaly Detection}

\author{
Yanning Hou\textsuperscript{\rm 2}\equalcontrib,
Jingyuan Zhang\textsuperscript{\rm 1}\equalcontrib,
Xiaoyun Wang\textsuperscript{\rm 1},
Qixiang Ma\textsuperscript{\rm 1},\\
Sihang Zhou\textsuperscript{\rm 2}\corresponding,
Ke Xu\textsuperscript{\rm 1}\corresponding
}
\affiliations{
\textsuperscript{\rm 1}Anhui University, Hefei, China\\
\textsuperscript{\rm 2}National University of Defense Technology, Changsha, China
}

\begin{document}

\maketitle

\begin{abstract}
Zero-shot visual anomaly detection has achieved remarkable progress, with recent vision-only approaches further improving performance while simplifying the inference pipeline. However, existing methods typically perform dense computation over all images and spatial tokens, despite the fact that normal samples dominate real-world scenarios and anomalies usually occupy only small regions. Token pruning offers a promising solution, but introduces an asymmetric pruning risk in anomaly detection: retaining normal tokens mainly incurs redundant computation, whereas removing anomalous tokens may eliminate the only evidence for detection and localization. This risk is particularly severe in early layers, where pruning provides the greatest computational benefit but anomaly semantics remain unreliable. We propose \method{}, a defect-preserving token pruning framework that formulates token selection as high-recall, anomaly-aware routing. In shallow layers, \method{} combines coverage-preserving selection over local $2\times2$ patch neighborhoods with deterministic anomaly rescue to reduce the risk of discarding subtle defects. In deeper layers, frozen normal and abnormal prototypes guide pruning under an image-adaptive token budget, aggressively removing low-risk normal tokens while preserving local anomaly evidence. Dense-to-sparse self-distillation further supervises early token routing without introducing additional inference overhead. Experiments on six industrial and seven medical zero-shot anomaly detection benchmarks show that \method{} reduces the token retention ratio to below $20\%$, while limiting the average degradation in image-level and pixel-level AUROC to within $2.7$ percentage points. At the most aggressive operating point, \method{} achieves a $7.9\times$ speedup over the strongest CLIP-based baseline.
\end{abstract}

\begin{links}
    \link{Code}{https://github.com/7HHHHH/KeepAD}
\end{links}

\section{Introduction}
\label{sec:introduction}

\begin{figure}[t]
\centering
\includegraphics[width=\linewidth]{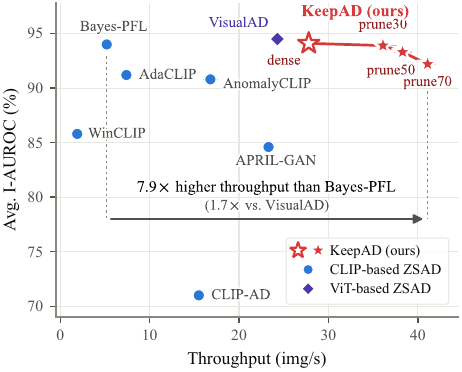}
\caption{\textbf{Accuracy--throughput trade-off.} At competitive accuracy, \method{} achieves up to $7.9\times$ and $1.7\times$ the throughput of Bayes-PFL and VisualAD, respectively; throughput is measured on a single NVIDIA RTX 5090 GPU.}
\label{fig:teaser}
\end{figure}

\begin{figure}[t]
\centering
\includegraphics[width=\linewidth]{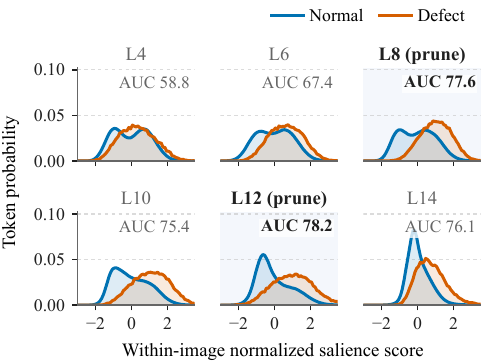}
\caption{\textbf{Depth-dependent token separability.} Results are obtained with CLIP ViT-L/14@336px. L8 and L12 balance token separability with remaining computation.}
\label{fig:intro_salience_depth}
\end{figure}

Zero-shot visual anomaly detection (ZSAD) seeks to recognize anomalous images and localize the responsible regions in categories unseen during training.
Recent methods increasingly exploit pretrained vision-language models and vision-only Vision Transformers (ViTs)~\cite{dosovitskiy2021vit,zhou2024anomalyclip,qu2025bayespfl,uniadnet,visualAD}, whose patch representations transfer well across object categories, textures, and anomaly types.
However, these models propagate every patch token through every Transformer layer, incurring substantial inference cost.

Efficient inference is particularly important for practical ZSAD systems.
Anomaly evidence is scarce at two levels: anomalous images are rare in deployment streams, and the decisive evidence within an anomalous image is usually confined to a few localized patch tokens.
Consequently, most computation is spent on normal images or normal regions that contribute little to the final prediction.
This dual sparsity makes token pruning a natural approach to accelerating ZSAD inference under practical workloads.

Pruning for anomaly detection, however, differs fundamentally from pruning for generic image classification.
Classification-oriented methods typically retain tokens that support dominant object semantics and discard less discriminative content with limited impact on the final prediction~\cite{rao2021dynamicvit,liang2022evit}.
In anomaly detection, a globally inconspicuous token may contain the only evidence required for image-level detection and pixel-level localization.
A pruning strategy centered on dominant semantics may therefore inadvertently remove precisely the subtle evidence that ZSAD must preserve.

Token pruning for anomaly detection consequently faces two core challenges.
\textbf{(i) Asymmetric pruning risk.}
Retaining redundant normal tokens mainly increases computation, whereas removing an anomaly-bearing token may irreversibly erase evidence required for detection and localization.
Token routing must therefore maintain high recall for potentially anomalous tokens and remove them only when they are confidently low risk.
\textbf{(ii) Early-pruning paradox.}
Earlier pruning saves more downstream computation, but shallow representations cannot yet reliably distinguish subtle defects from normal texture variations.
Deeper features provide more reliable anomaly cues, but leave progressively less computation to save.
Figure~\ref{fig:intro_salience_depth} illustrates this depth-dependent trade-off, indicating that the most computationally valuable stages are not necessarily the most reliable stages for anomaly-aware selection.

We address these challenges with \method{}, a progressive defect-preserving router that dynamically adapts both the token selection criterion and pruning strength as network depth increases and anomaly semantics mature.Our layer-wise statistical analysis of CLIP ViT-L/14@336px in Figure~\ref{fig:intro_salience_depth} reveals a depth-dependent trade-off between token separability and remaining computation.
Based on this observation, KeepAD adopts conservative spatial routing at L8 and transitions to more aggressive anomaly-aware pruning at L12.

At the early pruning stage, affinities to normal and anomaly prototypes are not yet sufficiently reliable.
KeepAD therefore combines prototype-free scoring with coverage-preserving selection on the original patch grid, retaining one representative from each local $2\times2$ block and allocating the residual budget to high-risk rescue candidates.
This design prevents global salience from concentrating the surviving sequence on dominant normal regions and reduces the risk of completely removing small or spatially isolated defects.

At the deeper pruning stage, token representations are more aligned with the anomaly detection objective.
KeepAD combines visual salience with affinities to frozen normal and anomaly prototypes, and assigns an image-adaptive token budget according to the distribution of anomaly evidence.
This enables more aggressive pruning after anomaly cues have become sufficiently informative.
Both stages preserve the original-grid identities of surviving tokens, maintaining the spatial correspondence required for dense anomaly localization.
Together, the two stages transition from conservative spatial preservation to anomaly-guided adaptive pruning.

Learning the shallow router remains challenging because tokens that appear uninformative at L8 may become anomaly-relevant in deeper layers, while hard selection obstructs direct gradient propagation.
KeepAD therefore introduces dense-to-sparse self-distillation.
During training, a frozen and unpruned dense branch provides detached deeper token-level anomaly priors as explicit preservation targets for the sparse branch.
A pruning-aware differentiable surrogate further allows the original image-level detection and pixel-level localization losses to optimize the both routers.
The teacher branch and differentiable surrogate are removed at inference, leaving a single hard-pruned forward pass.

For dense localization, KeepAD records the original-grid coordinates of surviving tokens and restores the patch-level anomaly map by assigning each dropped position to its nearest survivor. This recovery restores spatial resolution but cannot recover anomaly evidence removed during pruning, whose preservation instead depends on the risk-aware routing strategy. Image-level anomaly scores are aggregated only from physical survivors, thereby preventing recovered positions from introducing artificial anomaly evidence into image-level prediction during final inference.

Our contributions are threefold: (1) we formulate token pruning for ZSAD as a risk-aware routing problem and identify its asymmetric pruning risk and early-pruning paradox; (2) we propose KeepAD, a depth-aware progressive router that transitions from coverage-preserving shallow routing to prototype-guided adaptive pruning, with dense-to-sparse self-distillation supervising shallow token preservation; and (3) extensive experiments on 13 industrial and medical benchmarks demonstrate that KeepAD retains fewer than $20\%$ of tokens while maintaining competitive detection and localization accuracy, and achieves a $7.9\times$ speedup over the strongest CLIP-based baseline.

\section{Related Work}\label{sec:related work}

\subsection{Industrial Anomaly Detection}
\label{sec:rw_iad}

\hladd{
Classical unsupervised anomaly detection builds category-specific models from memory banks, feature statistics, or distillation~\cite{roth2022patchcore,defard2021padim,deng2022rd4ad}, whereas zero-shot anomaly detection (ZSAD) targets unseen categories.
CLIP-based ZSAD has progressed from WinCLIP's handcrafted multi-scale prompts~\cite{jeong2023winclip} to learned projections or dual paths in APRIL-GAN and CLIP-AD~\cite{chen2023aprilgan,chen2023clipad}, object-agnostic prompts and modified visual attention in AnomalyCLIP~\cite{zhou2024anomalyclip}, and hybrid or probabilistic prompting in AdaCLIP and Bayes-PFL~\cite{cao2024adaclip,qu2025bayespfl}.
Language-free methods such as UniADet and VisualAD remove the text encoder but retain task-specific weights or additional state tokens, attention, and alignment modules~\cite{uniadnet,visualAD}.
Despite these differences, all process the full patch grid through the ViT.
KeepAD instead reduces backbone cost by pruning within the vision encoder while protecting sparse defect evidence during inference.
}

\subsection{Token Pruning}
\label{sec:rw_token_pruning}

\hladd{
ViT token pruning accelerates inference by removing low-utility tokens using learned importance, structured schedules, attention graphs, or hardware-aware objectives~\cite{rao2021dynamicvit,liang2022evit,tang2022patchslimming,kong2022spvit,wang2024zerotprune}.
For dense prediction, SparseViT prunes window activations, DToP exits easy tokens early, and SViT reactivates previously removed tokens when needed~\cite{chen2023sparsevit,tang2023dtop,liu2024revisitingpruning}.
VLM pruning methods retain visual tokens using attention, redundancy, diversity, multiple cues, or adaptive budgets~\cite{chen2024fastv,wen2025dart,alvar2025divprune,huang2025dynamicllava,ye2025atpllava,takezoe2026learnpruner,fang2026prunesid}.
These methods target classification, known-category dense prediction, or generation rather than the complete-miss risk of rare defects.
VMAD is anomaly-specific, but its efficiency module operates only in the MLLM projector after dense visual encoding~\cite{deng2026vmad}.
KeepAD instead performs progressive token pruning inside the ViT, combines anomaly evidence with local coverage, and tracks physical survivors for dense localization.
}


\section{Method}
\label{sec:method}

\begin{figure*}[t]
    \centering
    \includegraphics[width=0.92\linewidth]{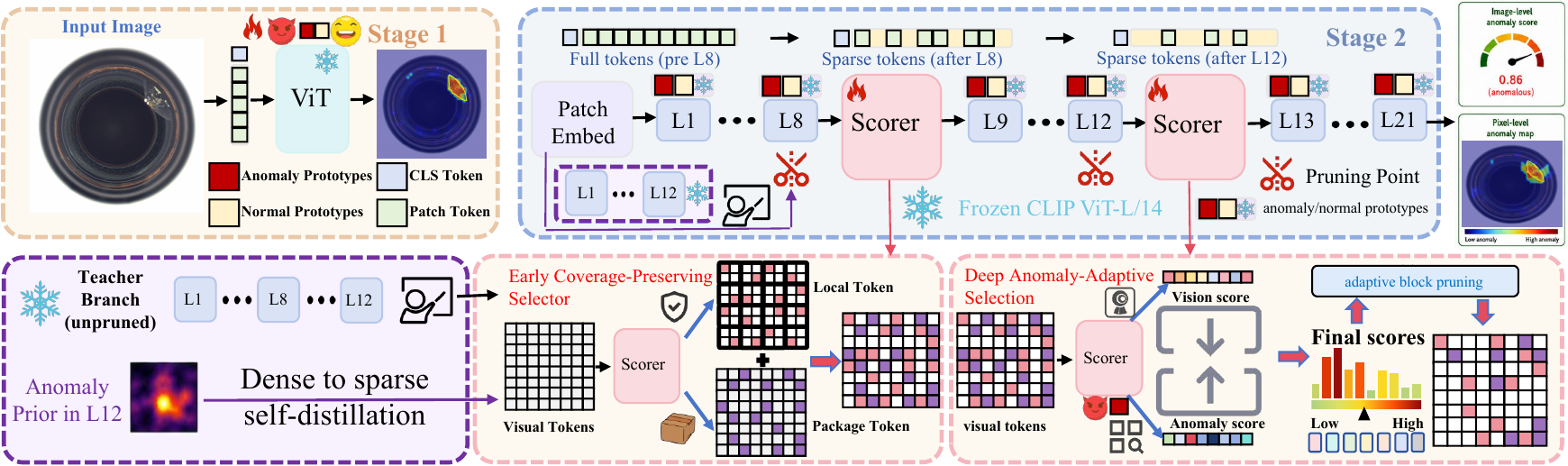}
    \caption{\textbf{Overview of KeepAD.}
    Stage~1 trains a dense ViT anomaly detector, which is then frozen.
    Stage~2 learns an L8 coverage-preserving selector and an L12 prototype-guided selector with an image-adaptive budget.
    The dense teacher and differentiable masks are used only for selector training.
    At inference, hard-selected tokens are propagated, their original-grid coordinates support dense recovery, and image-level scoring uses physical survivors only.}
    \label{fig:framework}
\end{figure*}

\subsection{Overview}
\label{sec:method_overview}
\label{sec:sparse_state}

KeepAD follows a two-stage design (Fig.~\ref{fig:framework}). Stage~1 trains a dense ViT anomaly detector $F$; Stage~2 freezes $F$ and learns two lightweight token selectors at L8 and L12, enabling sparse execution without retraining the detector. For CLIP ViT-L/14@336px, these pruning points are chosen to balance token separability with remaining computation, rather than to maximize separability in isolation.

\paragraph{Stage~1: dense detector.}
We keep the ViT backbone frozen and train only a lightweight anomaly head on top of a set of intermediate layers. For each such layer $l$, the head holds a pair of learnable normal/anomaly prototypes $(p_n^l,p_a^l)$; the anomaly response of a patch is the softmax probability of the anomaly prototype under cosine similarity, and the image-level score aggregates patch responses across layers. The head is optimized with a focal--Dice segmentation loss on the anomaly map and a cross-entropy loss on the image label, using synthetic anomalies produced by a collage-and-crop augmentation to expose the model to localized defects. This yields patch-level responses and an image-level score, and---crucially for Stage~2---a frozen normal/anomaly prototype pair at every head layer that later serves as stable, layer-specific prior knowledge for anomaly-aware pruning.

\paragraph{Stage~2: token selectors.}
We freeze the Stage~1 backbone \emph{and} head, and insert two selectors at layers~8 and~12 (L8, L12). They play complementary roles matched to depth: the shallow selector preserves local evidence before reliable anomaly semantics emerge, whereas the deep selector reuses the frozen L12 prototypes to concentrate computation on high-risk regions. Only the two lightweight selectors are trained; $F$ is never updated.

\paragraph{Sparse state on the original grid.}
Each live patch token $x_i^l$ carries its coordinate $\pi_i^l$ on the original $S\times S$ patch grid ($S=37$, $N_0=1369$ for our backbone). Whenever tokens are gathered, their coordinates are gathered by the same indices, so local selection and dense recovery are always defined on the original grid even after the sequence becomes sparse during subsequent processing. During training, hard decisions are replaced by differentiable masks and an unpruned branch supplies dense supervision; both are removed at inference, which uses a single hard-pruned forward pass.

\subsection{Progressive Defect-Preserving Pruning}
\label{sec:progressive_pruning}

\paragraph{Early coverage-preserving selection.}
\label{sec:early_pruning}
Pruning at L8 provides the largest computational saving, but shallow tokens do not yet support reliable anomaly classification. We therefore score them without anomaly prototypes. For $l\in\{8,12\}$, the prototype-free visual score is
\begin{equation}
\bar{x}^{l}=\frac{1}{N_l}\sum_i x_i^{l},
\qquad
v_i^{l}=\frac{(W_q^{l}x_i^{l})^\top(W_k^{l}\bar{x}^{l})}{\sqrt{h}}.
\label{eq:visual_score}
\end{equation}
Here $\bar{x}^{l}$ is the mean of the $N_l$ live tokens at layer $l$ ($N_0$ the original count), and $W_q^{l}$ and $W_k^{l}$ are the selector's query and key projections of width $h$. We partition the original grid into non-overlapping $2\times2$ blocks and let $\mathcal{B}_m^8$ contain the live tokens whose original coordinates fall in block $m$. To prevent global top-$k$ selection from concentrating all tokens in a few salient normal regions, we first retain one representative from every non-empty block,
\begin{equation}
\mathcal{K}_{\mathrm{cov}}^8
=
\bigcup_{m:\,\mathcal{B}_m^8\neq\emptyset}
\left\{\arg\max_{i\in\mathcal{B}_m^8}v_i^8\right\}.
\label{eq:l8_coverage}
\end{equation}
The L8 target is $K_8^{\mathrm{tar}}=\max\bigl(|\mathcal{K}_{\mathrm{cov}}^8|,\operatorname{round}(\rho_8N_0)\bigr)$, where $\rho_8$ is the L8 keep fraction set by the chosen operating point. Any remaining budget $R_8=K_8^{\mathrm{tar}}-|\mathcal{K}_{\mathrm{cov}}^8|$ is filled deterministically from the unselected tokens: \codefix{up to 16 slots} are assigned to \codefix{cell-best candidates} from \codefix{distinct cells} of an \codefix{$8\times8$ original-grid partition (64 cells)}, and \codefix{all remaining slots are filled by global score}, with no random sampling. Thus the coverage set provides a local safety floor---one representative per non-empty block---while the rescue set spends the residual budget on additional high-utility or spatially isolated evidence that a pure top-$k$ rule would discard.

\paragraph{Deep anomaly-adaptive selection.}
\label{sec:deep_pruning}
At L12, token features carry stronger task semantics, so pruning can be guided by anomaly prototypes. In addition to the visual score $v_i^{12}$ from Eq.~\eqref{eq:visual_score}, we compare each token with the frozen normal and anomaly prototypes $p_n$ and $p_a$ of the Stage-1 head. Let $q_i=W_q^c x_i^{12}$ and $k_{n/a}=W_k^c p_{n/a}$, and let $h_c$ be the width of this projection. We compute
\begin{equation}
\begin{aligned}
\alpha_i^{12}
&=\sigma\!\left(
\frac{\langle q_i,k_a\rangle-\langle q_i,k_n\rangle}{\sqrt{h_c}}
\right),\\
r_i^{12}
&=v_i^{12}+\tanh(\gamma)\alpha_i^{12}.
\end{aligned}
\label{eq:l12_risk}
\end{equation}
where $\sigma(\cdot)$ is the logistic sigmoid and the learnable gate $\gamma$ is initialized to zero. This initialization lets the selector begin from the stable visual branch and incorporate prototype evidence only when it becomes useful.

A single fixed budget is unnecessarily conservative for easy images and overly aggressive for evidence-rich images. We standardize the fused risks as
$u_i=\sigma((r_i^{12}-\mu_r)/(\sigma_r+\epsilon))$, where $\mu_r$ and $\sigma_r$ are the mean and standard deviation of $r_i^{12}$ over the live tokens, and summarize their distribution by
\begin{equation}
\begin{aligned}
E(I)
&=\tfrac{1}{2}E_{\mathrm{var}}(u)+\tfrac{1}{2}E_{\mathrm{tail}}(u),\\
\widetilde{\rho}(I)
&=\rho_{\min}+(\rho_{\max}-\rho_{\min})
  \sigma\!\left(\frac{E(I)-c}{T_b}\right),\\
K_{12}^{\mathrm{tar}}(I)
&=\operatorname{round}\!\left(N_{12}\widetilde{\rho}(I)^{\kappa}\right).
\end{aligned}
\label{eq:adaptive_budget}
\end{equation}
Here $E_{\mathrm{var}}$ is a bounded dispersion statistic of the standardized risks over all live tokens and $E_{\mathrm{tail}}$ is the mean of their top-$3\%$ upper tail; both are computed per image. A larger $E(I)$---sharper or heavier-tailed evidence---therefore assigns more computation to images containing pronounced local anomalies, while evidence-poor images fall back toward the safe minimum budget. Note that the realized ratio is bounded by $\rho_{\min}^{\kappa}$ and $\rho_{\max}^{\kappa}$ before block rounding, rather than directly by $\rho_{\min}$ and $\rho_{\max}$.

To retain local structure, L12 selects whole live blocks rather than isolated tokens. Each non-empty block receives the mean fused risk
$b_m=|\mathcal{B}_m^{12}|^{-1}\sum_{i\in\mathcal{B}_m^{12}}r_i^{12}$; blocks are ranked by $b_m$ and added until $K_{12}^{\mathrm{tar}}(I)$ is reached. This produces the final keep set $\mathcal{K}_{12}$ while avoiding highly fragmented spatial support.

\subsection{Pruning-Aware Learning}
\label{sec:training}

\begin{revision}
\paragraph{Differentiable selection.}
Hard keep decisions would block gradients into the selectors. During Stage~2, each selector therefore produces both a hard keep set {$\mathcal K_l$}, which gathers a physically shortened sequence exactly as at inference, and relaxed keep probabilities {$\widetilde m^l$}, which carry gradients through the surviving tokens. The relaxed masks are accumulated along the sparse path and reweight subsequent transformer blocks. The L8 relaxation follows the coverage-plus-rescue structure, whereas the L12 relaxation operates on block scores. This hard-topology/soft-gradient construction is distinct from a classical straight-through identity; because the detector is frozen, it updates only the two selectors.

\paragraph{Training objective.}
Stage~2 is trained to preserve the detector's task supervision while controlling sparse computation and protecting anomaly evidence:
\begin{equation}
\mathcal L_{\mathrm{S2}}
=
\mathcal L_{\mathrm{det}}
+\mathcal R_{\mathrm{eff}}
+\mathcal R_{\mathrm{pres}}.
\label{eq:loss}
\end{equation}
Here {$\mathcal L_{\mathrm{det}}$} is the detector's original image- and patch-level anomaly objective evaluated on the sparse path. {$\mathcal R_{\mathrm{eff}}$} regularizes expected sparse computation, while {$\mathcal R_{\mathrm{pres}}$} combines high-risk-token protection, supervision from available source-domain defect masks, and dense-to-sparse teacher alignment. Appendix~\ref{app:impl_details} provides their exact decomposition, weights, and relaxation settings.

\paragraph{Dense-to-sparse self-distillation.}
The detection objective reaches L8 only through many frozen layers and is therefore indirect about \emph{which shallow tokens will matter later}. We let the frozen detector provide this future evidence: an additional unpruned pass produces detached L12 anomaly responses on the original grid, and the relaxed L8 keep distribution is aligned with this deeper evidence within each local block. The resulting teacher signal explicitly guides early token retention tells the shallow selector which locations are likely to become anomaly-relevant before reliable semantics have emerged. This dense-to-sparse teacher exists only during Stage~2 training and adds no inference cost; its exact normalization and divergence are specified in Appendix~\ref{app:impl_details}.
\end{revision}

\subsection{Dense Recovery and Inference}
\label{sec:recovery}

\begin{table*}[t]
\centering
\setlength{\tabcolsep}{2.1pt}
\renewcommand{\arraystretch}{1.18}
\resizebox{\linewidth}{!}{
\begin{tabular}{@{}llcccccccc@{}}
\toprule
\textbf{Method}
& \textbf{Source}
& \begin{tabular}[c]{@{}c@{}}\textbf{MVTec-AD}\\[-1pt]{\scriptsize I-AUROC / P-AUROC}\end{tabular}
& \begin{tabular}[c]{@{}c@{}}\textbf{VisA}\\[-1pt]{\scriptsize I-AUROC / P-AUROC}\end{tabular}
& \begin{tabular}[c]{@{}c@{}}\textbf{BTAD}\\[-1pt]{\scriptsize I-AUROC / P-AUROC}\end{tabular}
& \begin{tabular}[c]{@{}c@{}}\textbf{KSDD2}\\[-1pt]{\scriptsize I-AUROC / P-AUROC}\end{tabular}
& \begin{tabular}[c]{@{}c@{}}\textbf{DAGM}\\[-1pt]{\scriptsize I-AUROC / P-AUROC}\end{tabular}
& \begin{tabular}[c]{@{}c@{}}\textbf{DTD-Synthetic}\\[-1pt]{\scriptsize I-AUROC / P-AUROC}\end{tabular}
& \begin{tabular}[c]{@{}c@{}}\textbf{Avg.}\\[-1pt]{\scriptsize I-AUROC / P-AUROC}\end{tabular}
& \begin{tabular}[c]{@{}c@{}}\textbf{Speed}\\[-1pt]{\scriptsize img/s}\end{tabular} \\
\midrule

\rowcolor{GroupRow}
\multicolumn{10}{@{}c@{}}{\textbf{CLIP-based methods}} \\
WinCLIP
& CVPR'23
& (90.4, 82.3)
& (75.6, 73.2)
& (68.2, 72.7)
& (93.5, 94.1)
& (91.8, 87.6)
& (95.1, 79.5)
& (85.8, 81.6)
& 1.9 $\pm$ 0.4 \\

APRIL-GAN
& CVPRW'23
& (86.1, 87.5)
& (77.4, 93.8)
& (73.7, 91.3)
& (90.4, 94.5)
& (94.4, 84.4)
& (85.5, 94.9)
& (84.6, 91.1)
& 23.3 $\pm$ 0.5 \\

CLIP-AD
& arXiv'23
& (74.1, 77.9)
& (66.2, 93.0)
& (66.7, 80.9)
& (81.7, 95.6)
& (62.1, 69.1)
& (75.1, 86.6)
& (71.0, 83.9)
& 15.5 $\pm$ 0.2 \\

AnomalyCLIP
& ICLR'24
& (91.6, 91.0)
& (81.0, 95.4)
& (88.7, 93.0)
& (91.9, 98.0)
& (98.0, \best{96.9})
& (93.7, 97.5)
& (90.8, \second{95.3})
& 16.8 $\pm$ 0.8 \\

AdaCLIP
& ECCV'24
& (92.0, 88.5)
& (79.7, 95.1)
& (90.0, 87.7)
& (94.9, 96.1)
& (98.3, 88.6)
& (92.1, 95.1)
& (91.2, 91.9)
& 7.4 $\pm$ 0.7 \\

Bayes-PFL
& CVPR'25
& (92.3, \best{91.9})
& (86.2, \best{95.9})
& (94.1, \best{95.7})
& (\second{97.8}, \best{99.5})
& (\second{98.7}, \second{96.2})
& (94.8, \second{98.3})
& (94.0, \best{96.3})
& 5.2 $\pm$ 0.3 \\

\midrule
\rowcolor{GroupRow}
\multicolumn{10}{@{}c@{}}{\textbf{ViT-based methods}} \\
UniADet
& arXiv'26
& (92.4, \second{91.8})
& (88.0, \second{95.8})
& (\best{96.4}, \best{95.7})
& --
& --
& (\best{98.1}, \best{98.4})
& --
& -- \\

VisualAD
& CVPR'26
& (92.2, 90.8)
& (84.7, \second{95.8})
& (\second{94.9}, 91.1)
& (\best{98.0}, 98.5)
& (\best{99.5}, 92.2)
& (\second{97.5}, 98.1)
& (\best{94.5}, 94.4)
& 24.3 $\pm$ 0.5 \\

\rowcolor{OursRow}
\method{} w/o pruning
& --
& (\second{92.7}, 91.2)
& (\best{88.8}, \second{95.8})
& (91.9, \second{94.2})
& (\second{97.8}, \second{99.3})
& (98.0, 93.5)
& (95.1, 96.9)
& (\second{94.1}, 95.2)
& 27.8 $\pm$ 0.7 \\

\rowcolor{OursRow}
\method{}-L8 prune30
& --
& (\best{93.1}\gain{0.4}, 89.6\drop{1.6})
& (\second{88.5}\drop{0.3}, 95.1\drop{0.7})
& (89.9\drop{2.0}, 91.9\drop{2.3})
& (97.4\drop{0.4}, 99.0\drop{0.3})
& (97.8\drop{0.2}, 93.0\drop{0.5})
& (96.5\gain{1.4}, 95.7\drop{1.2})
& (93.9\drop{0.2}, 94.1\drop{1.1})
& 36.1 $\pm$ 1.6\gain{8.3} \\

\rowcolor{OursRow}
\method{}-L8 prune50
& --
& (92.0\drop{0.7}, 89.1\drop{2.1})
& (87.7\drop{1.1}, 94.6\drop{1.2})
& (90.1\drop{1.8}, 91.4\drop{2.8})
& (97.3\drop{0.5}, 98.7\drop{0.6})
& (96.8\drop{1.2}, 92.9\drop{0.6})
& (96.0\gain{0.9}, 95.2\drop{1.7})
& (93.3\drop{0.8}, 93.7\drop{1.5})
& \secondpm{38.3}{1.0}\gain{10.5} \\

\rowcolor{OursRow}
\method{}-L8 prune70
& --
& (90.5\drop{2.2}, 88.6\drop{2.6})
& (86.1\drop{2.7}, 94.3\drop{1.5})
& (89.7\drop{2.2}, 90.7\drop{3.5})
& (97.3\drop{0.5}, 98.4\drop{0.9})
& (94.9\drop{3.1}, 92.3\drop{1.2})
& (94.8\drop{0.3}, 93.8\drop{3.1})
& (92.2\drop{1.9}, 93.0\drop{2.2})
& \bestpm{41.1}{0.9}\gain{13.3} \\

\bottomrule
\end{tabular}
}
\caption{Industrial ZSAD results. Dataset columns report I-AUROC/P-AUROC; Avg. is the mean over six benchmarks. Best and second-best results are \textbf{bold} and \underline{underlined}; green/red values show absolute changes from \method{} w/o pruning.}
\label{tab:industrial_benchmarks}
\end{table*}
\begin{table*}[t]
\centering
\setlength{\tabcolsep}{1.8pt}
\renewcommand{\arraystretch}{1.18}
\resizebox{\linewidth}{!}{
\begin{tabular}{@{}llccccccccc@{}}
\toprule
\textbf{Method}
& \textbf{Source}
& \begin{tabular}[c]{@{}c@{}}\textbf{OCT17}\\[-1pt]{\scriptsize I-AUROC}\end{tabular}
& \begin{tabular}[c]{@{}c@{}}\added{\textbf{BrainMRI}}\\[-1pt]{\scriptsize I-AUROC}\end{tabular}
& \begin{tabular}[c]{@{}c@{}}\textbf{Brain AD}\\[-1pt]{\scriptsize I-AUROC / P-AUROC}\end{tabular}
& \begin{tabular}[c]{@{}c@{}}\textbf{HIS}\\[-1pt]{\scriptsize I-AUROC}\end{tabular}
& \begin{tabular}[c]{@{}c@{}}\textbf{CVC-ClinicDB}\\[-1pt]{\scriptsize P-AUROC}\end{tabular}
& \begin{tabular}[c]{@{}c@{}}\textbf{Endo}\\[-1pt]{\scriptsize P-AUROC}\end{tabular}
& \begin{tabular}[c]{@{}c@{}}\textbf{Kvasir}\\[-1pt]{\scriptsize P-AUROC}\end{tabular}
& \begin{tabular}[c]{@{}c@{}}\textbf{Avg.}\\[-1pt]{\scriptsize I-AUROC / P-AUROC}\end{tabular}
& \begin{tabular}[c]{@{}c@{}}\textbf{Speed}\\[-1pt]{\scriptsize img/s}\end{tabular} \\
\midrule

\rowcolor{GroupRow}
\multicolumn{11}{@{}c@{}}{\textbf{CLIP-based methods}} \\
WinCLIP
& CVPR'23
& 55.2
& 86.6
& (72.5, 87.6)
& 47.1
& 70.3
& 68.2
& 69.7
& (65.4, 74.0)
& 1.8 $\pm$ 0.2 \\

APRIL-GAN
& CVPRW'23
& 30.4
& 89.3
& (58.8, 83.6)
& 59.8
& 82.4
& 82.7
& 77.6
& (59.6, 81.6)
& 23.5 $\pm$ 0.3 \\

CLIP-AD
& arXiv'23
& 58.1
& 83.0
& (72.1, 94.1)
& 44.7
& 76.5
& 78.1
& 73.6
& (64.5, 80.6)
& 15.3 $\pm$ 0.1 \\

AnomalyCLIP
& ICLR'24
& 63.7
& 96.4
& (69.0, 95.1)
& 55.2
& 84.6
& 86.5
& 82.0
& (71.1, \best{87.1})
& 16.7 $\pm$ 0.4 \\

AdaCLIP
& ECCV'24
& 77.3
& 94.9
& (80.0, 95.2)
& 59.9
& 84.3
& 82.9
& 80.3
& (78.0, 85.7)
& 7.2 $\pm$ 0.2 \\

Bayes-PFL
& CVPR'25
& 67.7
& \best{98.0}
& (82.7, 95.8)
& \best{61.4}
& 79.3
& \second{86.9}
& \second{83.3}
& (77.5, 86.3)
& 5.5 $\pm$ 0.2 \\

\midrule
\rowcolor{GroupRow}
\multicolumn{11}{@{}c@{}}{\textbf{ViT-based methods}} \\
UniADet
& arXiv'26
& --
& 95.7
& --
& --
& \best{90.1}
& \best{90.7}
& \best{87.5}
& --
& -- \\

VisualAD
& CVPR'26
& \best{88.9}
& \second{96.7}
& (80.8, 95.2)
& \second{60.1}
& \second{85.2}
& 84.9
& 80.3
& (\best{81.6}, \second{86.4})
& 24.7 $\pm$ 0.6 \\

\rowcolor{OursRow}
\method{} w/o pruning
& --
& 80.8
& \second{96.7}
& (83.2, \best{96.9})
& 56.0
& 78.7
& 81.4
& 80.3
& (79.2, 84.3)
& 27.1 $\pm$ 0.4 \\

\rowcolor{OursRow}
\method{}-L8 prune30
& --
& 83.7\gain{2.9}
& 93.6\drop{3.1}
& (\best{87.1}\gain{3.9}, \second{96.4}\drop{0.5})
& 58.7\gain{2.7}
& 78.9\gain{0.2}
& 77.0\drop{4.4}
& 77.7\drop{2.6}
& (\second{80.8}\gain{1.6}, 82.5\drop{1.8})
& 34.9 $\pm$ 0.6\gain{7.8} \\

\rowcolor{OursRow}
\method{}-L8 prune50
& --
& \second{84.1}\gain{3.3}
& 92.0\drop{4.7}
& (84.6\gain{1.4}, 95.6\drop{1.3})
& 58.3\gain{2.3}
& 79.5\gain{0.8}
& 76.7\drop{4.7}
& 77.0\drop{3.3}
& (79.8\gain{0.6}, 82.2\drop{2.1})
& \secondpm{36.9}{0.5}\gain{9.8} \\

\rowcolor{OursRow}
\method{}-L8 prune70
& --
& 81.8\gain{1.0}
& 89.9\drop{6.8}
& (\second{85.0}\gain{1.8}, 95.2\drop{1.7})
& 59.4\gain{3.4}
& 79.1\gain{0.4}
& 77.5\drop{3.9}
& 78.2\drop{2.1}
& (79.0\drop{0.2}, 82.5\drop{1.8})
& \bestpm{39.1}{0.6}\gain{12.0} \\

\bottomrule
\end{tabular}
}
\caption{Medical ZSAD results. Avg. summarizes I-AUROC/P-AUROC across the medical suite. Best and second-best results are \textbf{bold} and \underline{underlined}; green/red values show absolute changes from \method{} w/o pruning.}
\label{tab:medical_benchmarks}
\end{table*}

Let $\mathcal{S}$ be the final physical survivor set and $a_i$ the anomaly response of survivor $i$. We restore a dense patch map on the original grid by mapping each position {$p$} to a physical survivor {$o(p)$}: a surviving position maps to itself, and a dropped position maps to the spatially nearest survivor on the original patch grid (global nearest-survivor assignment; no same-block preference). The recovered response is
\begin{equation}
\widehat{A}_p=a_{o(p)}.
\label{eq:nearest_recovery}
\end{equation}
The patch map is then upsampled to the input resolution. Recovery copies only anomaly responses; it never reconstructs deleted token features. In contrast, image-level scoring never uses recovered positions. Patch evidence is reduced only over physical survivors (top-response mean of {$\{a_i:i\in\mathcal{S}\}$}) and is then mixed with a CLS pathway over selected intermediate layers for image-level scoring:
{$s_{\mathrm{base}}=(1-\lambda_p)\,s_{\mathrm{cls}}+\lambda_p\,s_{\mathrm{patch}}$}, where the blend weight {$\lambda_p$}, layer set, and top-response fraction are listed in Appendix~\ref{app:impl_details}.

For the reported image-level results, an inference-only mid-layer refinement branch forms {$s_{\mathrm{mid}}$} from coordinate-aligned responses of physical survivors at intermediate depths, and a source-calibrated soft sample gate {$q(I)\in[0,1]$} blends the two scores as
\begin{equation}
 s_{\mathrm{img}}(I)=(1-q(I))s_{\mathrm{base}}(I)+q(I)s_{\mathrm{mid}}(I).
\label{eq:sparse_mid}
\end{equation}
This branch reads already-computed features at survivor coordinates, so it changes neither token routing nor the recovered anomaly map; the gate temperature, decision thresholds, and mid-layer recipe are listed in Appendix~\ref{app:impl_details}. All reported pruned models use active features through L21 and terminate thereafter; the efficiency gain from this early exit is separated from the gain due to token pruning in Sec.~\ref{sec:exp_setup}.

\section{Experiments}
\label{sec:experiments}

We organize the experiments around four research questions (RQs): RQ1: How do pruning budget and evaluation domain affect the accuracy--throughput trade-off? RQ2: How do pruning stages and core routing components affect efficiency and defect-evidence retention? RQ3: At matched average computation, does image-adaptive budgeting outperform fixed or shuffled budgets? RQ4: How does routing evolve across stages, and how does it compare with generic token reduction in the same frozen detector?

\subsection{Experimental Setup}
\label{sec:exp_setup}

\paragraph{Datasets.}
We follow the cross-dataset ZSAD protocol~\cite{zhou2024anomalyclip} on six industrial benchmarks---MVTec-AD~\cite{bergmann2019mvtec}, VisA~\cite{zou2022visa}, BTAD~\cite{mishra2021btad}, KSDD2~\cite{bozic2021ksdd2}, DAGM~\cite{wieler2007dagm}, and DTD-Synthetic~\cite{aota2023dtd}---and seven medical benchmarks from the same suite. MVTec AD$\rightarrow$VisA and VisA$\rightarrow$MVTec AD denote source-to-target transfer; all other benchmarks use the VisA-trained checkpoint without adaptation.

\paragraph{Baselines.}
We compare against eight ZSAD detectors: six CLIP-based methods---WinCLIP~\cite{jeong2023winclip}, APRIL-GAN~\cite{chen2023aprilgan}, CLIP-AD~\cite{chen2023clipad}, AnomalyCLIP~\cite{zhou2024anomalyclip}, AdaCLIP~\cite{cao2024adaclip}, and Bayes-PFL~\cite{qu2025bayespfl}---and two ViT-based detectors, UniADet~\cite{uniadnet} and VisualAD~\cite{visualAD}.

\paragraph{Evaluation metrics.}
We evaluate image-level detection and pixel-level localization with I-AUROC and P-AUROC, routing quality with defect-token recall (DTR) and complete-miss rate (CMR), and efficiency with token retention and forward-only throughput. DTR and CMR are computed on physical survivors before dense recovery; Appendix~\ref{app:dtr} defines both metrics. All accuracy metrics use the full target test sets. Throughput is measured with batch-1, forward-only fp32 inference at $518\times518$ resolution on one NVIDIA RTX 5090; after 20 warm-up runs, we repeat each measurement five times and report the median. All methods use the same timing protocol, and suite-level speeds are sample-weighted means $\pm$ standard deviations across datasets.

\paragraph{Implementation details.}
We freeze the CLIP ViT-L/14@336px detector and optimize only the L8/L12 scorers (1.97M parameters) for 10 epochs. Under one shared training recipe across both directions, we use the epoch-7 checkpoint for MVTec$\rightarrow$VisA and epoch 10 for VisA$\rightarrow$MVTec. Inference combines fixed L8 keep ratios of $70/50/30\%$ with an adaptive L12 budget and layer-21 early exit; Appendix~\ref{app:impl_details} provides additional optimization details.

\begin{table}[t]
\centering
\setlength{\tabcolsep}{4.2pt}
\renewcommand{\arraystretch}{1.15}
\resizebox{\linewidth}{!}{
\begin{tabular}{@{}lcccccc@{}}
\toprule
\textbf{Variant}
& \textbf{I-AUROC}
& \textbf{P-AUROC}
& \textbf{L8 Keep (\%)}
& \textbf{L12 Keep (\%)}
& \textbf{Final Keep (\%)}
& \textbf{Speed (img/s)} \\
\midrule

\rowcolor{GroupRow}
\multicolumn{7}{@{}c@{}}{\textbf{MVTec AD$\rightarrow$VisA}} \\
\rowcolor{OursRow}
Full method
& \best{86.1}
& 94.3
& 30
& 47
& 14
& \best{41.1} \\

w/o L8 pruning
& \best{86.1}\textcolor{gray}{\scriptsize~(0.0)}
& \best{95.4}\gain{1.1}
& --
& 57
& 57
& 32.8\drop{8.3} \\

w/o L12 pruning
& 84.7\drop{1.4}
& \second{95.1}\gain{0.8}
& 30
& --
& 30
& \second{38.4}\drop{2.7} \\

w/o layer-21 early exit
& \best{86.1}\textcolor{gray}{\scriptsize~(0.0)}
& 94.3\textcolor{gray}{\scriptsize~(0.0)}
& 30
& 47
& 14
& 37.8\drop{3.3} \\

\midrule
\rowcolor{GroupRow}
\multicolumn{7}{@{}c@{}}{\textbf{VisA$\rightarrow$MVTec AD}} \\
\rowcolor{OursRow}
Full method
& \second{90.5}
& 88.6
& 30
& 59
& 18
& \best{38.7} \\

w/o L8 pruning
& \best{93.3}\gain{2.8}
& \best{90.3}\gain{1.7}
& --
& 59
& 59
& 32.2\drop{6.5} \\

w/o L12 pruning
& 90.2\drop{0.3}
& \second{89.5}\gain{0.9}
& 30
& --
& 30
& 35.5\drop{3.2} \\

w/o layer-21 early exit
& \second{90.5}\textcolor{gray}{\scriptsize~(0.0)}
& 88.6\textcolor{gray}{\scriptsize~(0.0)}
& 30
& 59
& 18
& \second{37.8}\drop{0.9} \\

\bottomrule
\end{tabular}
}
\caption{Stage-wise ablation under cross-dataset evaluation. Parentheses show changes from the full method in each transfer direction; best and second-best results are \textbf{bold} and \underline{underlined}.}
\label{tab:ablation_pruning_strategy}
\end{table}

\subsection{Main Results}
\label{sec:exp_main}

Tables~\ref{tab:industrial_benchmarks} and~\ref{tab:medical_benchmarks} compare KeepAD with dense ZSAD baselines and its three pruning budgets. Two observations stand out.

\paragraph{KeepAD yields a favorable industrial accuracy--efficiency trade-off.}
The dense model is competitive with Bayes-PFL and VisualAD, reaching $94.1/95.2$ average I/P-AUROC at $27.8$ img/s. As the budget tightens, prune30 reaches $36.1$ img/s ($1.30\times$) with only $0.2/1.1$-point average losses. At prune70, final retention is $14$--$18\%$ in the two transfer directions (Table~\ref{tab:ablation_pruning_strategy}), while throughput reaches $41.1$ img/s ($1.48\times$) with losses of $1.9/2.2$ points. Small gains on individual datasets also show that accuracy need not decrease monotonically on every target. Overall, speed increases monotonically while average accuracy degrades gradually.

\paragraph{The trade-off transfers without medical adaptation.}
Using the same VisA-trained checkpoint, prune70 delivers a $1.44\times$ speedup with average I/P-AUROC changes of only $-0.2/-1.8$ points. Moderate pruning even improves detection on several datasets, whereas the largest losses occur on BrainMRI, Endo, and Kvasir. Thus, the trade-off remains target-dependent in the medical domain.

\subsection{Ablation Study}
\label{sec:exp_ablation}

\begin{table}[t]
\centering
\setlength{\tabcolsep}{3.0pt}
\renewcommand{\arraystretch}{1.08}
\resizebox{0.94\linewidth}{!}{
\begin{tabular}{@{}lcccc@{}}
\toprule
\textbf{Variant}
& \textbf{I-AUROC}
& \textbf{P-AUROC}
& \textbf{DTR (\%)$\uparrow$}
& \textbf{CMR (\%)$\downarrow$} \\
\midrule

\rowcolor{GroupRow}
\multicolumn{5}{@{}c@{}}{\textbf{MVTec AD$\rightarrow$VisA}} \\
\rowcolor{OursRow}
Full method
& 86.1
& 94.3
& 56.12
& 0.42 \\

L8 global top-$k$
& 82.7\drop{3.4}
& 85.5\drop{8.9}
& 59.69
& 4.00 \\

L12 token top-$k$
& 85.9\drop{0.2}
& 94.4\textcolor{gray}{\scriptsize~(+0.1)}
& 55.93
& 0.50 \\

L12 fixed budget
& 84.3\drop{1.8}
& 93.9\drop{0.5}
& 54.50
& 0.75 \\

L12 visual-only scorer
& 83.4\drop{2.7}
& 93.9\drop{0.4}
& 38.64
& 11.25 \\

L12 anomaly-only scorer
& 86.0\drop{0.1}
& 94.7\gain{0.4}
& 56.50
& 0.50 \\

w/o self-distillation
& 83.4\drop{2.7}
& 94.4\textcolor{gray}{\scriptsize\,(+0.1)}
& 40.39
& 6.92 \\

\midrule
\rowcolor{GroupRow}
\multicolumn{5}{@{}c@{}}{\textbf{VisA$\rightarrow$MVTec AD}} \\
\rowcolor{OursRow}
Full method
& 90.5
& 88.6
& 39.87
& 0.95 \\

L8 global top-$k$
& 87.7\drop{2.8}
& 77.9\drop{10.7}
& 41.51
& 6.28 \\

L12 token top-$k$
& 90.3\drop{0.2}
& 88.6\textcolor{gray}{\scriptsize~(0.0)}
& 39.04
& 0.48 \\

L12 fixed budget
& 88.6\drop{1.9}
& 87.5\drop{1.1}
& 25.08
& 6.20 \\

L12 visual-only scorer
& 89.7\drop{0.8}
& 88.3\drop{0.3}
& 36.81
& 1.99 \\

L12 anomaly-only scorer
& 87.8\drop{2.7}
& 88.8\gain{0.2}
& 29.02
& 7.39 \\

w/o self-distillation
& 85.6\drop{4.9}
& 87.9\drop{0.7}
& 34.60
& 4.29 \\

\bottomrule
\end{tabular}
}
\caption{Core design ablation across the two transfer directions. Green/red values show absolute I-AUROC and P-AUROC changes from the full method; DTR and CMR are computed on final physical survivors before dense recovery.}
\label{tab:core_ablation}
\end{table}
Tables~\ref{tab:ablation_pruning_strategy} and~\ref{tab:core_ablation} isolate the efficiency stack and the main routing choices. We draw four findings.

\paragraph{Early pruning delivers most of the speedup.}
Removing L8 pruning costs $6.5$--$8.3$ img/s, more than removing L12 ($2.7$--$3.2$ img/s). Layer-21 early exit contributes an additional $0.9$--$3.3$ img/s without changing AUROC. Thus, L8 is the primary computational lever, while early exit provides a purely computational gain.

\paragraph{Coverage preservation prevents complete misses.}
Replacing L8 coverage with global top-$k$ slightly raises mean DTR but increases CMR from below $1\%$ to $4.00\%/6.28\%$ and loses $8.9/10.7$ P-AUROC points. The higher average recall therefore masks image-level tail failures; local coverage is valuable because it sharply reduces the risk of removing all defect evidence. By contrast, L12 token top-$k$ closely matches the full method, indicating that the coverage constraint matters primarily at the early pruning stage.

\paragraph{Adaptive dual-cue routing is robust across transfer directions.}
Fixed budgets and single-cue scorers fail in different transfer directions: CMR reaches $6.20\%$ with a fixed budget and up to $11.25\%$ with a single cue. Combining visual and anomaly cues with an image-adaptive budget consistently avoids these direction-specific failures.

\paragraph{Self-distillation improves routing.}
Without self-distillation, CMR rises from below $1\%$ to $6.92\%/4.29\%$, and I-AUROC drops by $2.7/4.9$ points. The small recovered P-AUROC change in one direction shows that physical-survivor metrics expose the teacher's contribution more directly than the recovered anomaly map.
\begin{figure}[t]
    \centering
    \includegraphics[width=0.5\textwidth]{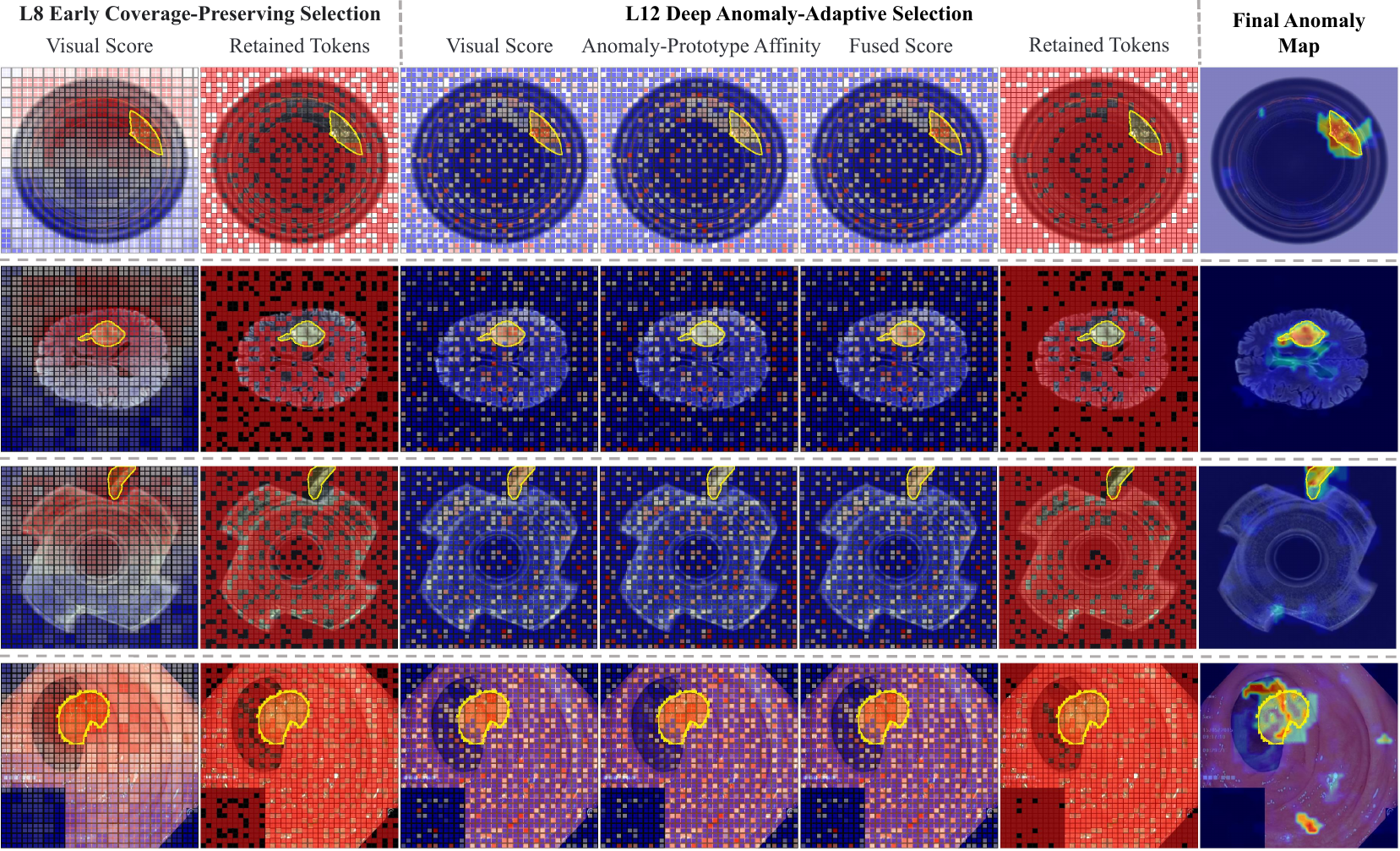}
    \caption{Layer-wise KeepAD routing at prune70. Each row shows one anomalous sample, including L8 and L12 routing results, retained tokens, and the final recovered anomaly map. Grid overlays indicate retained and removed tokens, while yellow contours mark ground-truth defects.}
    \label{fig:tokenvis}
\end{figure}
\subsection{Adaptive-Budget Analysis}
\label{sec:exp_analysis}

\begin{figure}[h]
    \centering
    \includegraphics[width=\linewidth]{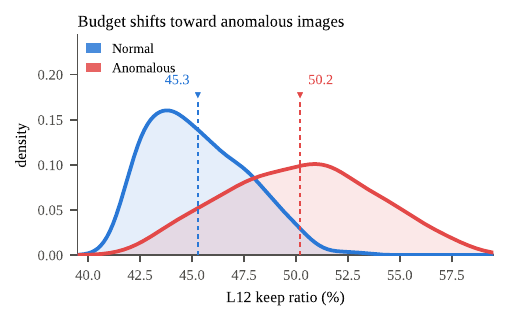}
    \caption{\textbf{Class-conditional L12 keep-ratio distributions at prune70.}
    The adaptive controller spends more on anomalous images ($50.2\%$ vs.\ $45.3\%$ mean keep)
    and keeps most normal images on a lower budget, yielding the main speedup.
    Dashed lines mark class means.}
    \label{fig:budget_distribution}
\end{figure}
\begin{table}[h]
\centering
\setlength{\tabcolsep}{2.5pt}
\renewcommand{\arraystretch}{1.05}
\resizebox{\linewidth}{!}{%
\begin{tabular}{@{}lrrrr@{}}
\toprule
\textbf{Method} & \textbf{I-AUROC} & \textbf{P-AUROC} & \textbf{DTR (\%)$\uparrow$} & \textbf{CMR (\%)$\downarrow$} \\
\midrule
\rowcolor{GroupRow}
\multicolumn{5}{@{}c@{}}{\textbf{MVTec AD$\rightarrow$VisA}} \\
Fixed-Mean & 82.0 & 91.4 & 46.15 & 5.50 \\
Shuffled-Adaptive & 84.8 & 92.3 & 49.17 & 4.50 \\
\rowcolor{OursRow}
Adaptive & \best{86.1} & \best{94.3} & \best{56.12} & \best{0.42} \\
\midrule
\rowcolor{GroupRow}
\multicolumn{5}{@{}c@{}}{\textbf{VisA$\rightarrow$MVTec AD}} \\
Fixed-Mean & 88.5 & 85.6 & 29.85 & 7.95 \\
Shuffled-Adaptive & 89.4 & 85.6 & 27.86 & 4.95 \\
\rowcolor{OursRow}
Adaptive & \best{90.5} & \best{88.6} & \best{39.87} & \best{0.95} \\
\bottomrule
\end{tabular}}
\caption{\addedcap{Adaptive-budget controls at matched mean L12 keep under prune70: $47\%$ on MVTec AD$\rightarrow$VisA and $59\%$ on the reverse transfer.}}
\label{tab:adaptive_controls}
\end{table}

\paragraph{Adaptive budgets follow image evidence.}
Figure~\ref{fig:budget_distribution} shows that anomalous images receive a modestly higher mean L12 keep ratio than normal images ($50.2\%$ versus $45.3\%$), while the broad overlap indicates graded allocation rather than implicit anomaly classification. Table~\ref{tab:adaptive_controls} explicitly controls for total compute using a shared mean budget and a shuffled version of the same adaptive budget multiset.

\paragraph{Image-wise budget matching drives the gain.}
Adaptive consistently leads all metrics in both transfer directions and lowers CMR by at least $4.0$ points relative to Shuffled-Adaptive. Because both use the same budget multiset, the gain comes from directly matching computation to each image's evidence rather than changing aggregate compute.

\subsection{Layer-Wise Routing and Generic Comparison}

\paragraph{Layer-wise routing follows the intended stage roles.}

Figure~\ref{fig:tokenvis} illustrates the intended progression: L8 maintains distributed spatial support, L12 concentrates the reduced budget on high-risk regions, and nearest-survivor recovery maps sparse responses back to the dense grid. This behavior is consistent with the stage-wise findings in Tables~\ref{tab:ablation_pruning_strategy} and~\ref{tab:core_ablation}.

\begin{table}[t]
\centering
\setlength{\tabcolsep}{3.2pt}
\renewcommand{\arraystretch}{1.15}
\resizebox{\linewidth}{!}{
\begin{tabular}{@{}lccccc@{}}
\toprule
\multirow{2}{*}{\textbf{Method}}
& \multicolumn{2}{c}{\textbf{MVTec AD$\rightarrow$VisA}}
& \multicolumn{2}{c}{\textbf{VisA$\rightarrow$MVTec AD}}
& \multirow{2}{*}{\speedhead} \\
\cmidrule(lr){2-3}\cmidrule(lr){4-5}
& \textbf{I-AUROC} & \textbf{P-AUROC}
& \textbf{I-AUROC} & \textbf{P-AUROC}
& \\
\midrule

ToMe
& 80.4 & 82.2
& 85.3 & 80.4
& 36.6 \\

ATS
& 81.4 & 82.9
& 85.3 & 77.2
& 34.3 \\

EViT
& 81.5 & 83.3
& 84.6 & 76.7
& 34.1 \\

Zero-TPrune
& 79.9 & 78.9
& 87.0 & 74.2
& 33.2 \\

\midrule
\rowcolor{OursRow}
\method{}-L8 prune70
& 86.1 & 94.3
& 90.5 & 88.6
& $\mathbf{41.1}$ \\

\bottomrule
\end{tabular}
}
\caption{Comparison with generic training-free token reduction in the same frozen detector. Only the selector changes; speed is reported on MVTec AD$\rightarrow$VisA under the common timing protocol, and \method{}-L8 prune70 is included for reference.}
\label{tab:generic_pruning}
\end{table}

\paragraph{Anomaly-aware routing outperforms generic token reduction.}
Under the same frozen detector, all four generic selectors~\cite{bolya2023tome,fayyaz2022ats,liang2022evit,wang2024zerotprune} trail KeepAD by $8.2$--$15.4$ P-AUROC points and are slower (at most $36.6$ versus $41.1$ img/s). Together with the global top-$k$ ablation, this further supports the importance of anomaly-aware, coverage-preserving routing over token reduction alone.
\FloatBarrier
\section{Conclusion}
\label{sec:conclusion}

In this work, we introduce KeepAD, a defect-preserving token pruning framework for efficient zero-shot anomaly detection. By combining coverage-preserving early selection with prototype-guided, image-adaptive deep pruning, KeepAD reconciles aggressive token reduction with the preservation of sparse anomaly evidence. Dense-to-sparse self-distillation and nearest-survivor recovery enable reliable routing and dense localization within a single sparse forward pass. Experiments across six industrial and seven medical benchmarks demonstrate that KeepAD substantially improves inference efficiency while maintaining competitive image-level detection and pixel-level localization performance.

\bibliography{aaai2027}

\clearpage
\appendix
\setcounter{secnumdepth}{2}

\begin{center}
{\Large\bfseries Technical Appendix}
\end{center}
\vspace{0.6em}

\noindent\textbf{Appendix overview.}
This appendix provides the full supporting material for KeepAD.
Appendix~\ref{app:impl_details} specifies the complete method configuration and reproducibility settings, including the backbone, sparse state, pruning layers, training procedure, and timing protocol.
Appendix~\ref{app:dtr} defines the physical survivor metrics DTR and CMR, which evaluate defect-token retention before dense recovery.
Appendix~\ref{app:backbone} studies backbone generality and pruning-layer placement.
Appendix~\ref{app:controlled} reports budget-matched accuracy--efficiency controls, including pruning-ratio sweeps, block geometry, and rescue design.
Appendix~\ref{app:loss_ablation} presents component-level and inference ablations, together with the evidence-conditioned adaptive-budget analysis.
Appendix~\ref{app:robustness} checks seed stability and medical-suite transfer, and Appendix~\ref{app:rw_zsad} expands the related-work discussion.

\section{Reproducibility and Full Method Specification}
\label{app:impl_details}
This section specifies the experimental protocol and the complete sparse inference path used by all reported KeepAD models. Unless a control explicitly changes one component, the Stage-1 detector, preprocessing, anomaly head, score aggregation, and evaluation code are held fixed. The purpose of this section is to separate invariant implementation choices from the variables examined in later ablations.

\subsection{Evaluation and Timing Protocol}
Accuracy metrics are computed on the full test set of each target dataset. DTR and CMR are measured on physical survivor indices before owner assignment or dense-map recovery (Appendix~\ref{app:dtr}). Throughput is forward-only, batch-1 inference at $518\times518$ resolution in fp32 without autocast on one RTX~5090. After 20 warmup iterations, each run is repeated five times and the median is reported; multi-dataset speed entries report the sample-count-weighted mean of the per-dataset median throughput together with the corresponding weighted population standard deviation across datasets. All throughput comparisons use the same hardware and input protocol. The reported no-pruning timing configuration uses a full-depth dense backbone through L24 with active head layers \texttt{[12,15,18,21]}. Baseline accuracies in the benchmark tables follow the cited evaluation source, whereas all KeepAD accuracies and controlled studies are produced by the evaluation pipeline used in this work.

\subsection{Backbone, Sparse State, and Active Layers}
The reference detector uses OpenAI CLIP ViT-L/14@336px with $518\times518$ inputs. Patch size $14$ produces a $37\times37$ original patch grid ($N_0=1369$). Each live patch token stores its original-grid index. This index yields the coordinate $\pi_i=(r_i,c_i)$, which is recovered only when spatial grouping is required. After every hard pruning operation, feature and original-index tensors are gathered with identical indices. The class token is excluded from patch budgets and local-block statistics. The original $2\times2$ block containing token $i$ is
\begin{equation}
\operatorname{bid}(\pi_i)
=\left(\left\lfloor\frac{r_i}{2}\right\rfloor,
\left\lfloor\frac{c_i}{2}\right\rfloor\right).
\label{eq:app_block_id}
\end{equation}
Because the grid side is odd, blocks on the final row or column contain only the available one or two patches; empty blocks are ignored.

Stage~1 aggregates dense features from layers \texttt{[12,15,18,21,24]}. Stage-2 training retains these five detection layers through L24, whereas reported sparse inference activates layers \texttt{[12,15,18,21]} and terminates after L21. The base image response uses layers \texttt{[12,21]}. For gate-selected samples, the inference-only sparse-mid extension additionally incorporates sparse L15/L18 responses and reduces the union \texttt{[12,15,18,21]}. Table~\ref{tab:implementation_summary} collects the numerical settings used in the reported configuration.

\begin{table*}[t]
\centering
\setlength{\tabcolsep}{5.0pt}
\renewcommand{\arraystretch}{1.10}
\small
\begin{tabular}{@{}p{0.22\textwidth}p{0.25\textwidth}p{0.43\textwidth}@{}}
\toprule
\textbf{Component} & \textbf{Setting} & \textbf{Role} \\
\midrule
Backbone and input & CLIP ViT-L/14@336; $518\times518$ & $37\times37$ original patch grid ($N_0=1369$) \\
Sparse insertion / exit & L8, L12 / after L21 & Early coverage preservation followed by deep anomaly-aware pruning \\
Feature / image-score layers & \texttt{[12,15,18,21]} / \texttt{[12,21]} & Sparse feature aggregation and survivor-based image scoring \\
L8 target keep & $0.70/0.50/0.30$ & prune30 / prune50 / prune70 operating points \\
L8 rescue pools & 64 score packages; 16 diversity candidates & Deterministic score--diversity rescue after the coverage floor \\
L12 budget bounds & $\rho_{\min}=0.25$, $\rho_{\max}=0.50$ & Evidence-conditioned controller parameters \\
L12 controller & $c=0.55$, $T_b=0.04$, $\kappa=0.65$ & Maps image evidence to the target keep mass \\
Evidence tail / relaxation & top $3\%$; temperature $0.12$ & Upper-tail evidence and differentiable block selection \\
Sparse-mid extension & token fraction $0.05$; gate temperature $0.5$ & Inference-only image-score refinement \\
Selector optimization & 1.97M parameters; AdamW; 10 epochs & lr $5\times10^{-5}$, weight decay $10^{-2}$, batch size 4 \\
Loss weights & $0.45/0.22/0.60/0.25$ & FLOPs / preserve / teacher / mask terms \\
\bottomrule
\end{tabular}
\caption{Implementation summary for the reported \method{} configuration. The prune30/50/70 operating points change the L8 keep fraction only; the L12 budget remains image-adaptive.}
\label{tab:implementation_summary}
\end{table*}

\subsection{L8 Coverage Floor and Deterministic Rescue}
Let $\mathcal K_{\mathrm{cov}}^8$ denote the one-token-per-nonempty-block set in Eq.~\eqref{eq:l8_coverage}. For the $37\times37$ patch grid, this coverage set contains $19\times19=361$ tokens. The residual budget and unselected live-token set are
\begin{equation}
R_8=K_8^{\mathrm{tar}}-|\mathcal K_{\mathrm{cov}}^8|,
\qquad
\mathcal R_8=\{1,\ldots,N_8\}\setminus\mathcal K_{\mathrm{cov}}^8.
\label{eq:app_l8_remainder}
\end{equation}
All three reported inference operating points satisfy $R_8>0$. More generally, if $K_8^{\mathrm{tar}}\leq|\mathcal K_{\mathrm{cov}}^8|$, the implementation retains only the highest-scoring subset of block winners up to the target budget, and the one-token-per-block coverage guarantee no longer applies.

For $R_8>0$, the original patch grid is partitioned into an $8\times8$ coarse grid. Each nonempty coarse cell contributes its highest-scoring remaining token as a diversity candidate. At most $\min(16,R_8)$ such candidates are retained, and all remaining residual slots are filled by global score:
\begin{equation}
\begin{aligned}
\mathcal K_{\rm div}^8
&=\operatorname{TopCell}_{\min(16,R_8)}(\mathcal R_8;v^8,\mathcal C_{8\times8}),\\
\mathcal K_{\rm score}^8
&=\operatorname{TopScore}_{R_8-|\mathcal K_{\rm div}^8|}(\mathcal R_8\setminus\mathcal K_{\rm div}^8;v^8),\\
\mathcal K_8
&=\mathcal K_{\rm cov}^8\cup\mathcal K_{\rm div}^8\cup\mathcal K_{\rm score}^8 .
\end{aligned}
\label{eq:app_rescue}
\end{equation}
Thus, the $8\times8$ grid defines 64 coarse spatial cells rather than 64 additional score packages, and the diversity tokens are included within the residual budget rather than appended to it. No random sampling is used.

For prune30, prune50, and prune70, the integer L8 target budgets are $958$, $684$, and $411$, respectively, yielding residual budgets $R_8=597$, $323$, and $50$. Accordingly, the rescue contains up to 16 diversity tokens and the remaining $581$, $307$, or $34$ slots are filled by global score, respectively. Stage-2 training uses a different fixed setting, \texttt{package\_count}$=64$, yielding $361+64=425$ L8 survivors; up to 16 residual slots are allocated to coarse-grid diversity and the remaining slots are filled by global score. The budget-matched controls in Appendix~\ref{app:rescue} and Appendix~\ref{app:rescue_decomposition} isolate the value of score and diversity selection.

\subsection{L12 Evidence-Conditioned Block Selection}
The fused L12 risks are standardized independently within each image and mapped to $u_i\in(0,1)$ as in Eq.~\eqref{eq:adaptive_budget}, where $N_{12}$ denotes the number of live patch tokens entering the L12 decision. With \texttt{topk\_fraction}$=0.03$, the evidence statistics are
\begin{equation}
\begin{aligned}
k_{\mathrm{tail}}
&=\max\!\left(1,\left\lceil 0.03N_{12}\right\rceil\right),\\
E_{\mathrm{var}}(u)
&=
\sqrt{
\operatorname{clip}\!\left(
4\operatorname{Var}(u),0,1
\right)
},\\
E_{\mathrm{tail}}(u)
&=
\operatorname{clip}\!\left(
\frac{\overline{u}_{\mathrm{top}\text{-}k_{\mathrm{tail}}}-\overline{u}}
{\max(1-\overline{u},\varepsilon)},
0,1
\right),\\
E(I)
&=\tfrac12E_{\mathrm{var}}(u)+\tfrac12E_{\mathrm{tail}}(u),
\end{aligned}
\label{eq:app_evidence}
\end{equation}
where $\overline{u}$ is the mean over all live-token risks, $\overline{u}_{\mathrm{top}\text{-}k_{\mathrm{tail}}}$ is the mean of the largest $k_{\mathrm{tail}}$ risks, and $\operatorname{Var}(u)$ is the population variance. Both statistics are computed independently for each image and exclude the class token. We set $\varepsilon=10^{-6}$ for numerical stability. The matched-budget decomposition in Appendix~\ref{app:evidence_decomposition} verifies that the two statistics provide complementary information.

The controller uses
\begin{equation}
\begin{aligned}
\rho_{\min}&=0.25, & \rho_{\max}&=0.50, & c&=0.55,\\
T_b&=0.04, & \kappa&=0.65.&&
\end{aligned}
\label{eq:app_budget_parameters}
\end{equation}
Ignoring finite-token discretization, the L12 keep fraction before atomic block selection lies in $[\rho_{\min}^{\kappa},\rho_{\max}^{\kappa}]\approx[0.406,0.637]$; hence $\rho_{\min}$ and $\rho_{\max}$ are controller parameters rather than exact realized endpoints. Each nonempty L12 block receives the mean fused risk of its live members. In our batch-size-one evaluation, blocks are retained atomically in descending risk order until the cumulative live-token count reaches the image-specific target $K_{12}^{\mathrm{tar}}(I)$. A full $2\times2$ block contains at most four live tokens, so the last selected block can overshoot the nominal target by at most three tokens.

\subsection{Differentiable Selection and Pruning-Aware Learning}
At inference, token pruning itself uses only hard selection and \texttt{gather}, without introducing token-level soft masks. Stage-2 training retains the same hard-\texttt{gather} topology while additionally maintaining a cumulative soft mask to provide surrogate gradients to the selectors. Specifically, after scoring at pruning layer $l$: (i) hard indices $K_l$ gather the surviving tokens and their coordinates on the original patch grid, yielding a physically shortened sequence; (ii) the same selector scores produce a relaxed retention vector $\tilde m_l$; and (iii) the cumulative mask is updated only on physical survivors:
\begin{equation}
\begin{aligned}
m_{\mathrm{cum}}^{+}
&=
\operatorname{Gather}
\left(
m_{\mathrm{cum}}^{-}\odot\tilde m_l,
K_l
\right),\\
x_l^{+}
&=
\operatorname{Gather}(x_l,K_l).
\end{aligned}
\label{eq:app_soft_hard}
\end{equation}
Thus, hard indices determine the actual sparse sequence and computation, whereas the relaxed masks provide surrogate gradient paths through the retained sequence. The discrete \texttt{gather} itself is not differentiated through, so this mechanism is distinct from a conventional straight-through estimator with an identity backward pass.

At L8, the within-block local-representative selection is continuously relaxed with temperature $T_8=0.20$, while positions chosen by the deterministic hard rescue selection are clamped to one in the relaxed mask. At L12, the ranking over $2\times2$ blocks is relaxed with temperature $0.12$, and each relaxed block value is broadcast to its live member tokens. Importantly, the updated cumulative mask is not multiplied into the gathered features immediately. Instead, each subsequent Transformer block first uses $\log(m_{\mathrm{cum}}+\epsilon)$ as an additive attention bias for the corresponding patch-key positions and then reapplies $m_{\mathrm{cum}}$ to the output patch rows. The L12 soft mask is additionally applied explicitly in the segmentation head, whereas the L8 mask influences later head features through cumulative backbone modulation. Both masks also participate in their corresponding auxiliary training objectives.

The compact objective in the main text is instantiated as
\begin{equation}
\begin{aligned}
R_{\mathrm{eff}}
&=0.45L_{\mathrm{flops}},\\
R_{\mathrm{pres}}
&=0.22L_{\mathrm{preserve}}
+0.60L_{\mathrm{teacher}}
+0.25L_{\mathrm{mask}},\\
L_{\mathrm{S2}}
&=L_{\mathrm{det}}+R_{\mathrm{eff}}+R_{\mathrm{pres}}.
\end{aligned}
\label{eq:app_stage2_loss}
\end{equation}
Here, $L_{\mathrm{det}}$ denotes the original image-level classification and dense anomaly-localization losses evaluated along the sparse training path. $L_{\mathrm{flops}}$ penalizes the mean relaxed L12 retention over the live tokens entering L12, while $L_{\mathrm{preserve}}$ discourages the adaptive L12 mask from suppressing high-risk anomaly evidence. The L8 relaxed mask is instead supervised through $L_{\mathrm{mask}}$, $L_{\mathrm{teacher}}$, and the detection objective. $L_{\mathrm{teacher}}$ provides dense-to-sparse self-distillation, and $L_{\mathrm{mask}}$ protects L8 tokens overlapping source-domain defect masks. Synthetic CAA is disabled during Stage~2. Thus, $R_{\mathrm{eff}}$ regularizes an L12 retention-based proxy for sparse computation, whereas $R_{\mathrm{pres}}$ combines complementary evidence-preservation signals.

For dense-to-sparse self-distillation, the dense and sparse branches share the frozen L1--L7 prefix. From this shared state, a no-grad unpruned continuation through L12 produces detached original-grid patch-level anomaly responses $z_i^T$, while the sparse branch separately continues with hard pruning at L8 and L12. Within each nonempty L8 block $\mathcal B_m^8$, the teacher responses are normalized with temperature $T_t=0.08$ to form a block-level teacher distribution $t_m$. The student is represented by the corresponding relaxed L8 retention values, which are not renormalized in this loss. We define the teacher contrast as
\begin{equation}
w_{bm}
=\max_{i\in\mathcal B_m^8} z_{b,i}^T
-\operatorname{mean}_{i\in\mathcal B_m^8} z_{b,i}^T .
\end{equation}
For each sample, we supervise the highest-contrast quarter of nonempty blocks according to $w_{bm}$, with cutoff ties retained. Let $\Omega_b$ denote the resulting block set and let $\mathcal V$ contain samples with positive total contrast. The implemented teacher loss is
\begin{equation}
\begin{aligned}
\ell_{bm}
&=
-\sum_{i\in\mathcal B_m^8}
t_{bm,i}
\log\!\left(
\max\!\left(m_{b,i}^8,\epsilon\right)
\right),\\
\mathcal L_{\mathrm{teacher}}
&=
\frac{1}{|\mathcal V|}
\sum_{b\in\mathcal V}
\frac{
\sum_{m\in\Omega_b}w_{bm}\ell_{bm}
}{
\max\!\left(
\sum_{m\in\Omega_b}w_{bm},
\epsilon
\right)
}.
\end{aligned}
\label{eq:app_teacher}
\end{equation}
The teacher targets and contrast weights are detached from gradient computation. Setting the coefficient of $\mathcal L_{\mathrm{teacher}}$ to zero removes teacher supervision without changing the remaining objectives or the physical pruning topology.

\subsection{Optimization and Execution Procedures}
Stage~2 trains only the two selectors ($1{,}966{,}596$, approximately 1.97M parameters) for 10 epochs with AdamW, learning rate $5\times10^{-5}$, weight decay $10^{-2}$, and batch size $4$. Optimization uses AMP/autocast with GradScaler and seed 42; the Stage-2 data configuration uses the source-domain test split, with synthetic CAA disabled. The released checkpoints are selected at epoch 7 for MVTec-to-VisA and epoch 10 for VisA-to-MVTec. During Stage-2 training, L8 uses the fixed \texttt{package\_count}$=64$ configuration described above. At inference, the prune30/50/70 operating points instead set the L8 keep fraction to $70\%/50\%/30\%$, respectively; the L12 target remains image-adaptive.

For each Stage-2 minibatch, the frozen L1--L7 prefix is computed once and shared by the dense and sparse branches. From this shared state, a no-grad unpruned continuation through L12 produces the detached teacher responses, while the sparse branch separately continues with hard token gathering at L8 and L12 and cumulative relaxed masks for optimization. Stage-2 detection losses retain the original five detection layers \texttt{[12,15,18,21,24]}. We evaluate the complete Stage-2 objective in Eq.~\eqref{eq:app_stage2_loss} and update only the selector parameters.

At inference, KeepAD requires a single hard-pruned forward pass. It first constructs the L8 coverage-and-rescue set and gathers the corresponding features and coordinates. The prototype-guided L12 selector then determines the final survivor set, which is propagated through the active sparse layers and terminates after L21. The base image response uses layers \texttt{[12,21]}; for gate-selected samples, the inference-only sparse-mid extension additionally incorporates sparse L15/L18 responses over \texttt{[12,15,18,21]}. Finally, owner-based recovery maps the sparse patch responses back to the original grid. Neither the dense teacher branch nor the relaxed masks are used at inference.

\section{Evaluation Protocol and Physical Defect-Token Retention}
\label{app:dtr}

Owner-based sparse-to-dense recovery can assign the response
of a nearby survivor to a deleted defect location. Consequently, post-recovery P-AUROC cannot verify whether defect-overlapping tokens survive pruning. High localization performance may instead result from reconstruction using neighboring survivors. We therefore evaluate defect preservation directly from survivor indices before owner assignment, dense recovery, and anomaly-map construction.

\paragraph{Protocol and metrics.}
Ground-truth masks undergo exactly the same $518\times518$ spatial transform as the model input and are mapped onto the $37\times37$ token grid (patch size 14). We use the conservative \emph{any-overlap} rule: a patch belongs to the defect set $G_i$ of anomalous image $i$ if it contains at least one anomalous pixel. Let $\mathcal{S}_i^{l}$ denote the original-grid positions of tokens that physically survive pruning up to layer $l$: $\mathcal{S}_i^{8}$ are the L8 survivors and $\mathcal{S}_i^{12}$ the cumulative survivors after both L8 and L12 pruning. Over the $N_{\mathrm{anom}}$ anomalous test images we report image-macro defect-token recall (DTR)
\begin{equation}
\mathrm{DTR}_l=\frac{1}{N_{\mathrm{anom}}}\sum_{i=1}^{N_{\mathrm{anom}}}\frac{\bigl|\mathcal{S}_i^{l}\cap G_i\bigr|}{|G_i|},
\label{eq:dtr}
\end{equation}
and complete-miss rate (CMR)
\begin{equation}
\mathrm{CMR}_l=\frac{1}{N_{\mathrm{anom}}}\sum_{i=1}^{N_{\mathrm{anom}}}\mathbf{1}\!\left[\,\bigl|\mathcal{S}_i^{l}\cap G_i\bigr|=0\,\right].
\label{eq:cmr}
\end{equation}
The two metrics must be read jointly: a selector can reach a high \emph{mean} DTR by concentrating its budget on a few easy images while still deleting all defect tokens of others, which only CMR exposes.

\paragraph{Results.}
At L8 ($30.0\%$ keep), KeepAD retains $57.9\%/53.0\%$ of defect-overlapping tokens with only $0.33\%/0.00\%$ CMR. At final keeps of $14.3\%/17.8\%$, it retains $56.12\%/39.87\%$ DTR with $0.42\%/0.95\%$ CMR ($5/1200$ and $12/1258$ misses). Budget-matched uniform-random pruning would instead yield DTR equal to the keep ratio and $25.93\%/4.14\%$ CMR.

Table~\ref{tab:cmr_size} stratifies the final CMR by defect size. The advantage of the $2\times2$ coverage design over budget-matched global top-$k$ is most pronounced exactly where the asymmetric risk is worst: for defects covering $1$--$16$ tokens, global top-$k$ misses up to $9.4\%$ of images entirely, while the coverage design stays at $\le2.7\%$.

\paragraph{Interpreting the core ablations.}
Global top-$k$ can raise mean DTR by concentrating survivors in easy high-response regions while still missing whole defects; this explains its higher DTR but much worse CMR and P-AUROC in Table~\ref{tab:core_ablation}. Likewise, post-recovery P-AUROC and physical retention measure different effects because owner recovery can fill a deleted position from a nearby survivor.

Without dense-to-sparse self-distillation, DTR falls from $56.12\%/39.87\%$ to $40.39\%/34.60\%$, CMR rises from $0.42\%/0.95\%$ to $6.92\%/4.29\%$, and I-AUROC drops by $2.7/4.9$ points. These physical-survivor metrics expose the teacher's effect more directly than the recovered pixel map.

\begin{table}[t]
\centering
\setlength{\tabcolsep}{5pt}
\renewcommand{\arraystretch}{1.12}
\resizebox{\dimexpr\linewidth-6pt\relax}{!}{%
\begin{tabular}{@{}llrrr@{}}
\toprule
\textbf{Setting} & $|G_i|$ & \textbf{\#Images} & \textbf{CMR local (\%)} & \textbf{CMR global (\%)} \\
\midrule
MVTec AD$\rightarrow$VisA & 1--4 & 225 & \best{2.2} & 5.8 \\
MVTec AD$\rightarrow$VisA & 5--16 & 602 & \best{0.0} & 5.3 \\
VisA$\rightarrow$MVTec AD & 1--4 & 9 & 0.0 & 0.0 \\
VisA$\rightarrow$MVTec AD & 5--16 & 299 & \best{2.7} & 9.4 \\
\bottomrule
\end{tabular}}
\caption{CMR (\%) at the final prune70 budget, stratified by defect size $|G_i|$ (number of GT-overlapping tokens). ``local'' is the full $2\times2$ coverage design; ``global'' is budget-matched L8 global top-$k$. Strata with $|G_i|>16$ are not broken out here.}
\label{tab:cmr_size}
\end{table}

\paragraph{Caveats.}
Two caveats motivate our choice of metrics and wording. First, the L8 coverage rule guarantees one representative token per non-empty $2\times2$ block. However, the retained token may not overlap the ground-truth defect. Thus, block coverage does not guarantee defect retention. We therefore compute DTR and CMR using survivor indices rather than region coverage. Second, owner recovery copies the features of survivors and never restores the original feature of a deleted token. So we do not use post-recovery P-AUROC as a substitute for DTR. Accordingly, the paper's claims are phrased as \emph{preferentially retaining defect-overlapping tokens while substantially reducing complete-miss risk}, not as guaranteeing the survival of every defect token.

\section{Backbone Generality and Pruning-Layer Placement}

This section tests whether the proposed routing principle is specific to one feature space or one hand-picked layer pair. Backbone experiments evaluate the full pipeline, whereas the depth analyses separate fixed-scorer salience transfer from end-to-end retraining at matched final keep.

\subsection{Backbone Generality}
\label{app:backbone}

Table~\ref{tab:backbones} reports the backbone-comparison protocol for four backbones under the two cross-dataset settings. All four backbones use the fixed 15-epoch comparison protocol; every delta is relative to the corresponding backbone's no-pruning row in this table. All I/P-AUROC, DTR, and CMR values use the full test sets. The pruned rows run the full pipeline with both scorers, adaptive L12 budget, rescue, and owner recovery. Three observations stand out. (i) The P-AUROC cost of pruning remains small: the largest drop across the $24$ direction--budget results is $3.5$ (DINOv3 ViT-L/16, VisA$\rightarrow$MVTec AD, prune70). (ii) I-AUROC sensitivity is backbone-dependent: CLIP ViT-B/16 is essentially insensitive on MVTec AD$\rightarrow$VisA (deltas within $\pm0.7$ even at prune70), and DINOv2 ViT-L/14-Reg even \emph{gains} up to $+1.8$ I-AUROC at prune30/50, again consistent with the denoising effect of removing low-risk tokens from the image-level aggregation. (iii) DINOv3 variants lose the most I-AUROC, particularly on VisA$\rightarrow$MVTec AD (up to $-7.1$ for ViT-L/16 at prune70), indicating that the salience scorer transfers less reliably to this feature space; understanding this gap is left for future work. The DTR/CMR columns expose the physical defect-token survival at the final prune layer and should be read jointly rather than as a replacement for AUROC.

\begin{table*}[t]
\centering
\setlength{\tabcolsep}{2.2pt}
\renewcommand{\arraystretch}{1.05}
\scriptsize\resizebox{0.98\textwidth}{!}{%
\begin{tabular}{@{}llcccccccccc@{}}
\toprule
\multirow{2}{*}{\textbf{Backbone}}
& \multirow{2}{*}{\textbf{Point}}
& \multicolumn{5}{c}{\textbf{MVTec AD$\rightarrow$VisA}}
& \multicolumn{5}{c}{\textbf{VisA$\rightarrow$MVTec AD}} \\
\cmidrule(lr){3-7}\cmidrule(lr){8-12}
& & \textbf{I-AUROC} & \textbf{P-AUROC}
& \begin{tabular}[c]{@{}c@{}}\textbf{Final keep (\%)}\end{tabular}
& \textbf{DTR (\%)$\uparrow$} & \textbf{CMR (\%)$\downarrow$}
& \textbf{I-AUROC} & \textbf{P-AUROC}
& \begin{tabular}[c]{@{}c@{}}\textbf{Final keep (\%)}\end{tabular}
& \textbf{DTR (\%)$\uparrow$} & \textbf{CMR (\%)$\downarrow$} \\
\midrule
\rowcolor{GroupRow}
\multirow{4}{*}{\shortstack[l]{CLIP\\ViT-B/16}}
& w/o pruning & 80.2 & 94.2 & 100 & 100 & 0 & 84.1 & 89.9 & 100 & 100 & 0 \\
& prune30 & 80.0\drop{0.2} & 93.3\drop{0.9} & 43.0 & 90.80 & 1.58 & 84.5\gain{0.4} & 87.9\drop{1.9} & 42.5 & 78.37 & 0.64 \\
& prune50 & 80.9\gain{0.7} & 93.0\drop{1.2} & 30.3 & 83.86 & 1.08 & 84.6\gain{0.5} & 87.8\drop{2.1} & 30.1 & 66.40 & 0.32 \\
& prune70 & 80.3\gain{0.1} & 92.4\drop{1.8} & 18.3 & 61.55 & 1.92 & 82.9\drop{1.2} & 87.6\drop{2.2} & 21.2 & 45.30 & 0.16 \\
\cmidrule(lr){2-12}
\rowcolor{GroupRow}
\multirow{4}{*}{\shortstack[l]{DINOv2\\ViT-L/14-Reg}}
& w/o pruning & 91.0 & 96.0 & 100 & 100 & 0 & 92.4 & 91.7 & 100 & 100 & 0 \\
& prune30 & 92.7\gain{1.7} & 96.0\same{0.0} & 42.3 & 96.94 & 0.42 & 91.2\drop{1.3} & 90.2\drop{1.5} & 41.5 & 66.05 & 3.34 \\
& prune50 & 92.8\gain{1.8} & 95.9\drop{0.1} & 29.6 & 92.87 & 0.58 & 90.5\drop{1.9} & 90.3\drop{1.4} & 29.5 & 57.74 & 3.02 \\
& prune70 & 90.0\drop{1.0} & 95.6\drop{0.4} & 17.9 & 69.48 & 2.08 & 90.1\drop{2.3} & 90.4\drop{1.3} & 17.8 & 39.17 & 2.62 \\
\cmidrule(lr){2-12}
\rowcolor{GroupRow}
\multirow{4}{*}{\shortstack[l]{DINOv3\\ViT-B/16}}
& w/o pruning & 91.0 & 96.2 & 100 & 100 & 0 & 92.3 & 92.8 & 100 & 100 & 0 \\
& prune30 & 87.0\drop{4.0} & 95.2\drop{1.0} & 46.9 & 63.54 & 11.42 & 91.1\drop{1.2} & 90.1\drop{2.7} & 49.7 & 84.45 & 0.08 \\
& prune50 & 86.6\drop{4.3} & 95.3\drop{0.9} & 33.8 & 65.41 & 11.00 & 89.9\drop{2.4} & 90.0\drop{2.8} & 35.5 & 75.10 & 0.48 \\
& prune70 & 86.2\drop{4.8} & 95.0\drop{1.2} & 20.1 & 49.75 & 14.67 & 89.4\drop{2.9} & 89.5\drop{3.3} & 21.1 & 54.35 & 0.79 \\
\cmidrule(lr){2-12}
\rowcolor{GroupRow}
\multirow{4}{*}{\shortstack[l]{DINOv3\\ViT-L/16}}
& w/o pruning & 92.7 & 96.3 & 100 & 100 & 0 & 93.0 & 91.8 & 100 & 100 & 0 \\
& prune30 & 91.6\drop{1.1} & 96.1\drop{0.2} & 41.1 & 95.38 & 0.00 & 88.4\drop{4.7} & 88.7\drop{3.1} & 49.8 & 77.13 & 0.72 \\
& prune50 & 89.2\drop{3.5} & 95.9\drop{0.5} & 34.6 & 83.97 & 0.33 & 86.9\drop{6.1} & 88.8\drop{3.0} & 35.7 & 58.25 & 1.35 \\
& prune70 & 87.0\drop{5.7} & 95.3\drop{1.1} & 20.8 & 59.52 & 1.42 & 85.9\drop{7.1} & 88.3\drop{3.5} & 21.3 & 38.37 & 1.51 \\
\bottomrule
\end{tabular}}
\caption{Backbone generality across four backbones. Colored annotations on pruned rows give absolute I-AUROC and P-AUROC changes from the corresponding no-pruning row in this table. Final keep is the realized patch-token retention relative to the original grid. DTR and CMR are measured on physical survivor indices at the final prune layer before owner fill; all entries use the full test sets.}
\label{tab:backbones}
\end{table*}

\subsection{Fixed-Scorer Depth Scan}
\label{app:layerscan}

To measure how well token salience separates defect from normal tokens at each depth, we transplant the trained L8 scorer weights to layer $l$, score all patch tokens of every anomalous test image, and compute the AUROC of this score against the any-overlap token labels of Appendix~\ref{app:dtr} (\emph{Token AUC}). No per-layer retraining is performed, so the scan measures depth sensitivity under a fixed scorer, not a per-layer upper bound after re-tuning.

\begin{figure*}[t]
    \centering
    \includegraphics[width=0.90\textwidth]{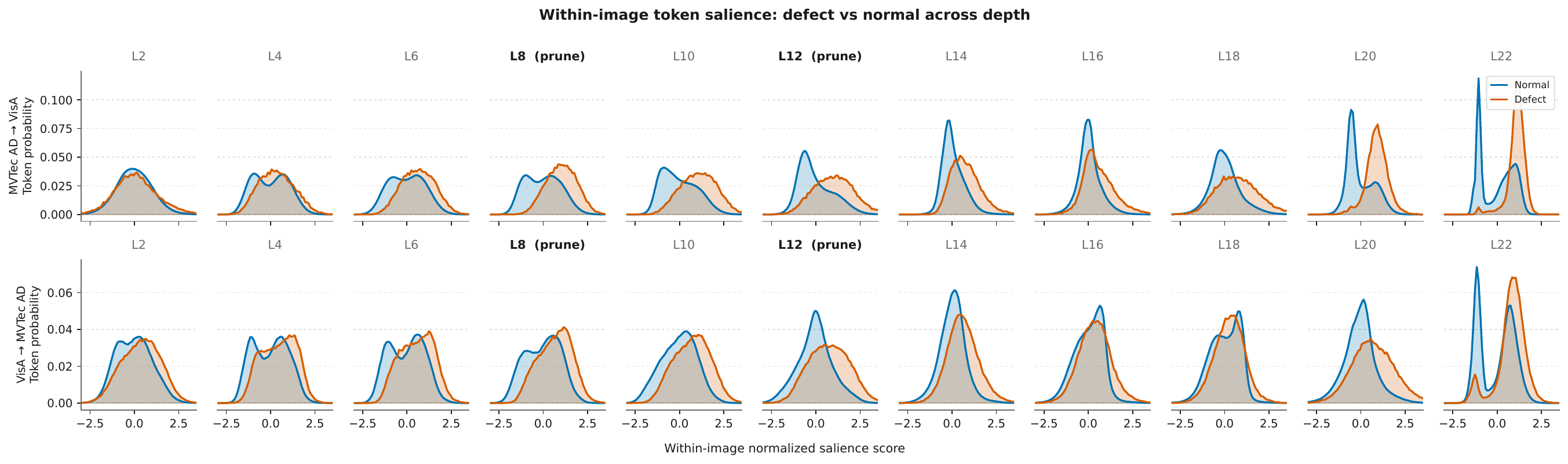}
    \caption{\textbf{Within-image salience-score distributions across encoder depth.} Probability densities of within-image normalized salience scores are shown for normal and defect tokens at every even encoder layer from L2 to L22. The two panels report MVTec AD$\rightarrow$VisA (top) and VisA$\rightarrow$MVTec AD (bottom); L8 and L12 are the pruning layers, and token labels use the any-overlap rule in Appendix~\ref{app:dtr}.}
    \label{fig:score_distributions}
\end{figure*}

Figure~\ref{fig:score_distributions} shows the salience distributions at every even encoder layer from L2 to L22. The corresponding Token AUC is $77.6$/$68.8$ at L8, dips to $68.8$/$57.2$ at L18, and recovers to $83.6$/$73.0$ at L22. Token separability is weak at L2--L6 and clearly \emph{non-monotonic} in depth. Although L20/L22 reach the highest separability, pruning that late saves almost no computation, since nearly the whole network has already run at full width. The choice of L8/L12 should therefore be read as a trade-off among computational savings, spatial coverage, and separability---not as the depths with the globally strongest salience.

\subsection{End-to-End Pruning-Layer Placement}
\label{app:layer_pair_control}

The fixed-scorer depth scan above measures salience transfer but does not establish the best end-to-end pruning locations. Table~\ref{tab:layer_pair_control} therefore retrains the selectors at an earlier pair, the proposed pair, and a later pair. All variants use the same $2\times2$ geometry, rescue rule, adaptive-budget mechanism, Stage-2 schedule, and layer-21 exit. Their final physical keeps are calibrated on source validation to match the proposed pair separately in each transfer direction.

\begin{table}[t]
\centering
\setlength{\tabcolsep}{2.0pt}
\renewcommand{\arraystretch}{1.12}
\resizebox{\linewidth}{!}{%
\footnotesize
\begin{tabular}{@{}lcccccc@{}}
\toprule
\textbf{Layers} & \textbf{I-AUROC} & \textbf{P-AUROC} & \textbf{Keep (\%)} & \textbf{DTR (\%)$\uparrow$} & \textbf{CMR (\%)$\downarrow$} & \textbf{img/s $\uparrow$} \\
\midrule
\rowcolor{GroupRow}
\multicolumn{7}{@{}c@{}}{\textbf{MVTec AD$\rightarrow$VisA}} \\
L6 / L10 & 84.2 & 92.1 & 14.3 & 48.50 & 1.83 & 44.8 \\
\rowcolor{OursRow}
L8 / L12 (ours) & 86.1 & 94.3 & 14.3 & 56.12 & 0.42 & 41.1 \\
L10 / L14 & 86.4 & 94.6 & 14.3 & 57.80 & 0.33 & 36.5 \\
\midrule
\rowcolor{GroupRow}
\multicolumn{7}{@{}c@{}}{\textbf{VisA$\rightarrow$MVTec AD}} \\
L6 / L10 & 88.9 & 86.4 & 17.8 & 34.20 & 2.38 & 42.3 \\
\rowcolor{OursRow}
L8 / L12 (ours) & 90.5 & 88.6 & 17.8 & 39.87 & 0.95 & 38.7 \\
L10 / L14 & 90.8 & 88.9 & 17.8 & 41.20 & 0.79 & 34.2 \\
\bottomrule
\end{tabular}}
\caption{End-to-end pruning-layer placement at a matched final physical keep. Each selector pair is independently trained.}
\label{tab:layer_pair_control}
\end{table}

Moving the selectors from L8/L12 to L6/L10 raises throughput by $+3.7$/$+3.6$ img/s but lowers P-AUROC by $2.2$/$2.2$ points and raises CMR by $1.41$/$1.43$ points. Moving them to L10/L14 lowers throughput by $4.6$/$4.5$ img/s while gaining only $+0.3$/$+0.3$ I-AUROC and $+0.3$/$+0.3$ P-AUROC. The resulting accuracy--throughput ordering places L8/L12 at the non-dominated knee in both directions: L6/L10 is faster but localization-brittle under early semantic noise, whereas L10/L14 yields only marginal accuracy gains at a clear speed cost.

\section{Controlled Accuracy--Efficiency Studies}
\label{app:controlled}

The controls in this section hold the surrounding pipeline fixed and change only the named budget or selector component. I-AUROC, P-AUROC, DTR, and CMR use the full test sets. Unless a caption states otherwise, controlled throughput is measured on a fixed 100-image subset per transfer direction using serial isolated reruns and is intended for within-table comparison; headline throughput follows the protocol in Appendix~\ref{app:impl_details}. The official prune70 full-method results are $(86.1,94.3)$ and $41.1$ img/s for MVTec AD$\rightarrow$VisA, and $(90.5,88.6)$ and $38.7$ img/s for VisA$\rightarrow$MVTec AD.

\subsection{Verified Efficiency Contributions}
\label{app:efficiency_stack}

Table~\ref{tab:efficiency_stack} retains only measured configurations. Dense-24 is the unpruned Stage-1 reference. The single-stage rows enable only L8 or only L12 pruning and use the L21 exit; the two-stage row without early exit keeps both selectors but continues to L24; the final row is the reported full method. We intentionally omit an unverified matched-depth dense row, so Dense-24 is a total-reference point rather than a causal estimate of early-exit gain by itself.

\begin{table}[t]
\centering
\setlength{\tabcolsep}{1.8pt}
\renewcommand{\arraystretch}{1.10}
\resizebox{\linewidth}{!}{%
\footnotesize
\begin{tabular}{@{}lcccrrrr@{}}
\toprule
\textbf{Variant} & \textbf{L8} & \textbf{L12} & \textbf{EE} & \textbf{I-AUROC} & \textbf{P-AUROC} & \textbf{Final Keep (\%)} & \textbf{img/s $\uparrow$} \\
\midrule
\rowcolor{GroupRow}
\multicolumn{8}{@{}c@{}}{\textbf{MVTec AD$\rightarrow$VisA}} \\
Dense-24 & & & & 88.8 & 95.8 & 100 & 27.8 \\
L8 only & \checkmark & & \checkmark & 84.7 & 95.1 & 30 & 38.4 \\
L12 only & & \checkmark & \checkmark & 86.1 & 95.4 & 57 & 32.8 \\
L8$+$L12 (no EE) & \checkmark & \checkmark & & 86.1 & 94.3 & 14 & 37.8 \\
\rowcolor{OursRow}
Full (L8$+$L12$+$EE) & \checkmark & \checkmark & \checkmark & 86.1 & 94.3 & 14 & \best{41.1} \\
\midrule
\rowcolor{GroupRow}
\multicolumn{8}{@{}c@{}}{\textbf{VisA$\rightarrow$MVTec AD}} \\
Dense-24 & & & & 92.7 & 91.2 & 100 & 27.8 \\
L8 only & \checkmark & & \checkmark & 90.2 & 89.5 & 30 & 35.5 \\
L12 only & & \checkmark & \checkmark & 93.3 & 90.3 & 59 & 32.2 \\
L8$+$L12 (no EE) & \checkmark & \checkmark & & 90.5 & 88.6 & 18 & 37.8 \\
\rowcolor{OursRow}
Full (L8$+$L12$+$EE) & \checkmark & \checkmark & \checkmark & 90.5 & 88.6 & 18 & \best{38.7} \\
\bottomrule
\end{tabular}}
\caption{Verified dense-to-pruned efficiency stack at the prune70 operating point. Checkmarks denote active L8 pruning, L12 pruning, and the L21 early exit (EE). All rows are measured results already used by the main benchmark or pruning-stage ablation.}
\label{tab:efficiency_stack}
\end{table}

The verified rows support two conclusions. First, early deletion matters: L8-only reaches $38.4/35.5$ img/s, whereas L12-only reaches $32.8/32.2$ img/s because it preserves the full sequence through more layers. Second, the L21 exit provides an additional gain on the already-pruned path: enabling it changes throughput from $37.8$ to $41.1/38.7$ img/s while leaving I/P-AUROC unchanged. Accuracy is not monotone in the number of active components---for example, L12-only retains more tokens for image aggregation---so the table should be interpreted as an efficiency decomposition, not as a nested accuracy ladder.

\subsection{Same-Pipeline Pruning-Ratio Sweep}
\label{app:prunecurve}

\begin{figure*}[t]
    \centering
    \includegraphics[width=0.48\textwidth]{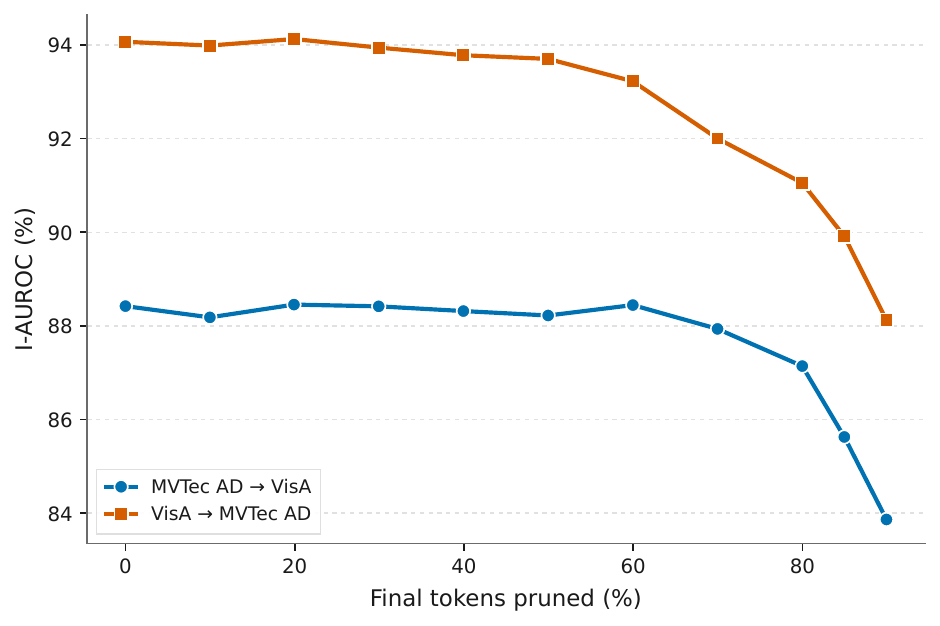}\hfill
    \includegraphics[width=0.48\textwidth]{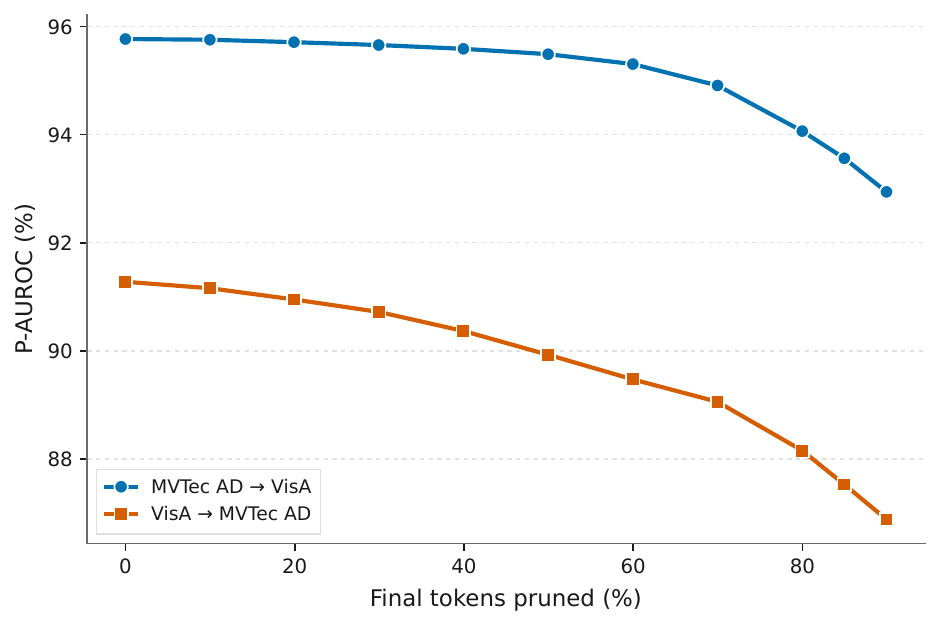}
    \caption{\textbf{Same-pipeline pruning-ratio curves.} I-AUROC (left) and P-AUROC (right) are plotted against the final total percentage of tokens pruned for both transfer directions. The common grid contains 0\%, 10\%, 20\%, 30\%, 40\%, 50\%, 60\%, 70\%, 80\%, 85\%, and 90\%; every point runs the same scorer, selector, and sparse-path pipeline.}
    \label{fig:prune_curve}
\end{figure*}

Figure~\ref{fig:prune_curve} plots the same pipeline over a common \emph{final total} token-pruning grid of 0--90\% (the overall fraction of original patch tokens removed after all pruning stages, not an L8-only or stage-local target). At prune${}=0$ the scorers, selectors, and sparse-path inference all run and only the physical token deletion is disabled; the $\approx0.96\times$ relative throughput at $0\%$ therefore isolates the scorer/selector overhead ($\approx 4\%$), and the Stage-1 dense baseline (which contains none of these modules) serves as the $1.0\times$ reference. Accuracy degrades gracefully: on MVTec AD$\rightarrow$VisA, I-AUROC is flat through $60\%$ pruning and P-AUROC loses only $0.5$; on VisA$\rightarrow$MVTec AD, I-AUROC is flat through $30\%$ and loses $\le0.4$ through $50\%$. Beyond $80$--$90\%$, CMR rises sharply ($10.57\%$ at $90\%$ on VisA$\rightarrow$MVTec AD), marking the regime where the coverage floor itself is squeezed out.

\subsection{Budget-Matched Block Geometry}
\label{app:blockgeom}

Table~\ref{tab:blockgeom} fixes the final keep of every geometry to exactly $14.97\%$ in both directions and varies only the local block size. With $1\times1$ blocks every block holds a single token, so the per-block floor degenerates into a global count: the $1\times1$ row \emph{is} budget-matched global top-$k$. Global selection attains the highest mean DTR by concentrating the budget on a few high-response regions, but pays with the worst CMR and P-AUROC in both directions. The $2\times2$ geometry achieves the best P-AUROC and the lowest CMR in both directions at a $\approx2\%$ throughput cost. I-AUROC is mixed ($4\times4$ is best on MVTec AD$\rightarrow$VisA, $2\times2$ on VisA$\rightarrow$MVTec AD), so larger blocks are \emph{not} uniformly worse; $2\times2$ is the most stable coverage/localization trade-off in this controlled comparison.

\begin{table}[t]
\centering
\setlength{\tabcolsep}{2.0pt}
\renewcommand{\arraystretch}{1.12}
\footnotesize
\begin{tabular}{@{}lrrrrr@{}}
\toprule
\textbf{Block} & \textbf{I-AUROC} & \textbf{P-AUROC} & \textbf{DTR (\%)$\uparrow$} & \textbf{CMR (\%)$\downarrow$} & \textbf{img/s $\uparrow$} \\
\midrule
\rowcolor{GroupRow}
\multicolumn{6}{@{}c@{}}{\textbf{MVTec AD$\rightarrow$VisA}} \\
$1\times1$ (global) & 84.0 & 85.7 & \best{70.01} & 3.83 & \best{34.39} \\
\rowcolor{OursRow}
$2\times2$ (control) & 84.4 & \best{94.4} & 56.06 & \best{0.42} & 33.66 \\
$3\times3$ & 85.2 & 93.4 & 66.36 & 1.00 & 33.60 \\
$4\times4$ & \best{85.6} & 92.5 & 68.40 & 1.83 & 34.30 \\
\midrule
\rowcolor{GroupRow}
\multicolumn{6}{@{}c@{}}{\textbf{VisA$\rightarrow$MVTec AD}} \\
$1\times1$ (global) & 87.6 & 78.0 & \best{41.05} & 5.72 & 34.03 \\
\rowcolor{OursRow}
$2\times2$ (control) & \best{90.7} & \best{88.2} & 33.82 & \best{1.11} & 33.29 \\
$3\times3$ & 89.8 & 87.0 & 39.55 & 1.67 & \best{34.08} \\
$4\times4$ & 88.7 & 86.3 & 40.45 & 1.91 & 33.38 \\
\bottomrule
\end{tabular}
\caption{Controlled block geometry under an identical final keep of $14.97\%$. The $1\times1$ geometry is equivalent to budget-matched global top-$k$ (the local floor degenerates to a global count).}
\label{tab:blockgeom}
\end{table}

\subsection{Score-Based vs. Random Rescue}
\label{app:rescue}

Table~\ref{tab:rescue} compares score-based rescue with uniform-random rescue at identical rescued-token counts (final keep $16.51\%$); the random baseline uses three deterministic seeds ($11/29/47$). Score-based rescue improves I-AUROC by $+1.0$/$+0.6$, raises DTR by $+16.9$/$+11.7$, lowers CMR in both directions, and attains higher P-AUROC ($+0.6$/$+0.4$). Its largest and most robust margins are on image-level evidence and tail risk---\emph{which} tokens survive at all---while pixel-level maps remain governed primarily by the coverage floor.

\begin{table}[t]
\centering
\setlength{\tabcolsep}{1.5pt}
\renewcommand{\arraystretch}{1.15}
\footnotesize
\begin{tabular}{@{}lccccc@{}}
\toprule
\textbf{Rescue} & \textbf{I-AUROC} & \textbf{P-AUROC} & \textbf{DTR (\%)$\uparrow$} & \textbf{CMR (\%)$\downarrow$} & \textbf{img/s $\uparrow$} \\
\midrule
\rowcolor{GroupRow}
\multicolumn{6}{@{}c@{}}{\textbf{MVTec AD$\rightarrow$VisA}} \\
random & 84.50$\pm$0.28 & 93.80 & 44.87$\pm$0.18 & 0.47$\pm$0.08 & \best{33.93} \\
\rowcolor{OursRow}
score & \best{85.50} & \best{94.40} & \best{61.81} & \best{0.25} & 33.73 \\
\midrule
\rowcolor{GroupRow}
\multicolumn{6}{@{}c@{}}{\textbf{VisA$\rightarrow$MVTec AD}} \\
random & 90.47$\pm$0.09 & 87.93 & 26.10 & 1.32 & \best{34.93} \\
\rowcolor{OursRow}
score & \best{91.10} & \best{88.30} & \best{37.75} & \best{0.87} & 33.91 \\
\bottomrule
\end{tabular}
\caption{Score-based vs.\ uniform-random rescue at identical rescued-token counts (final keep $16.51\%$). Random-rescue entries are three-seed means; $\pm$std is reported for the columns where seed-level variation was logged.}
\label{tab:rescue}
\end{table}

\subsection{Rescue-Component Decomposition}
\label{app:rescue_decomposition}

The score-versus-random control above establishes that rescue should not be random, but the implemented rule combines high-score and spatial-diversity candidates. Table~\ref{tab:rescue_decomposition} separates those two sources at exactly the same rescued-token count and final keep. Ties, the L8 coverage floor, the L12 selector, checkpoint, and inference protocol are held fixed.

\begin{table}[t]
\centering
\setlength{\tabcolsep}{1.8pt}
\renewcommand{\arraystretch}{1.10}
\resizebox{\linewidth}{!}{%
\footnotesize
\begin{tabular}{@{}lccccc@{}}
\toprule
\textbf{Rescue rule} & \textbf{I-AUROC} & \textbf{P-AUROC} & \textbf{DTR (\%)$\uparrow$} & \textbf{CMR (\%)$\downarrow$} & \textbf{Keep (\%)} \\
\midrule
\rowcolor{GroupRow}
\multicolumn{6}{@{}c@{}}{\textbf{MVTec AD$\rightarrow$VisA}} \\
uniform random & 84.50 & 93.80 & 44.87 & 0.47 & 16.51 \\
high-score only & 85.30 & 94.20 & 58.40 & 0.33 & 16.51 \\
spatial-diversity only & 84.90 & 94.10 & 50.20 & 0.25 & 16.51 \\
\rowcolor{OursRow}
score + diversity (full) & 85.50 & 94.40 & 61.81 & 0.25 & 16.51 \\
\midrule
\rowcolor{GroupRow}
\multicolumn{6}{@{}c@{}}{\textbf{VisA$\rightarrow$MVTec AD}} \\
uniform random & 90.47 & 87.93 & 26.10 & 1.32 & 16.51 \\
high-score only & 90.90 & 88.15 & 35.20 & 1.03 & 16.51 \\
spatial-diversity only & 90.70 & 88.05 & 30.50 & 0.95 & 16.51 \\
\rowcolor{OursRow}
score + diversity (full) & 91.10 & 88.30 & 37.75 & 0.87 & 16.51 \\
\bottomrule
\end{tabular}}
\caption{Decomposition of the L8 rescue set at the same rescued-token count and final keep ($16.51\%$).}
\label{tab:rescue_decomposition}
\end{table}

High-score-only rescue raises DTR by $+13.53$/$+9.10$ points over random rescue, whereas diversity-only rescue lowers CMR by $0.22$/$0.37$ points. Their combination raises I/P-AUROC by $+0.20$/$+0.20$ and $+0.20$/$+0.15$ and matches or slightly improves CMR relative to the stronger single-source alternative ($0.25$/$0.87$ versus $0.25$/$0.95$). This pattern indicates that the two candidate sources are complementary---high-score rescue recovers defect-bearing tokens while diversity rescue reduces complete misses---so the combined rule is retained.

\subsection{Exploratory Accuracy--Throughput Envelopes}

\begin{figure*}[t]
    \centering
    \includegraphics[width=0.91\textwidth]{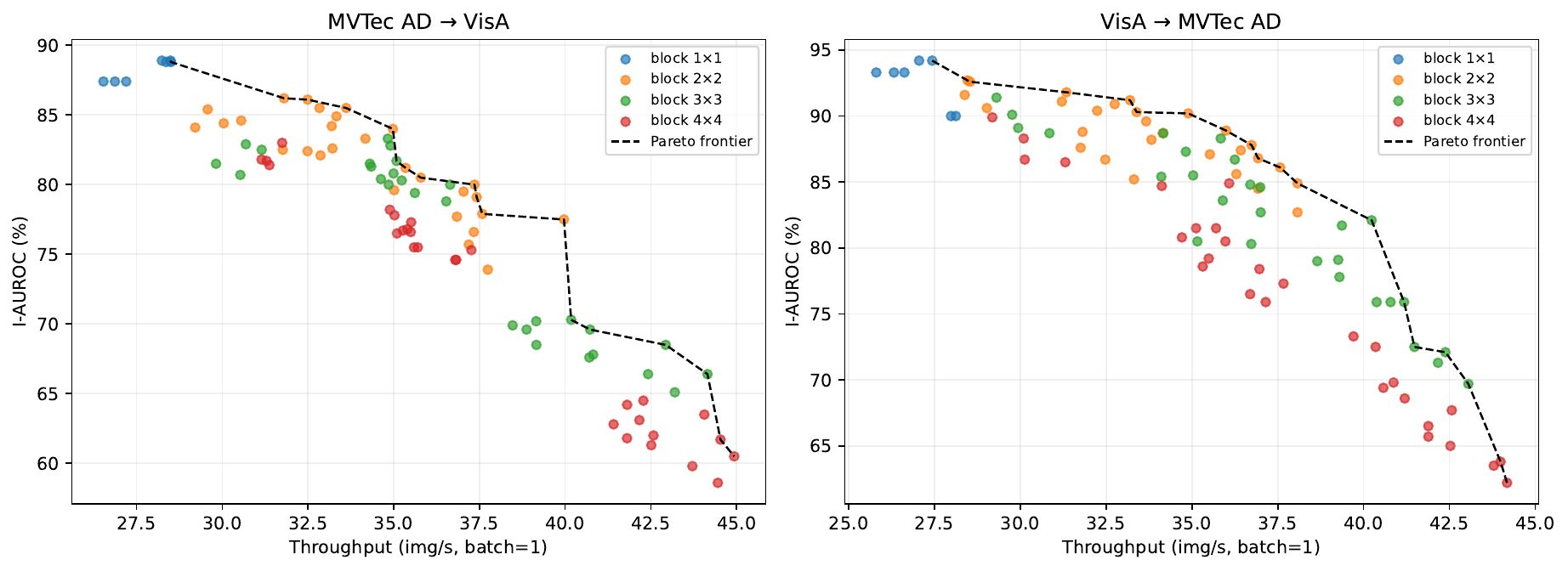}\par\vspace{-0.7em}
    \includegraphics[width=0.91\textwidth]{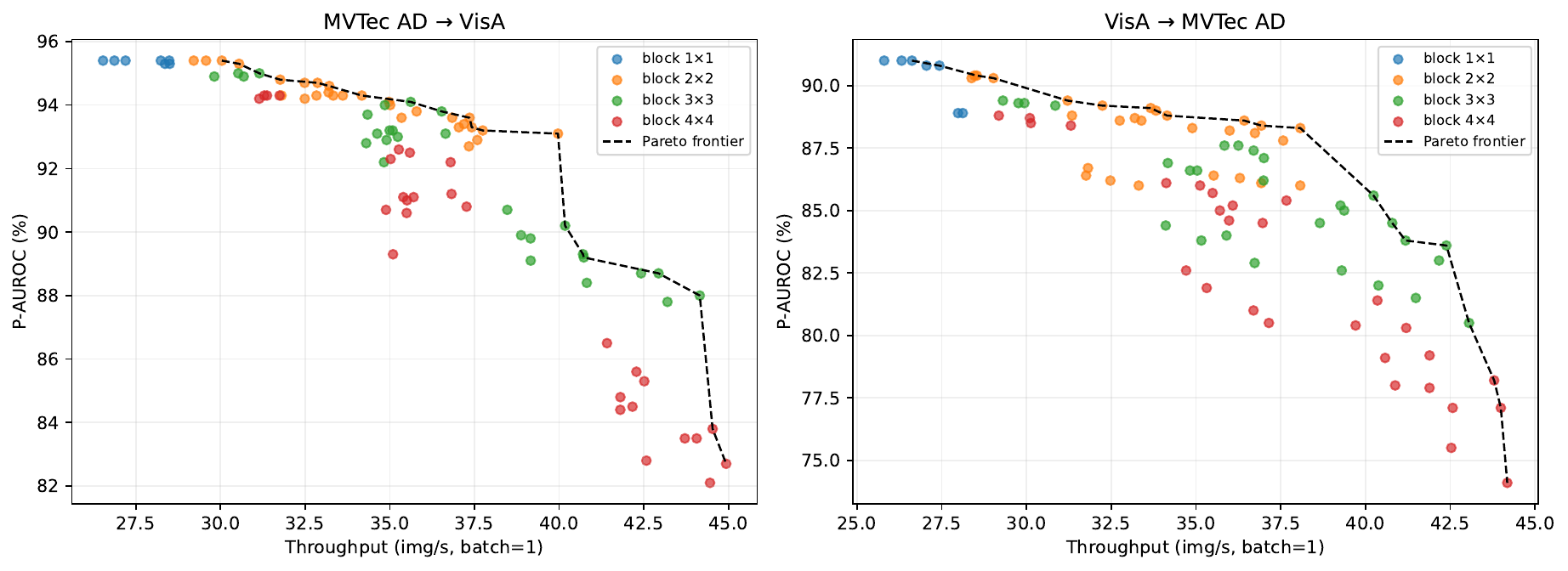}
    \caption{\textbf{Exploratory accuracy--throughput Pareto frontiers after the serial latency rerun.} I-AUROC (top) and P-AUROC (bottom) are plotted against batch-1 throughput for both transfer directions. Points are colored by local block geometry and jointly vary the pruning-layer pair, block size, and rescue ratio; the dashed curves mark the non-dominated envelopes.}
    \label{fig:pareto_frontiers}
\end{figure*}

\begin{figure*}[t]
    \centering
    \includegraphics[width=0.91\textwidth]{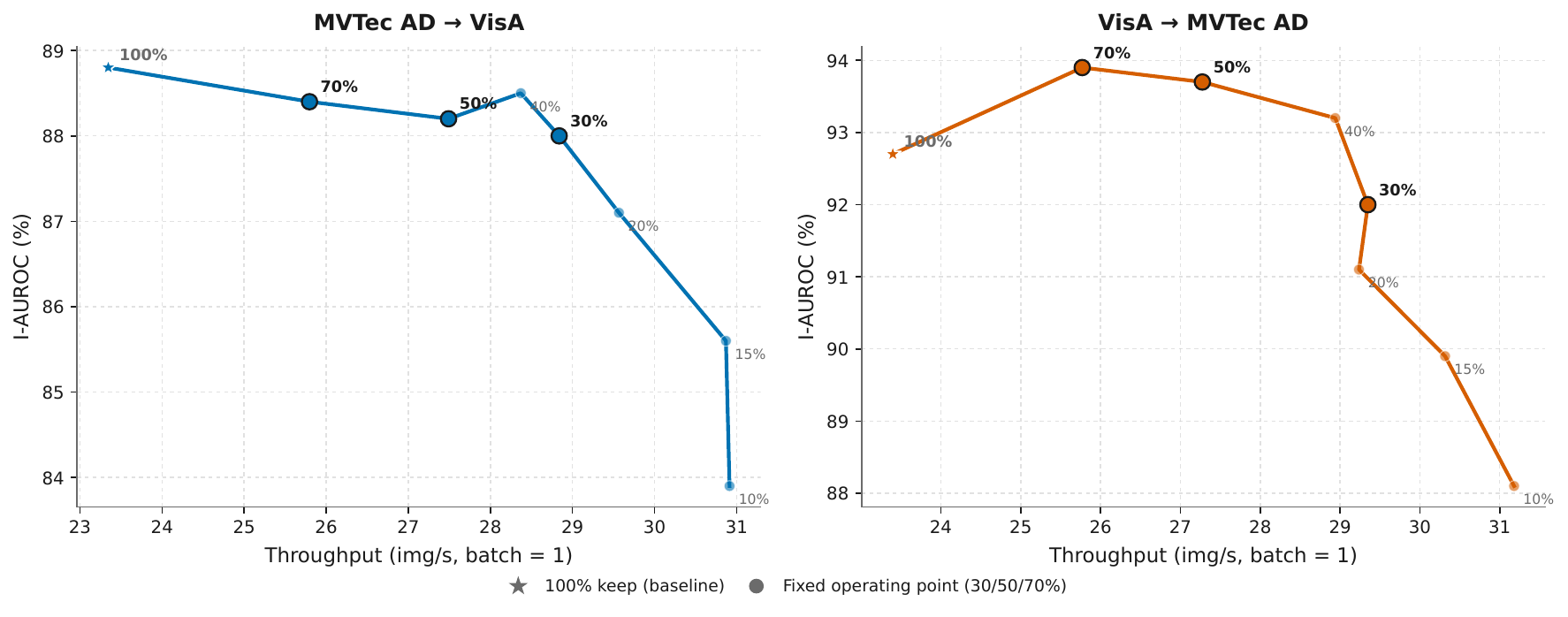}\par\vspace{0.4em}
    \includegraphics[width=0.91\textwidth]{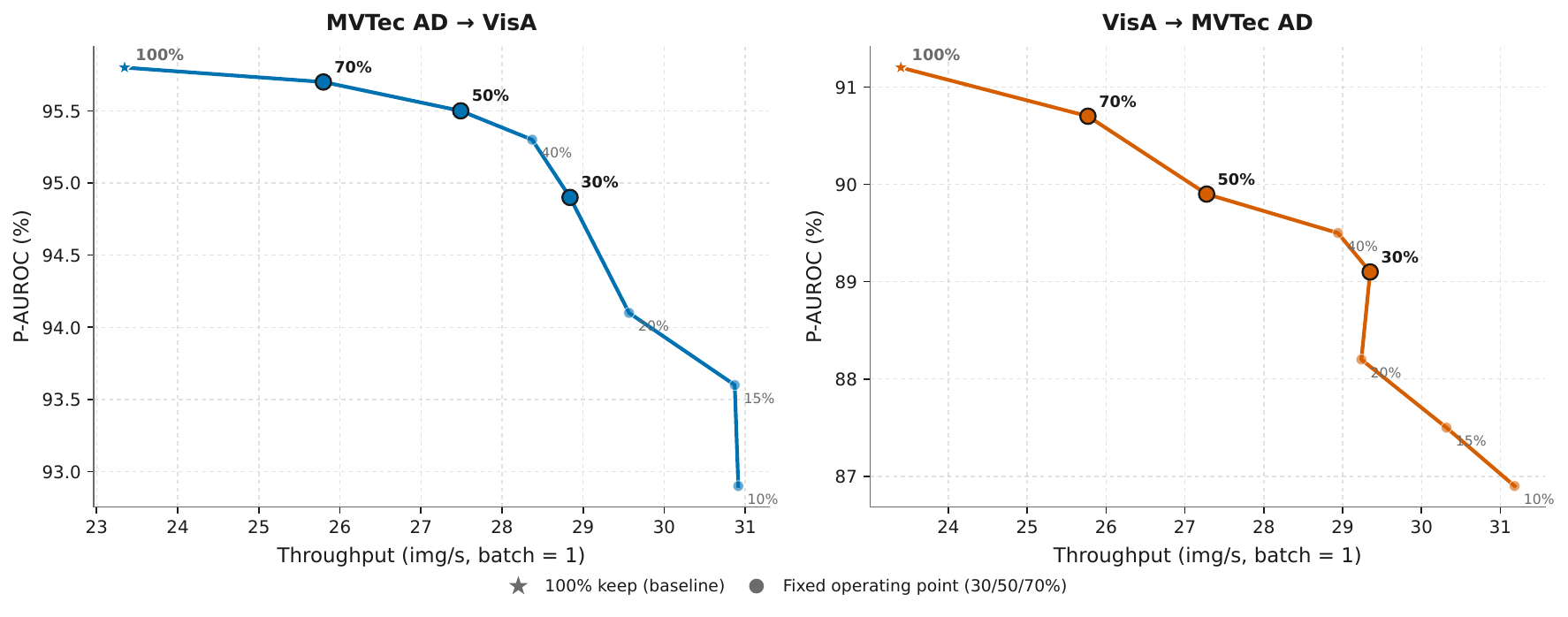}
    \caption{\textbf{Final-budget accuracy--throughput trajectories after the serial latency rerun.} I-AUROC (top) and P-AUROC (bottom) are shown for both transfer directions. Percentage annotations denote target final keep ratios; the solid lines connect the evaluated budget sequence and the dashed lines trace its Pareto envelope. The $100\%$ endpoint is the Stage-1 dense reference.}
    \label{fig:budget_trajectories}
\end{figure*}

Figures~\ref{fig:pareto_frontiers} and~\ref{fig:budget_trajectories} summarize two exploratory accuracy--throughput sweeps using full-test accuracy and the locked serial batch-1 latency rerun. The first sweep varies layer-pair, block geometry, and rescue ratio jointly, with colors encoding block geometry and dashed curves marking non-dominated envelopes. The second follows the final-budget trajectory, with annotations giving target final keep ratios. These plots summarize the attainable design space; because the sweeps vary multiple factors and the second includes the Stage-1 dense endpoint, component-level attribution relies on the budget-matched controls below.

Exploratory sweeps that vary one component often change the effective token budget or the surrounding pipeline at the same time, which confounds attribution:
\begin{itemize}
\item varying the block size also changed the realized final keep, granting smaller blocks a hidden budget advantage;
\item the original $100\%$-keep point used the Stage-1 baseline, which does not run the scorer/selector modules at all;
\item changing the rescue ratio changed the total token count, confounding score-based rescue with simply keeping more tokens.
\end{itemize}

\section{Component and Inference Ablations}

This section separates train-time objectives from inference-only decisions. Objective ablations retrain the selectors while keeping the frozen detector and nominal operating point fixed; recovery, image scoring, and evidence-budget controls reuse fixed physical routing wherever the experimental question permits.

\subsection{Stage-2 Objective Components}
\label{app:loss_ablation}

Table~\ref{tab:loss_components} isolates the train-time terms in Eq.~\eqref{eq:app_stage2_loss}. Each variant is retrained from the same frozen Stage-1 detector, data order, initialization protocol, and nominal prune70 target; only the named term is disabled. Hard L8 keep and the L12 keep bounds remain in force for all variants, so removing $\mathcal L_{\mathrm{flops}}$ can only push the realized final keep toward the hard upper bound rather than above it. Every metric in a row comes from the same natural-keep run; we do not mix budget-matched and unmatched evaluations in one row.

\begin{table}[t]
\centering
\setlength{\tabcolsep}{1.4pt}
\renewcommand{\arraystretch}{1.10}
\resizebox{\linewidth}{!}{%
\footnotesize
\begin{tabular}{@{}lcccccc@{}}
\toprule
\textbf{Variant} & \textbf{I-AUROC} & \textbf{P-AUROC} & \textbf{Keep (\%)} & \textbf{DTR (\%)$\uparrow$} & \textbf{CMR (\%)$\downarrow$} & \textbf{img/s $\uparrow$} \\
\midrule
\rowcolor{GroupRow}
\multicolumn{7}{@{}c@{}}{\textbf{MVTec AD$\rightarrow$VisA}} \\
\rowcolor{OursRow}
Full method & 86.1 & 94.3 & 14.3 & 56.12 & 0.42 & 41.1 \\
w/o $\mathcal L_{\mathrm{flops}}$ & 86.8 & 94.9 & 19.2 & 63.50 & 0.25 & 36.5 \\
w/o $\mathcal L_{\mathrm{preserve}}$ & 84.8 & 93.7 & 14.5 & 48.25 & 1.67 & 40.8 \\
w/o $\mathcal L_{\mathrm{mask}}$ & 85.2 & 94.0 & 14.8 & 51.30 & 1.08 & 40.5 \\
w/o $\mathcal L_{\mathrm{teacher}}$ & 83.4 & 94.4 & 14.6 & 40.39 & 6.92 & 40.9 \\
\midrule
\rowcolor{GroupRow}
\multicolumn{7}{@{}c@{}}{\textbf{VisA$\rightarrow$MVTec AD}} \\
\rowcolor{OursRow}
Full method & 90.5 & 88.6 & 17.8 & 39.87 & 0.95 & 38.7 \\
w/o $\mathcal L_{\mathrm{flops}}$ & 91.2 & 89.2 & 19.3 & 49.80 & 0.40 & 35.0 \\
w/o $\mathcal L_{\mathrm{preserve}}$ & 88.9 & 87.8 & 18.0 & 32.40 & 2.86 & 38.4 \\
w/o $\mathcal L_{\mathrm{mask}}$ & 89.4 & 88.1 & 18.2 & 35.60 & 1.91 & 38.1 \\
w/o $\mathcal L_{\mathrm{teacher}}$ & 85.6 & 87.9 & 18.1 & 34.60 & 4.29 & 38.3 \\
\bottomrule
\end{tabular}}
\caption{Stage-2 objective ablation. Full and w/o $\mathcal L_{\mathrm{teacher}}$ reuse the corresponding core-ablation results.}
\label{tab:loss_components}
\end{table}

The completed comparison is summarized along two axes. Removing $\mathcal L_{\mathrm{flops}}$ inflates final keep by $+4.9$/$+1.5$ percentage points toward the hard L12 upper bound ($\approx19.3\%$) and lowers throughput by $4.6$/$3.7$ img/s, showing that the soft efficiency term is what pulls the controller below the hard cap. The modest accuracy gains under this larger natural keep are therefore budget effects, not better routing. At nearly matched keeps, removing $\mathcal L_{\mathrm{preserve}}$ lowers DTR by $7.87$/$7.47$ points and raises CMR by $1.25$/$1.91$ points, while removing $\mathcal L_{\mathrm{mask}}$ lowers DTR by $4.82$/$4.27$ points and raises CMR by $0.66$/$0.96$ points. Together with the existing teacher ablation---the largest routing-quality drop---these results show that $\mathcal L_{\mathrm{flops}}$ enforces the intended budget, $\mathcal L_{\mathrm{preserve}}$ and $\mathcal L_{\mathrm{mask}}$ protect high-risk tokens, and dense-to-sparse distillation provides the strongest defect-preservation signal.

\subsection{Recovery and Image-Score Controls}
\label{app:recovery_mid_control}

Table~\ref{tab:recovery_mid_controls} separates two post-routing decisions using the same physical survivor responses. The recovery panel changes only how dropped grid locations receive a patch score; the sparse-mid panel keeps the full recovery rule fixed and changes only image-level aggregation. Consequently, recovery should affect P-AUROC but not I-AUROC, whereas sparse-mid scoring should affect I-AUROC but not P-AUROC, DTR, or CMR. This invariance is checked numerically rather than assumed.

\begin{table}[t]
\centering
\setlength{\tabcolsep}{1.4pt}
\renewcommand{\arraystretch}{1.10}
\resizebox{\linewidth}{!}{%
\footnotesize
\begin{tabular}{@{}lccccc@{}}
\toprule
\textbf{Variant}
& \begin{tabular}[c]{@{}c@{}}\textbf{M$\rightarrow$V}\\[-1pt]\textbf{I-AUROC}\end{tabular}
& \begin{tabular}[c]{@{}c@{}}\textbf{M$\rightarrow$V}\\[-1pt]\textbf{P-AUROC}\end{tabular}
& \begin{tabular}[c]{@{}c@{}}\textbf{V$\rightarrow$M}\\[-1pt]\textbf{I-AUROC}\end{tabular}
& \begin{tabular}[c]{@{}c@{}}\textbf{V$\rightarrow$M}\\[-1pt]\textbf{P-AUROC}\end{tabular}
& \textbf{img/s $\uparrow$} \\
\midrule
\rowcolor{GroupRow}
\multicolumn{6}{@{}c@{}}{\textbf{(a) Dense-map recovery}} \\
zero fill & 86.1 & 91.8 & 90.5 & 85.2 & 41.4 / 39.0 \\
global nearest owner & 86.1 & 93.5 & 90.5 & 87.4 & 41.0 / 38.6 \\
\rowcolor{OursRow}
same-block-first owner (full) & 86.1 & 94.3 & 90.5 & 88.6 & 41.1 / 38.7 \\
\midrule
\rowcolor{GroupRow}
\multicolumn{6}{@{}c@{}}{\textbf{(b) Image-level scoring}} \\
survivor base only & 85.4 & 94.3 & 89.8 & 88.6 & 41.3 / 38.9 \\
fixed $0.5$ base--mid blend & 85.7 & 94.3 & 90.1 & 88.6 & 41.2 / 38.8 \\
\rowcolor{OursRow}
calibrated sparse-mid gate (full) & 86.1 & 94.3 & 90.5 & 88.6 & 41.1 / 38.7 \\
recovered-map TopKMean & 87.2 & 94.3 & 91.4 & 88.6 & 40.9 / 38.5 \\
\bottomrule
\end{tabular}}
\caption{Inference-only controls for dense recovery and image-level scoring. M$\rightarrow$V and V$\rightarrow$M denote the two transfer directions. Speed follows the same direction order. Physical routing is fixed, so DTR and CMR are omitted.}
\label{tab:recovery_mid_controls}
\end{table}

Relative to zero fill, global nearest-survivor recovery raises P-AUROC by $+2.5$/$+3.4$ points, while the corresponding I-AUROC differences remain $0.0$ because physical routing is fixed. For image scoring, the calibrated sparse-mid gate raises I-AUROC by $+0.7$/$+0.7$ points over survivor-only scoring at an overhead of only $-0.2$/$-0.2$ img/s. Scoring the recovered map raises I-AUROC by $+1.1$/$+0.9$ over the full method, illustrating that duplicated owner values inflate image evidence; survivor-based aggregation is therefore retained as the primary image score.

\subsection{Evidence-Conditioned Adaptive Budget}
\label{app:evidence_budget}

\paragraph{Matched-budget controls.}
Table~\ref{tab:adaptive_controls} compares Adaptive with two equal-cost controls. Fixed-Mean assigns every image Adaptive's mean budget; Shuffled-Adaptive preserves the exact budget multiset but permutes it across images. Adaptive is best on I-AUROC, P-AUROC, DTR, and CMR in both directions, showing that the gain comes from matching computation to the current image rather than from the marginal budget distribution alone.

\begin{figure}[t]
    \centering
    \includegraphics[width=\linewidth]{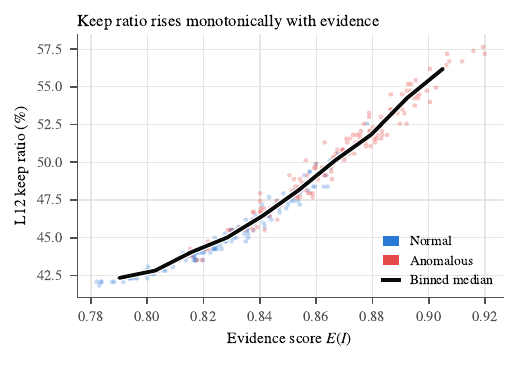}
    \caption{\textbf{Image-adaptive budget behavior} on MVTec AD$\rightarrow$VisA (prune70). The realized L12 keep ratio rises smoothly with the image-level evidence score $E(I)$ computed from normalized L12 token risks, confirming that the controller spends more computation on evidence-rich images.}
    \label{fig:budget}
\end{figure}

Figure~\ref{fig:budget} complements the class-conditional view in Figure~\ref{fig:budget_distribution} by plotting the image-level evidence score $E(I)$ against the realized L12 keep ratio. The keep ratio rises smoothly with $E(I)$, showing that the budget is not fixed per image but adapts to the normalized token-risk distribution and preserves more deep-layer tokens when localized high-risk evidence appears.

\subsection{Evidence-Statistic Decomposition}
\label{app:evidence_decomposition}

Table~\ref{tab:evidence_decomposition} tests whether the dispersion and upper-tail terms in Eq.~\eqref{eq:adaptive_budget} provide complementary information. For each single-statistic controller, only the scalar intercept is calibrated on source validation so that its average L12 keep matches the joint controller; the selector checkpoint, keep bounds, temperature, exponent, and all target-test inputs remain fixed. This prevents a larger mean budget from masquerading as a better evidence statistic.

\begin{table}[t]
\centering
\setlength{\tabcolsep}{1.4pt}
\renewcommand{\arraystretch}{1.10}
\resizebox{\linewidth}{!}{%
\footnotesize
\begin{tabular}{@{}lcccccc@{}}
\toprule
\textbf{Evidence} & \textbf{I-AUROC} & \textbf{P-AUROC} & \textbf{Keep$_{12}$ (\%)} & \textbf{Keep$_f$ (\%)} & \textbf{DTR (\%)$\uparrow$} & \textbf{CMR (\%)$\downarrow$} \\
\midrule
\rowcolor{GroupRow}
\multicolumn{7}{@{}c@{}}{\textbf{MVTec AD$\rightarrow$VisA}} \\
$E_{\mathrm{var}}$ only & 84.8 & 93.5 & 47.0 & 14.3 & 51.20 & 1.25 \\
$E_{\mathrm{tail}}$ only & 85.2 & 93.8 & 47.0 & 14.3 & 53.40 & 0.83 \\
\rowcolor{OursRow}
$\frac12E_{\mathrm{var}}+\frac12E_{\mathrm{tail}}$ & 86.1 & 94.3 & 47.0 & 14.3 & 56.12 & 0.42 \\
\midrule
\rowcolor{GroupRow}
\multicolumn{7}{@{}c@{}}{\textbf{VisA$\rightarrow$MVTec AD}} \\
$E_{\mathrm{var}}$ only & 89.2 & 87.4 & 59.0 & 17.8 & 34.50 & 2.07 \\
$E_{\mathrm{tail}}$ only & 89.6 & 87.9 & 59.0 & 17.8 & 36.80 & 1.43 \\
\rowcolor{OursRow}
$\frac12E_{\mathrm{var}}+\frac12E_{\mathrm{tail}}$ & 90.5 & 88.6 & 59.0 & 17.8 & 39.87 & 0.95 \\
\bottomrule
\end{tabular}}
\caption{Evidence-statistic decomposition at a matched average L12 budget.}
\label{tab:evidence_decomposition}
\end{table}

At a matched mean L12 keep, the joint statistic raises I/P-AUROC by $+1.3$/$+0.8$ over $E_{\mathrm{var}}$ alone and by $+0.9$/$+0.5$ over $E_{\mathrm{tail}}$ alone on MVTec AD$\rightarrow$VisA, and by $+1.3$/$+1.2$ and $+0.9$/$+0.7$ on VisA$\rightarrow$MVTec AD. The corresponding CMR reductions are $0.83$/$1.12$ and $0.41$/$0.48$. These results show that dispersion and tail strength are complementary across both transfer directions: neither single statistic matches the joint controller on localization or complete-miss control, so both terms are retained.

\section{Robustness and Cross-Domain Transfer}
\label{app:robustness}

The final empirical section checks whether the principal component conclusions persist across Stage-2 optimization seeds and under transfer from industrial training data to a seven-dataset medical suite.

\subsection{Stage-2 Seed Stability}

Table~\ref{tab:seed_stability} repeats Stage-2 training with three fixed seeds while keeping Stage~1, the data split, checkpoint rule, and evaluation code unchanged. We report mean$\pm$standard deviation across seeds. CMR is computed from per-image complete-miss indicators; patch tokens are not treated as independent samples.

\begin{table}[t]
\centering
\setlength{\tabcolsep}{1.6pt}
\renewcommand{\arraystretch}{1.12}
\resizebox{\linewidth}{!}{%
\footnotesize
\begin{tabular}{@{}lccccc@{}}
\toprule
\textbf{Variant} & \textbf{I-AUROC} & \textbf{P-AUROC} & \textbf{Keep (\%)} & \textbf{DTR (\%)$\uparrow$} & \textbf{CMR (\%)$\downarrow$} \\
\midrule
\rowcolor{GroupRow}
\multicolumn{6}{@{}c@{}}{\textbf{MVTec AD$\rightarrow$VisA}} \\
\rowcolor{OursRow}
Full & 86.1$\pm$0.15 & 94.3$\pm$0.10 & 14.3$\pm$0.2 & 55.9$\pm$0.8 & 0.44$\pm$0.08 \\
w/o teacher & 83.3$\pm$0.35 & 94.3$\pm$0.20 & 14.6$\pm$0.4 & 40.1$\pm$1.5 & 7.11$\pm$0.55 \\
\midrule
\rowcolor{GroupRow}
\multicolumn{6}{@{}c@{}}{\textbf{VisA$\rightarrow$MVTec AD}} \\
\rowcolor{OursRow}
Full & 90.4$\pm$0.22 & 88.5$\pm$0.15 & 17.8$\pm$0.3 & 39.6$\pm$0.9 & 0.98$\pm$0.12 \\
w/o teacher & 85.5$\pm$0.48 & 87.8$\pm$0.28 & 18.2$\pm$0.5 & 34.2$\pm$1.8 & 4.45$\pm$0.70 \\
\bottomrule
\end{tabular}}
\caption{Stage-2 stability over three training seeds. Every entry reports mean$\pm$std; CMR is accompanied by the pooled complete-miss count in the experiment record.}
\label{tab:seed_stability}
\end{table}

Across seeds, the full method has small variability in both transfer directions. The full-versus-no-teacher gaps in I-AUROC, DTR, and CMR are substantially larger than the corresponding seed-to-seed standard deviations, supporting the robustness of the self-distillation conclusion.

\subsection{Medical-Suite Component Transfer}

To test whether the component conclusions survive a larger domain shift, Table~\ref{tab:medical_component_transfer} evaluates the VisA-trained prune70 checkpoints and inference controls on the seven medical datasets without retraining or medical-set calibration. Besides the suite macro-average, we report the mean P-AUROC on Endo and Kvasir, where the main results identify large diffuse lesions as the most challenging case. DTR and CMR are averaged only over datasets that provide compatible pixel masks.

\begin{table}[t]
\centering
\setlength{\tabcolsep}{1.5pt}
\renewcommand{\arraystretch}{1.10}
\resizebox{\linewidth}{!}{%
\footnotesize
\begin{tabular}{@{}lccccc@{}}
\toprule
\textbf{Variant}
& \begin{tabular}[c]{@{}c@{}}\textbf{Avg.}\\[-1pt]\textbf{I-AUROC}\end{tabular}
& \begin{tabular}[c]{@{}c@{}}\textbf{Avg.}\\[-1pt]\textbf{P-AUROC}\end{tabular}
& \begin{tabular}[c]{@{}c@{}}\textbf{E/K}\\[-1pt]\textbf{P-AUROC}\end{tabular}
& \textbf{DTR (\%)$\uparrow$}
& \textbf{CMR (\%)$\downarrow$} \\
\midrule
\rowcolor{OursRow}
Full method & 79.0 & 82.5 & 77.9 & 42.30 & 1.80 \\
L8 global top-$k$ & 76.8 & 74.2 & 68.5 & 48.50 & 8.60 \\
L12 fixed budget & 77.5 & 80.8 & 75.1 & 36.20 & 3.40 \\
w/o self-distillation & 76.2 & 81.1 & 75.8 & 33.50 & 5.20 \\
\bottomrule
\end{tabular}}
\caption{Transfer of key component ablations to the seven-dataset medical suite. Avg.\ I-AUROC and Avg.\ P-AUROC are suite averages; E/K P-AUROC denotes mean P-AUROC over Endo and Kvasir. All choices are fixed before medical evaluation.}
\label{tab:medical_component_transfer}
\end{table}

Across the medical suite, replacing local coverage with global top-$k$ changes average P-AUROC and CMR by $-8.3$/$+6.8$, a fixed L12 budget changes them by $-1.7$/$+1.6$, and removing self-distillation changes them by $-1.4$/$+3.4$. On the diffuse-lesion subset (Endo/Kvasir), the corresponding P-AUROC changes are $-9.4$, $-2.8$, and $-2.1$. These results show that local coverage remains the dominant transferable component under medical domain shift, while adaptive budgeting and self-distillation still help with smaller absolute margins; the main medical-specific caveat is that global top-$k$ is especially harmful for large diffuse lesions.

\section{Extended Related Work}

This section situates KeepAD along two axes: the increasing adaptation complexity of zero-shot anomaly detectors and the evolution of token compression from classification to dense and vision--language settings.

\subsection{Evolution of Zero-Shot Anomaly Detection}
\label{app:rw_zsad}

The progression below is not a strictly monotonic increase in runtime: some
methods remove one source of cost while introducing another.  Rather, it shows
how accuracy improvements have increasingly relied on specialized adaptation
components and additional token interactions, without reducing the dense patch
sequence processed by the vision backbone.

\paragraph{From prompt engineering to multi-component CLIP adaptation.}
WinCLIP keeps CLIP frozen and combines large handcrafted normal/abnormal prompt
ensembles with windows at multiple scales~\cite{jeong2023winclip}.  It avoids
training an adapter, but localization requires repeated window-level feature
extraction and prompt comparisons.  APRIL-GAN instead trains linear projection
layers on auxiliary anomaly data to map intermediate image features into the
joint image--text space~\cite{chen2023aprilgan}.  CLIP-AD further combines
representative text-vector selection with a staged dual path, multi-level
features, architecture/feature surgery, and, in its SDP+ variant, learned linear
alignment layers~\cite{chen2023clipad}.

AnomalyCLIP learns object-agnostic normal and abnormal prompts under joint
image- and pixel-level supervision and inserts learnable prompt tokens into the
frozen text encoder~\cite{zhou2024anomalyclip}.  On the vision side, its
diagonally prominent attention map (DPAM) modifies deeper self-attention blocks:
the default value-to-value (V--V) path forms attention from value features
rather than the conventional query--key map, emphasizing diagonal/local
relations that are more useful for patch localization.  Intermediate visual
features are then aligned with the learned prompts.  This design reduces the
multiple image-window passes of WinCLIP, but adds prompt optimization, a
modified attention path, and multi-level image--text supervision.

AdaCLIP extends prompt adaptation to both CLIP branches~\cite{cao2024adaclip}.
Selected visual and textual transformer blocks concatenate $M$ hybrid prompt
tokens to their original sequences; each hybrid token is the sum of a shared
static prompt and an image-conditioned dynamic prompt, so each prompted stream
grows by $M$, not $2M$, tokens.  The complete pipeline additionally contains a
dynamic-prompt generator, patch projection, multi-layer anomaly-map fusion, and
cluster-based hybrid semantic fusion.  Bayes-PFL moves from point prompts to
prompt distributions~\cite{qu2025bayespfl}: prompt banks parameterize
image-specific and image-agnostic distributions, normalizing flows transform
them, Monte-Carlo samples produce diverse normal/abnormal prompts, and residual
cross-modal attention refines every sampled text pair against patch features.
Its coverage of prompt space therefore comes with stochastic flow inference,
repeated text embeddings, patch--text interactions, and sample ensembling.

\paragraph{Language-free adaptation.}
Removing language does not necessarily remove token-side adaptation.  UniADet
is a deliberately lightweight exception that replaces the text branch with
decoupled weights for different tasks and feature levels~\cite{uniadnet}.
VisualAD instead prepends two persistent learnable state tokens, one normal and
one anomalous, to the frozen ViT~\cite{visualAD}.  With a class token and $N$
patches, every backbone block therefore processes $N+3$ rather than $N+1$
tokens.  The two state tokens interact with all patches through self-attention;
layer-specific spatial-aware cross-attention (SCA) injects local evidence, a
self-alignment function (SAF) recalibrates patch features, and multi-layer
fusion forms the final scores.  Thus, VisualAD is language-free and
parameter-efficient relative to prompt-heavy CLIP adapters, yet it still adds
state-token interactions and auxiliary attention/alignment paths.  Across
these designs, the adaptation mechanism changes substantially, but the dense
patch grid remains intact throughout the ViT; this unchanged cost is the target
of KeepAD.

\subsection{Evolution of Token Compression}
\label{app:rw_pruning}

\paragraph{Classification-oriented reduction.}
DynamicViT, EViT, and A-ViT learn token selection, fusion, or token-wise halting; ATS and ToMe use attention sampling or similarity-based merging~\cite{rao2021dynamicvit,liang2022evit,yin2022avit,fayyaz2022ats,bolya2023tome}. Patch Slimming, Evo-ViT, SPViT, and Zero-TPrune add top-down supervision, fast/slow token paths, package tokens, latency objectives, or training-free graph centrality~\cite{tang2022patchslimming,xu2022evovit,kong2022spvit,wang2024zerotprune}. These methods establish effective adaptive computation, but their routing signals are optimized primarily for preserving a global classification representation. That objective is poorly aligned with anomaly detection, where a single low-salience patch may be the only positive evidence.

\paragraph{Dense prediction.}
SparseViT reduces selected window computation while retaining spatial resolution; DToP exits easy segmentation tokens through auxiliary heads; and SViT caches pruned tokens so later layers can reactivate them under an image-dependent budget~\cite{chen2023sparsevit,tang2023dtop,liu2024revisitingpruning}. These designs are closer to KeepAD because they preserve or reconstruct dense outputs. Their routing cues, however, rely on known semantic categories or generic object context. KeepAD instead targets the earlier and harder regime in which a rare defect is locally weak, semantically unresolved, and costly to miss completely.

\paragraph{Vision--language and anomaly-specific compression.}
Recent VLM pruning ranks or clusters visual tokens using attention, redundancy, diversity, instruction relevance, or cross-modal agreement; learned variants add differentiable sparsification and input-dependent budgets~\cite{chen2024fastv,zou2025holov,zhang2025sparsevlm,wen2025dart,alvar2025divprune,yang2025visionzip,shang2025prumerge,yu2026visiontrim,fang2026prunesid,zhang2026htcvlm,xu2026visiondrop,yu2026instructionclustering,huang2025dynamicllava,ye2025atpllava,takezoe2026learnpruner}. Such criteria suit classification, segmentation, or generation, but do not explicitly control the complete-miss risk of rare abnormal patches; merging can also destroy the spatial identity required by a pixel anomaly map. VMAD is the closest anomaly-specific precedent, but compresses multi-level features inside an MLLM projector after the dense vision encoder has already processed the full grid~\cite{deng2026vmad}. KeepAD instead performs physical, progressive pruning inside the ViT and couples it with defect-aware scoring, a local coverage floor, survivor tracking, and sparse-to-dense recovery.

\section{Visualizations}
\begin{figure*}[t]
    \centering
    \includegraphics[width=\textwidth]{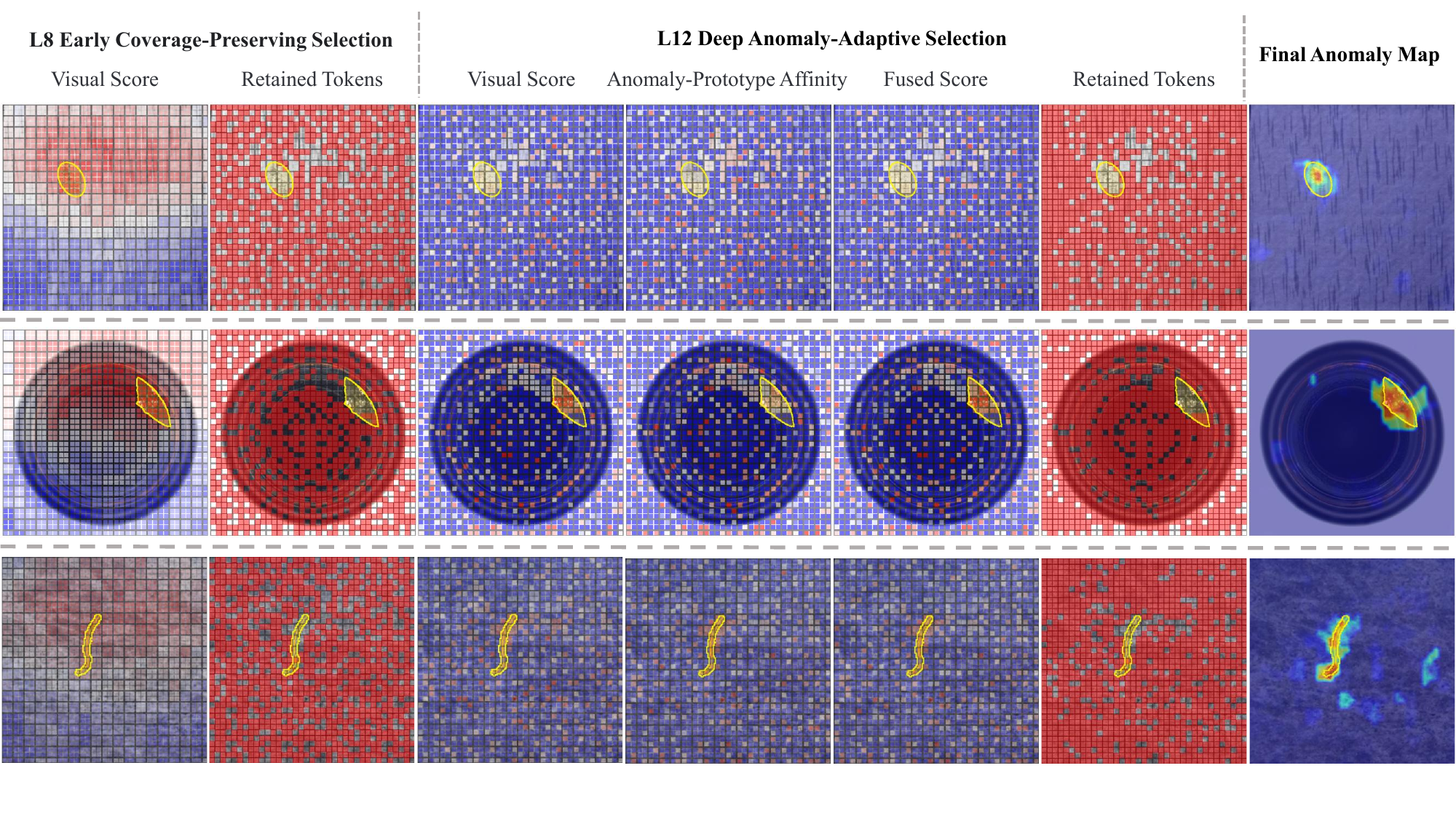}
\end{figure*}

\begin{figure*}[t]
    \centering
    \includegraphics[width=\textwidth]{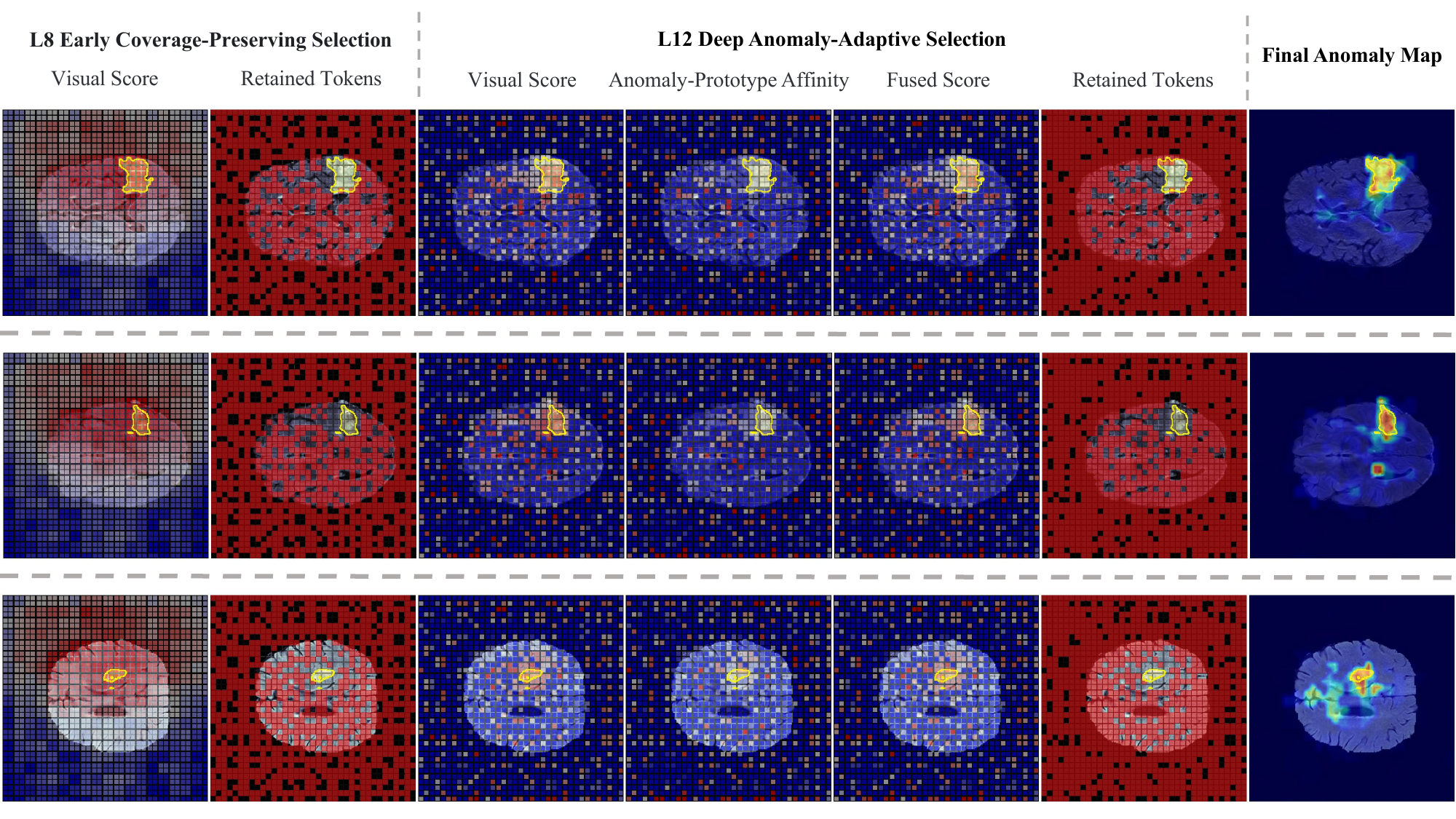}
\end{figure*}

\begin{figure*}[t]
    \centering
    \includegraphics[width=\textwidth]{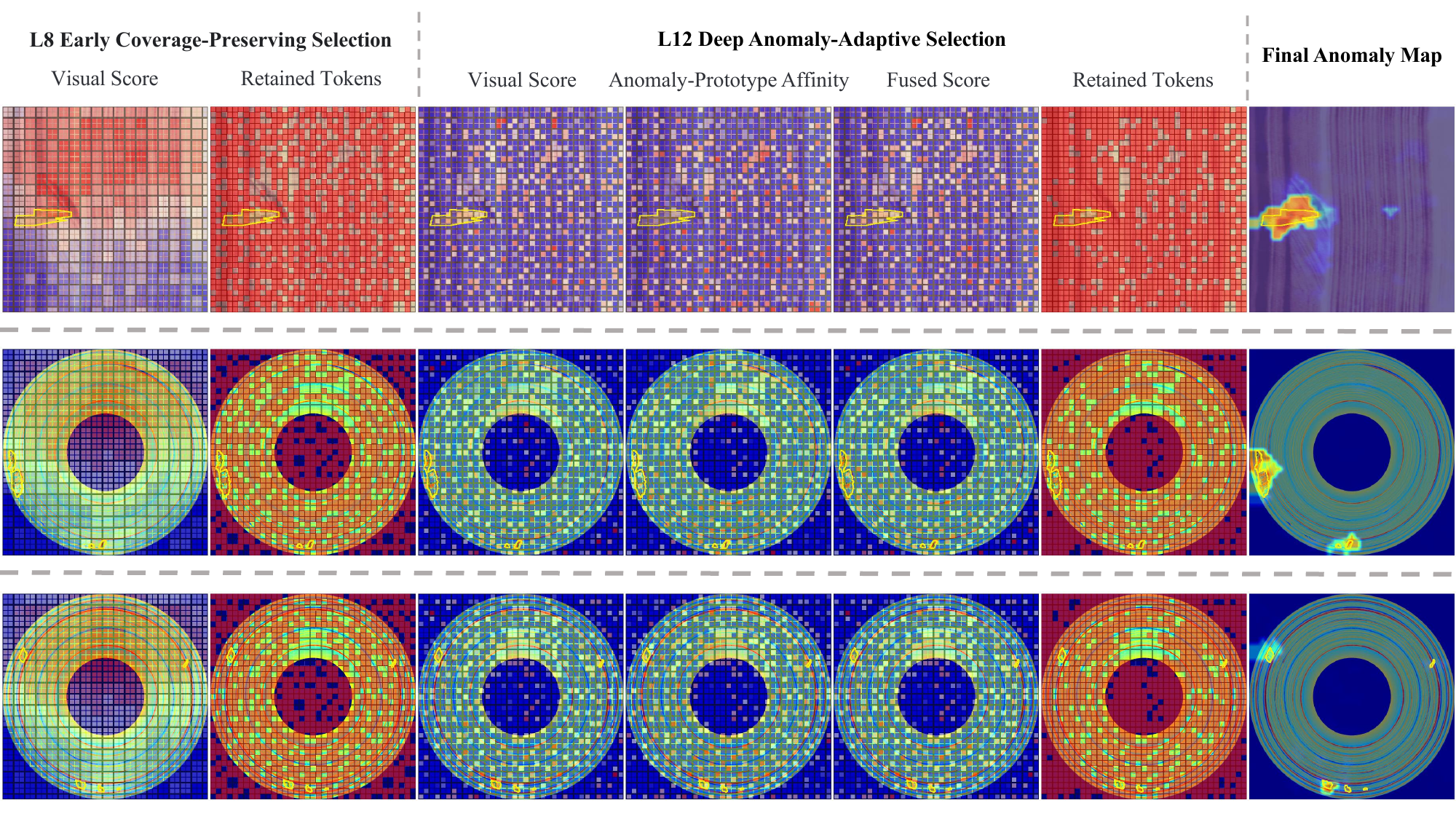}
\end{figure*}

\begin{figure*}[t]
    \centering
    \includegraphics[width=\textwidth]{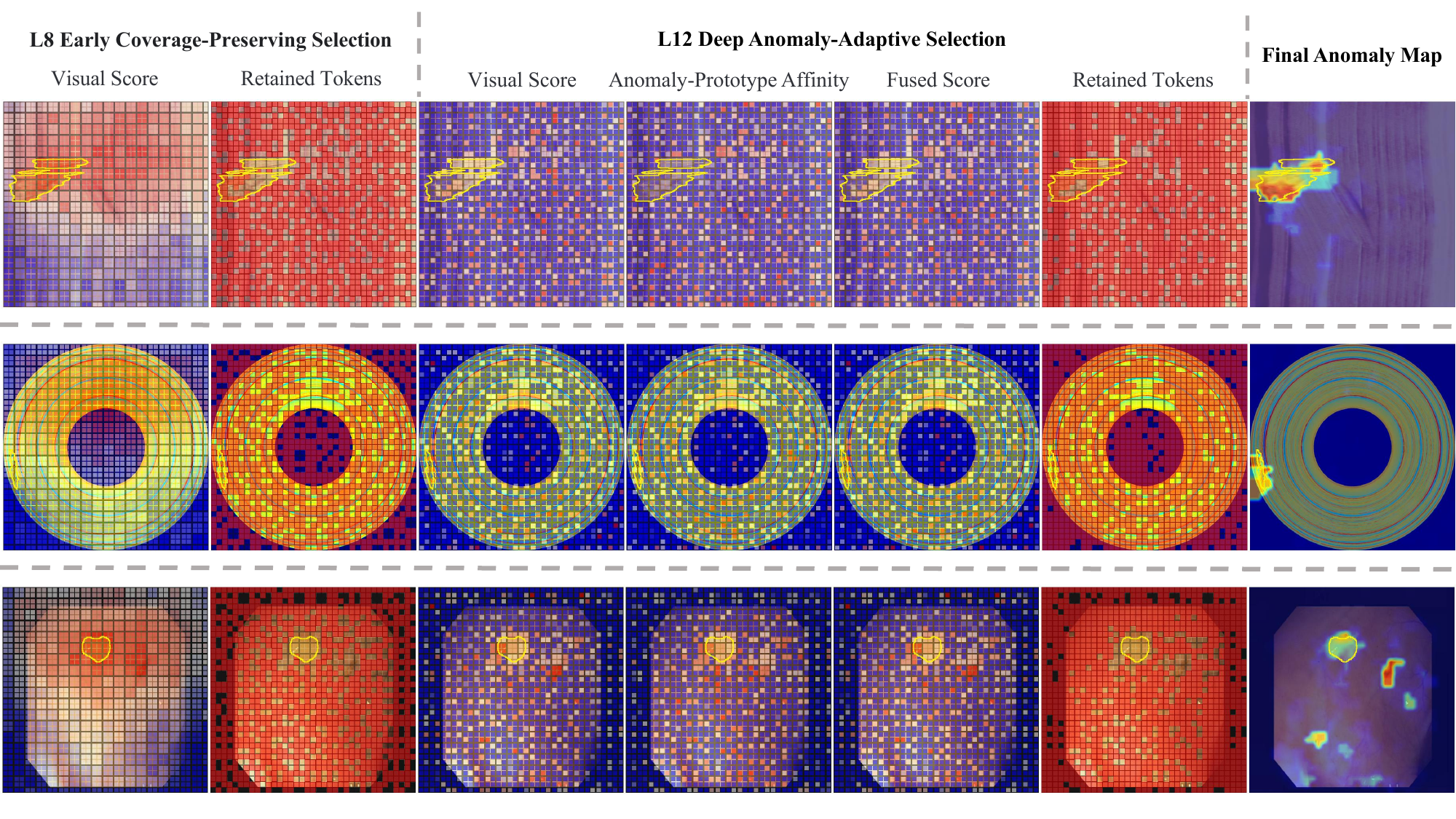}
\end{figure*}

\begin{figure*}[t]
    \centering
    \includegraphics[width=\textwidth]{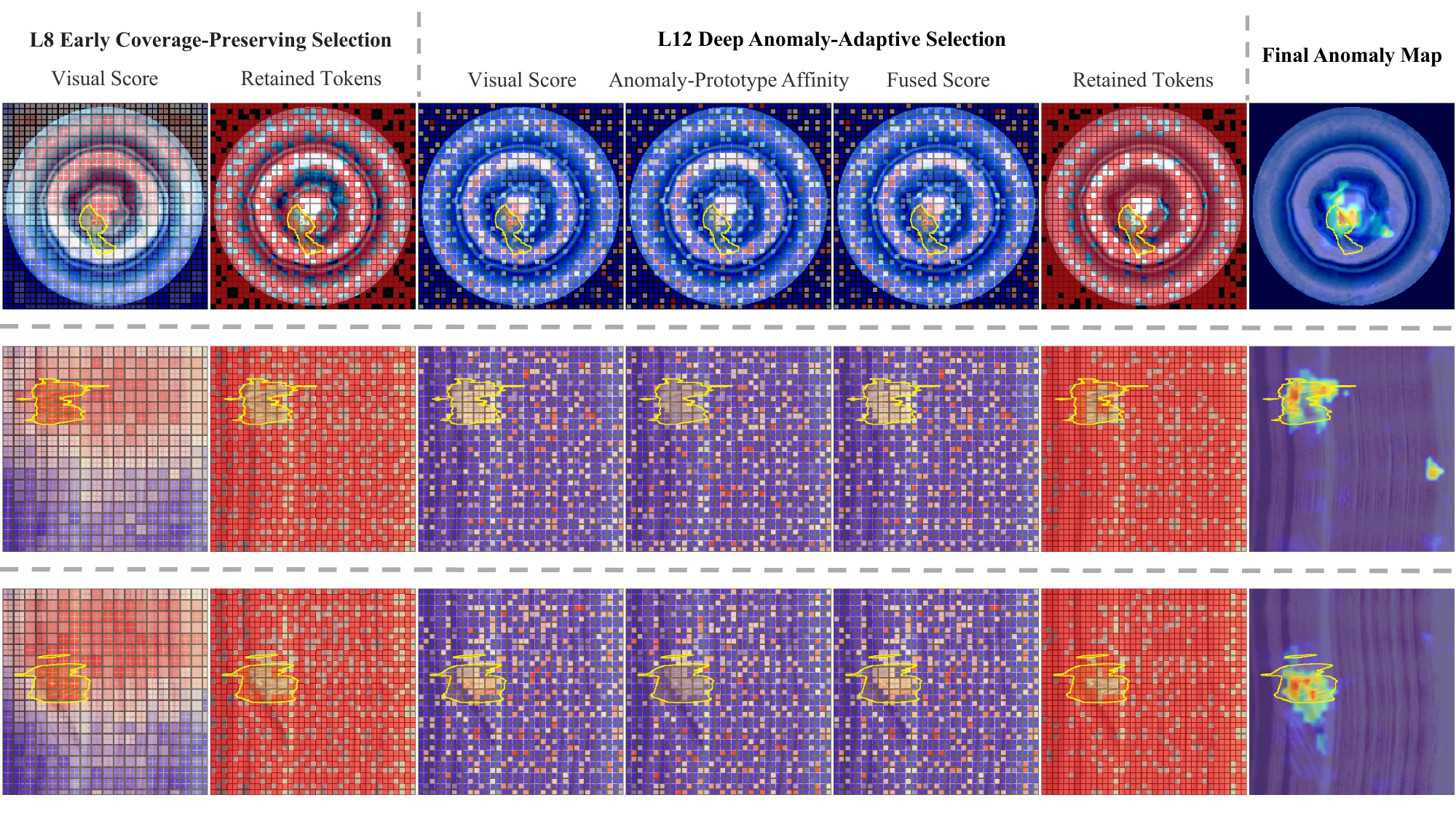}
\end{figure*}

\begin{figure*}[t]
    \centering
    \includegraphics[width=\textwidth]{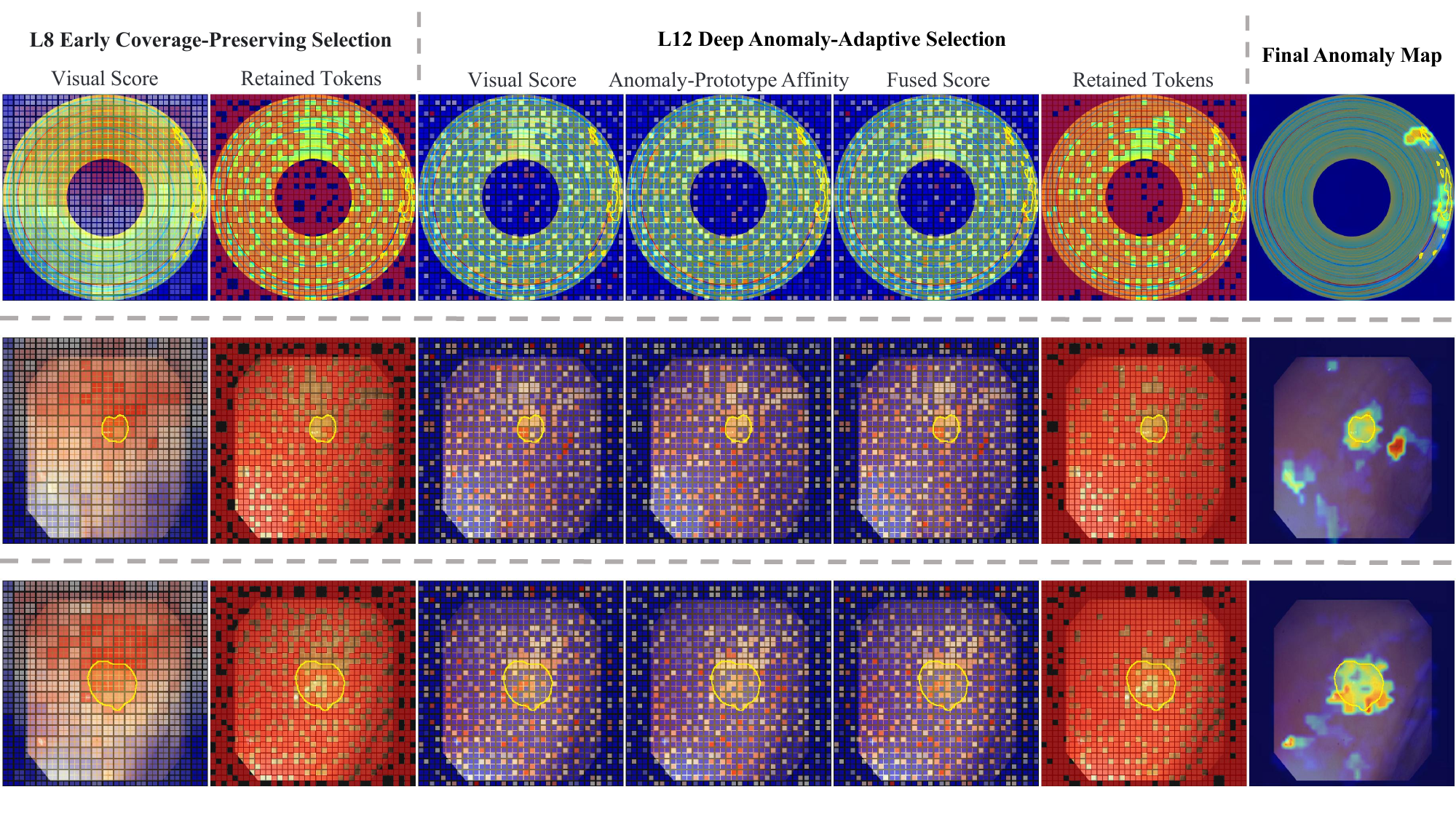}
\end{figure*}

\begin{figure*}[t]
    \centering
    \includegraphics[width=\textwidth]{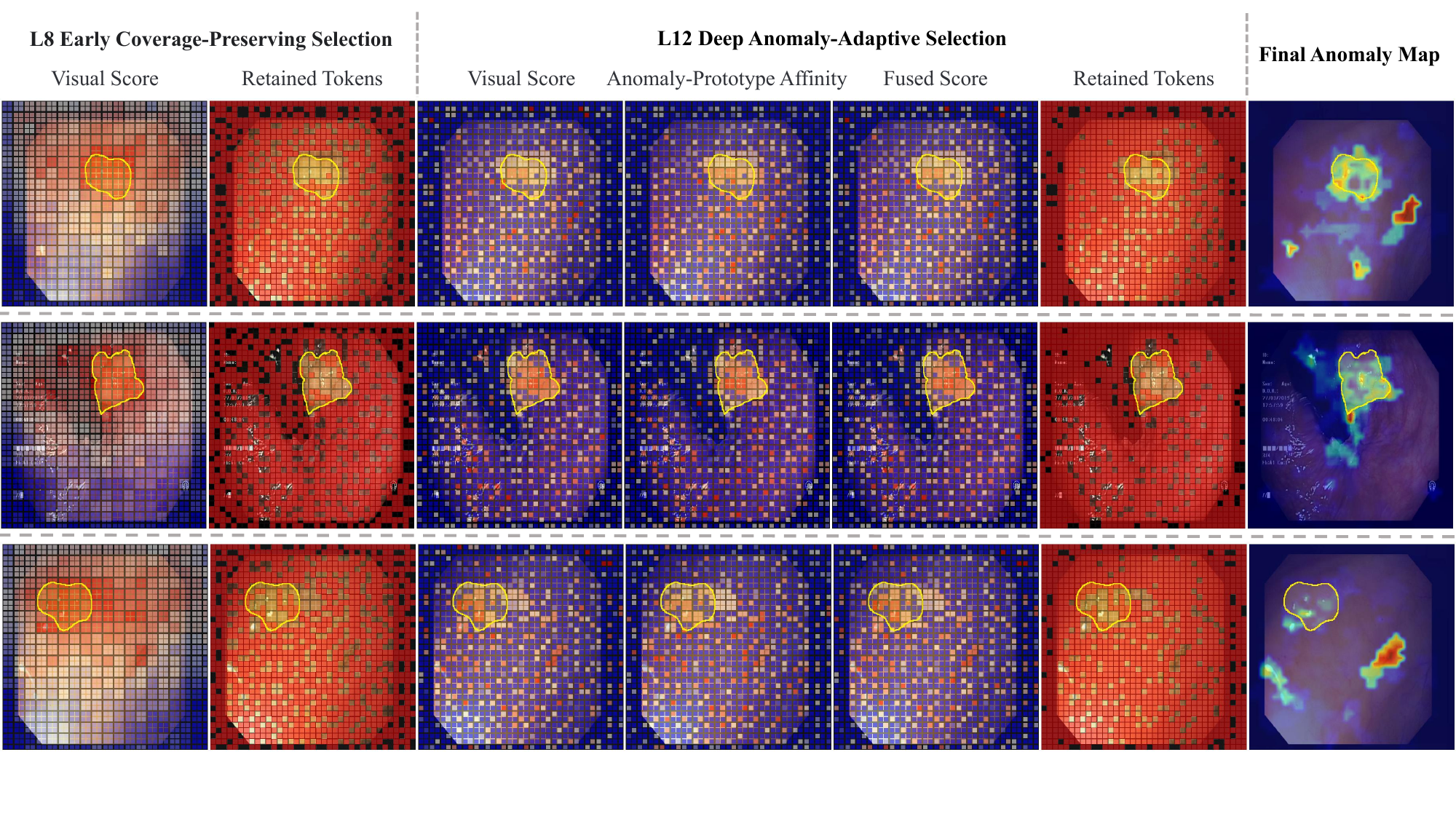}
\end{figure*}

\begin{figure*}[t]
    \centering
    \includegraphics[width=\textwidth]{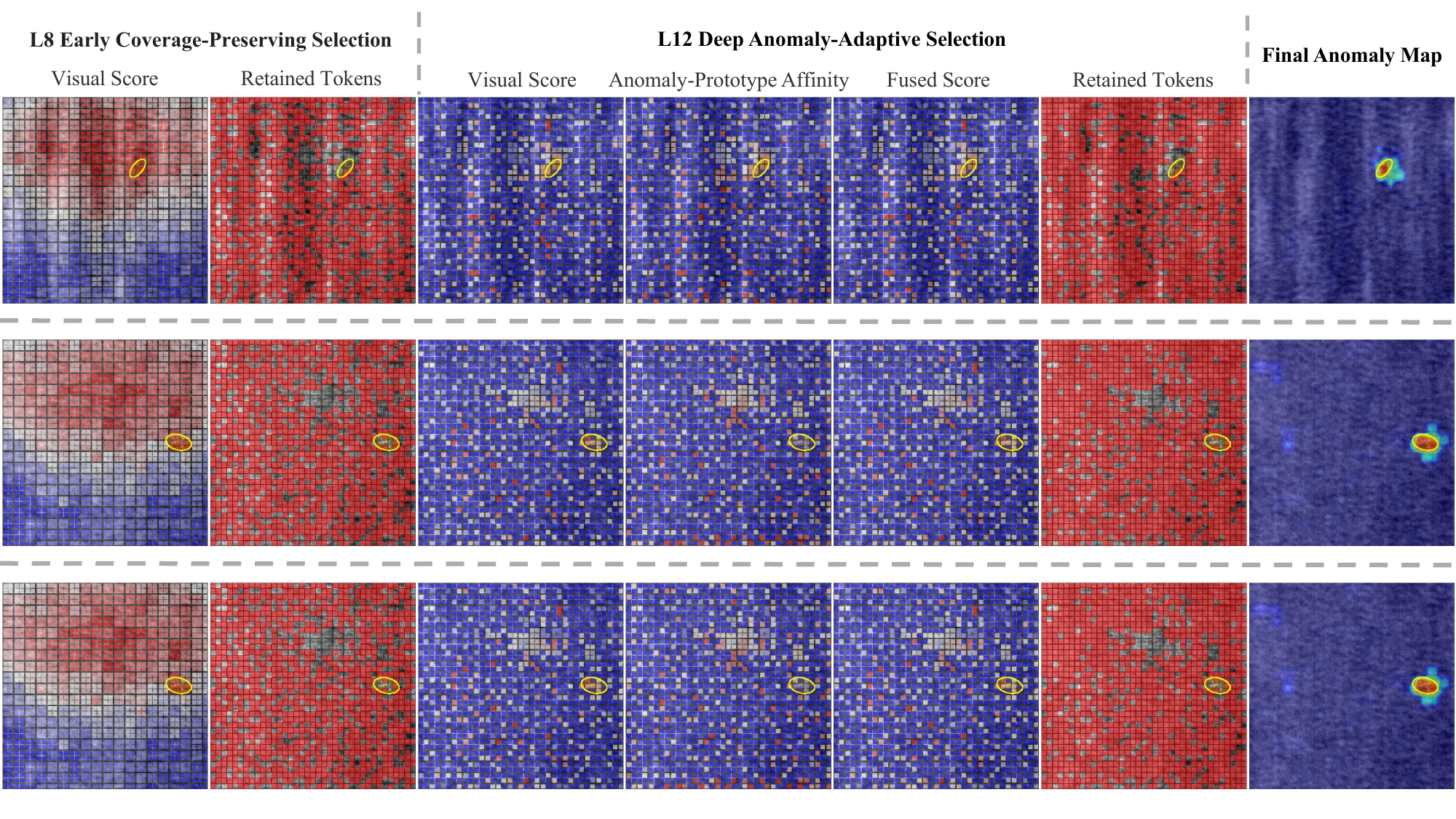}
\end{figure*}

\begin{figure*}[t]
    \centering
    \includegraphics[width=\textwidth]{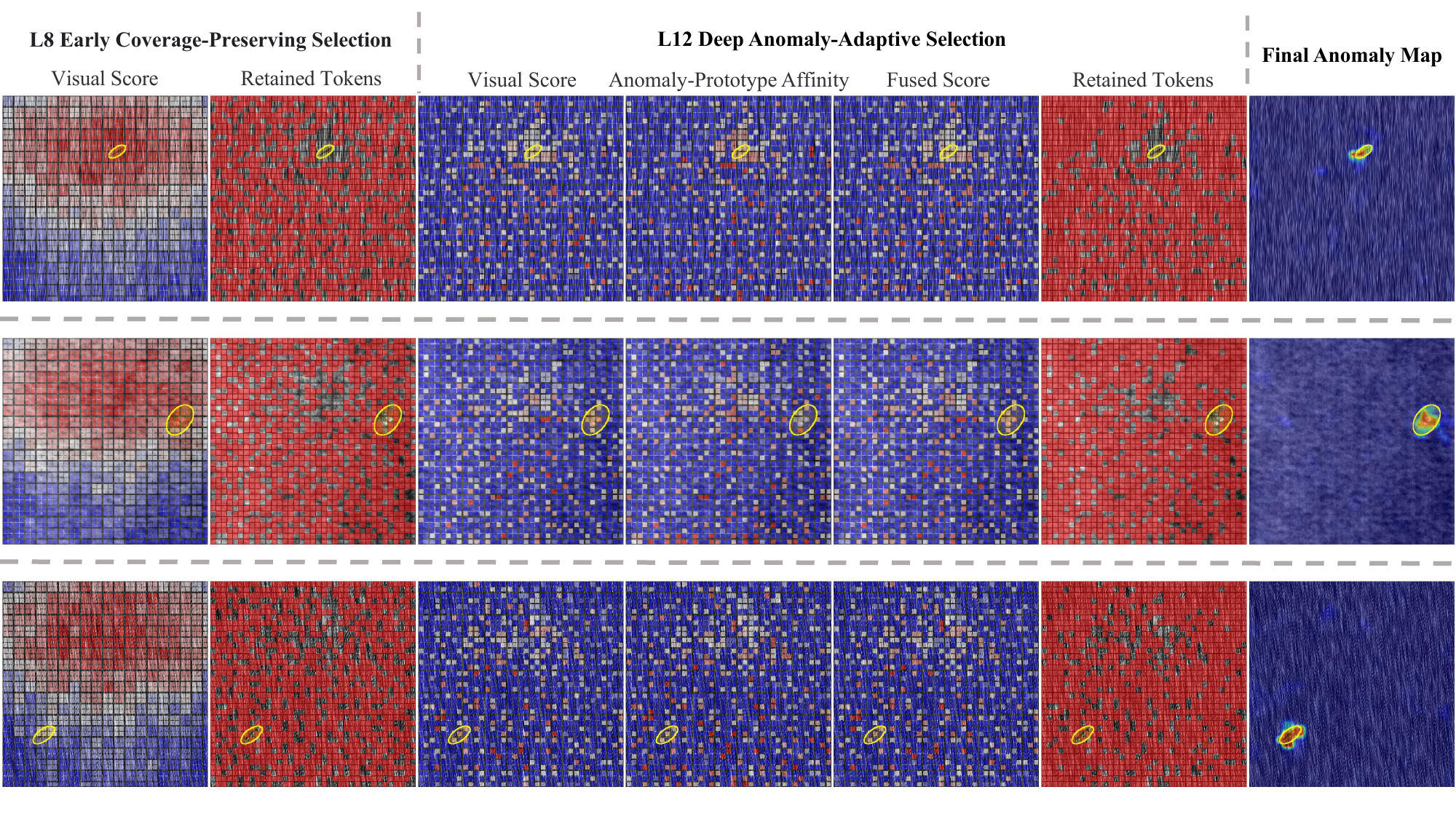}
\end{figure*}

\begin{figure*}[t]
    \centering
    \includegraphics[width=\textwidth]{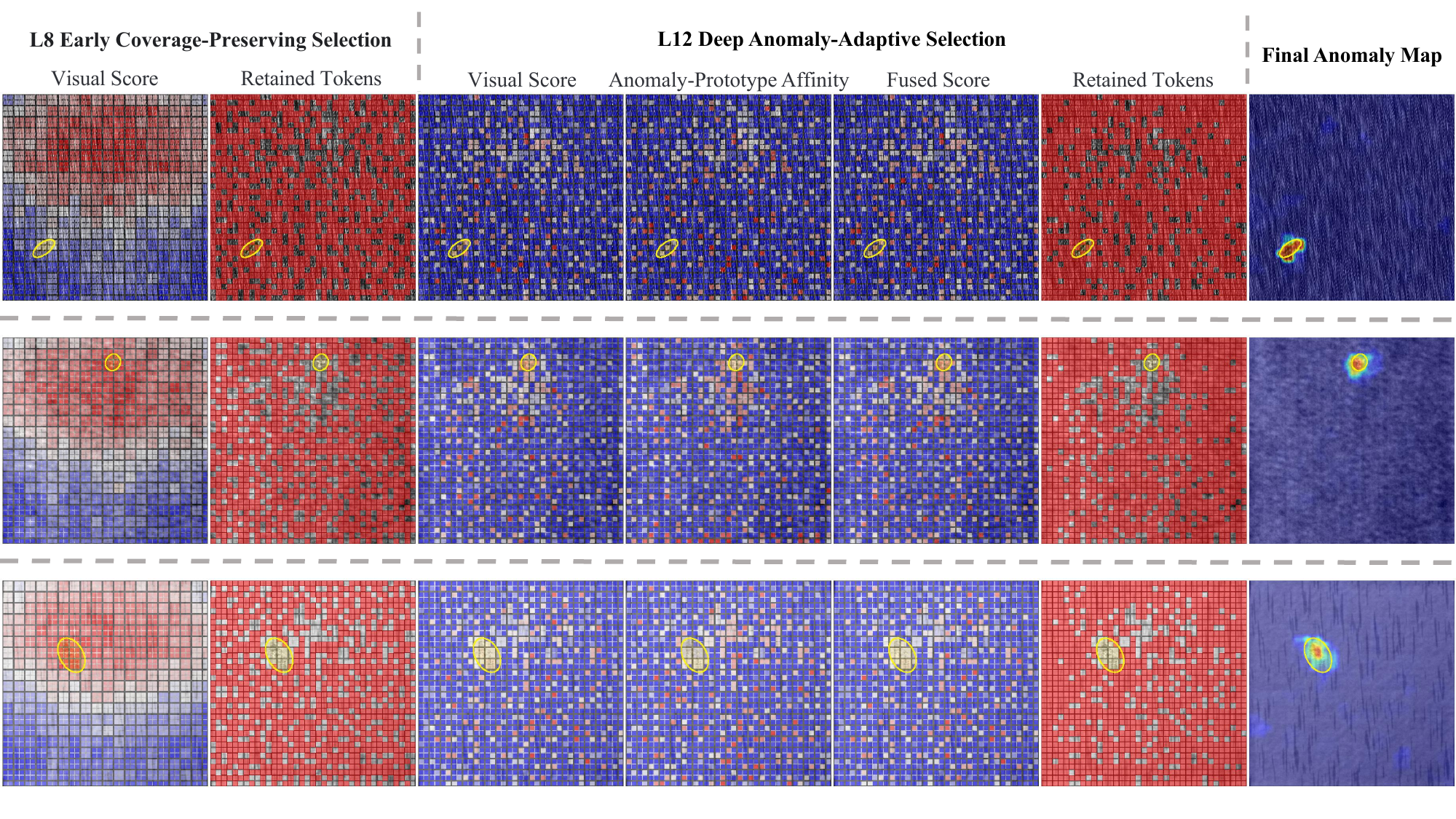}
\end{figure*}

\begin{figure*}[t]
    \centering
    \includegraphics[width=\textwidth]{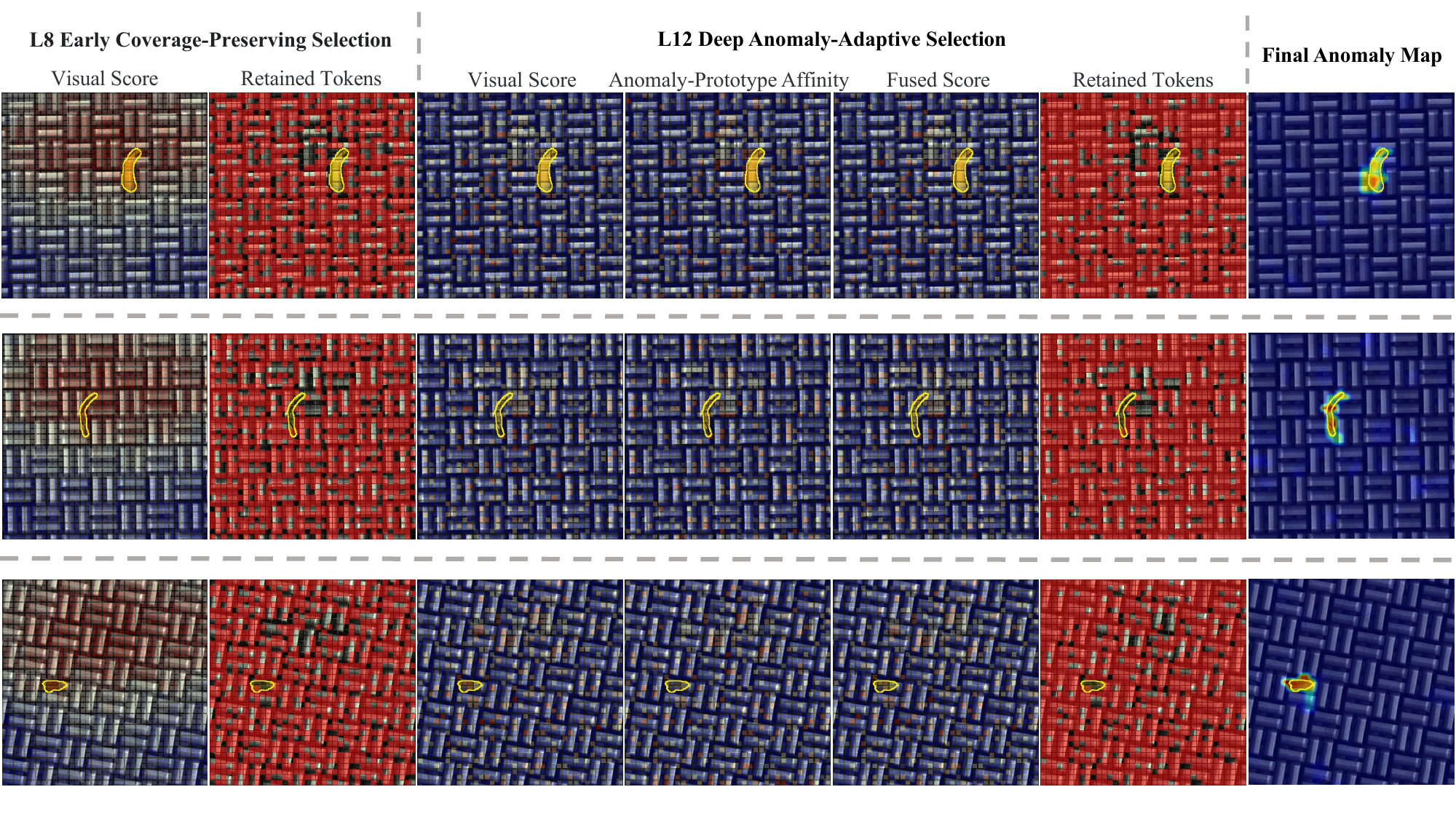}
\end{figure*}

\begin{figure*}[t]
    \centering
    \includegraphics[width=\textwidth]{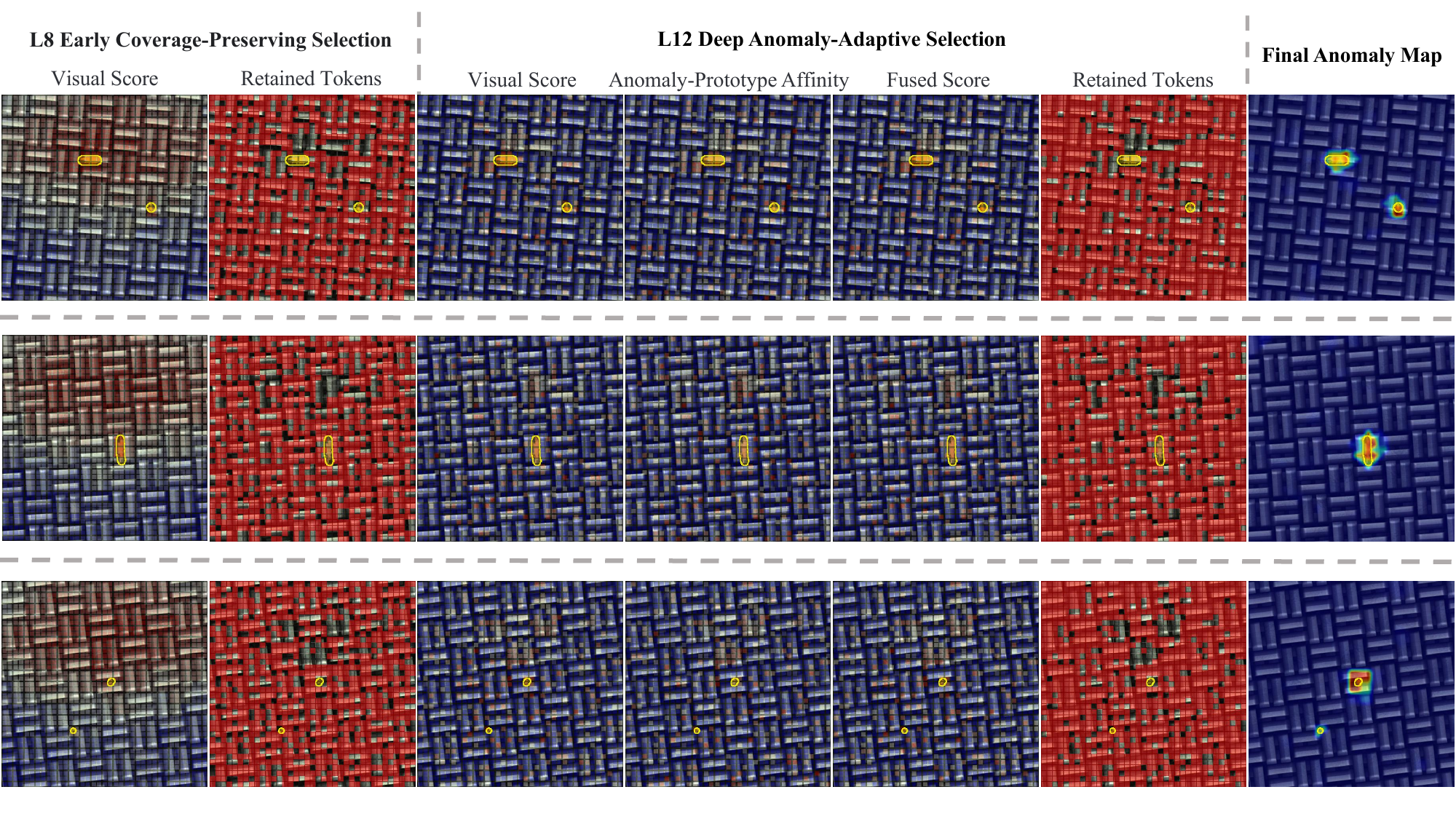}
\end{figure*}

\begin{figure*}[t]
    \centering
    \includegraphics[width=\textwidth]{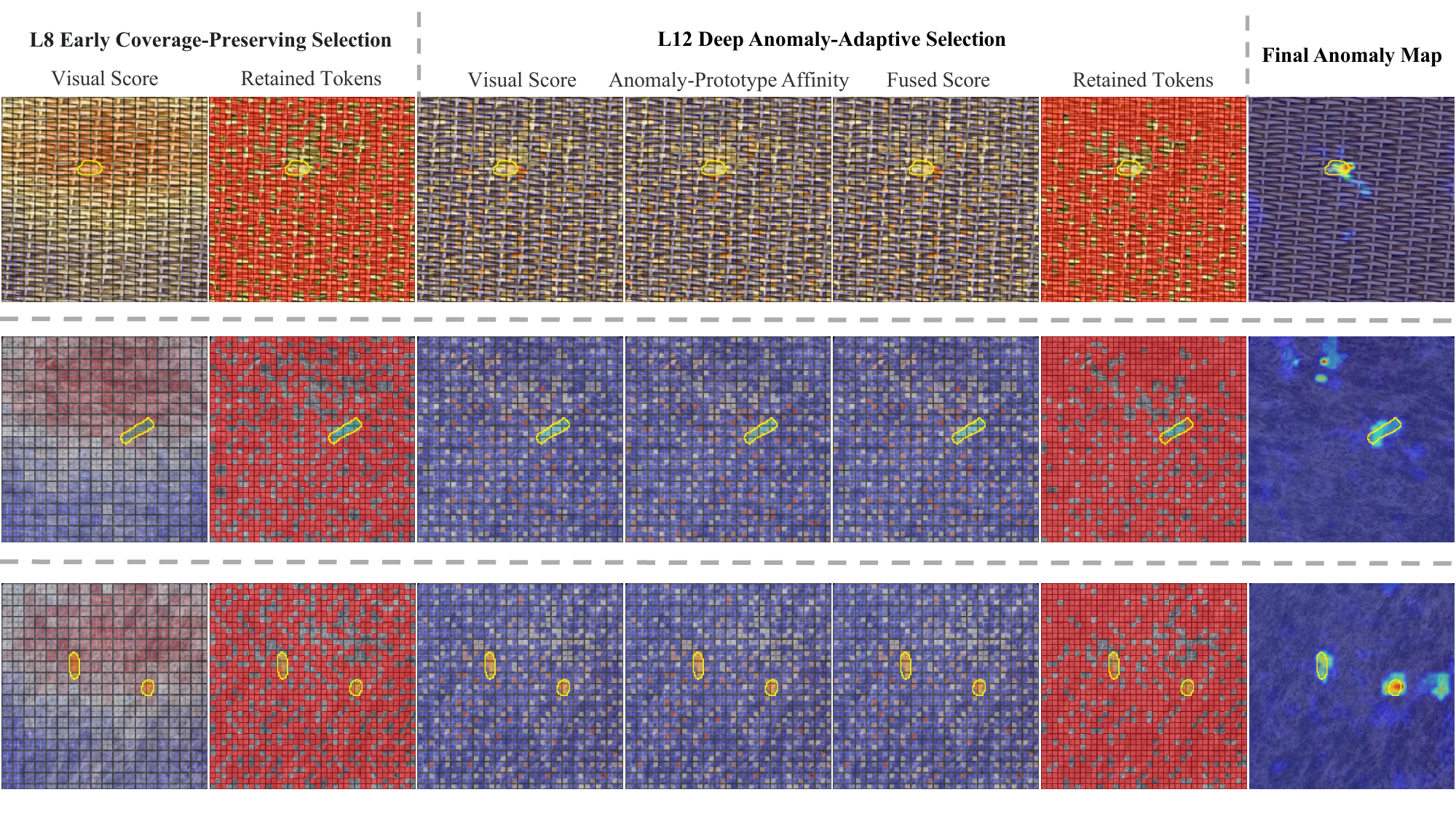}
\end{figure*}

\begin{figure*}[t]
    \centering
    \includegraphics[width=\textwidth]{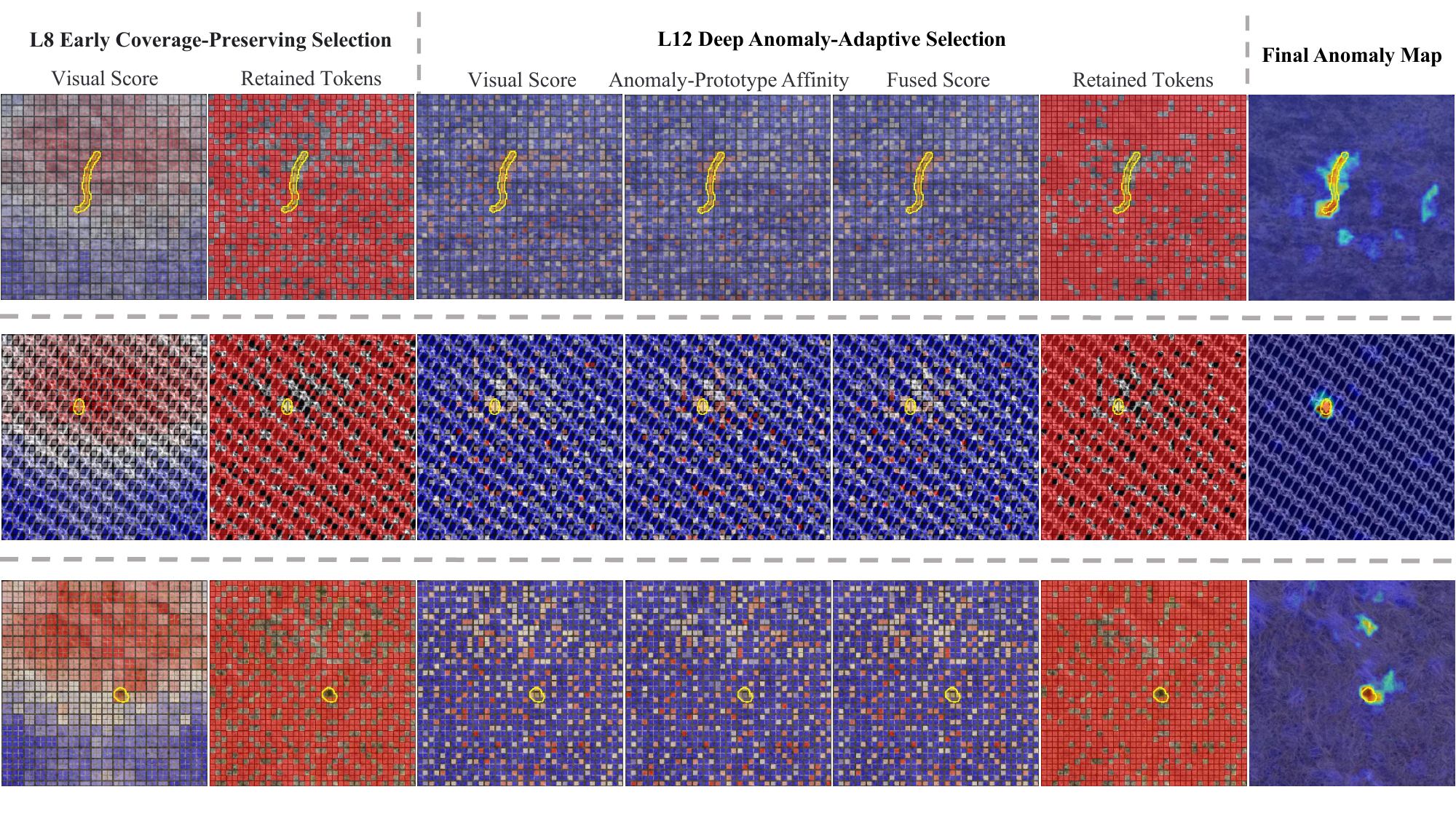}
\end{figure*}

\begin{figure*}[t]
    \centering
    \includegraphics[width=\textwidth]{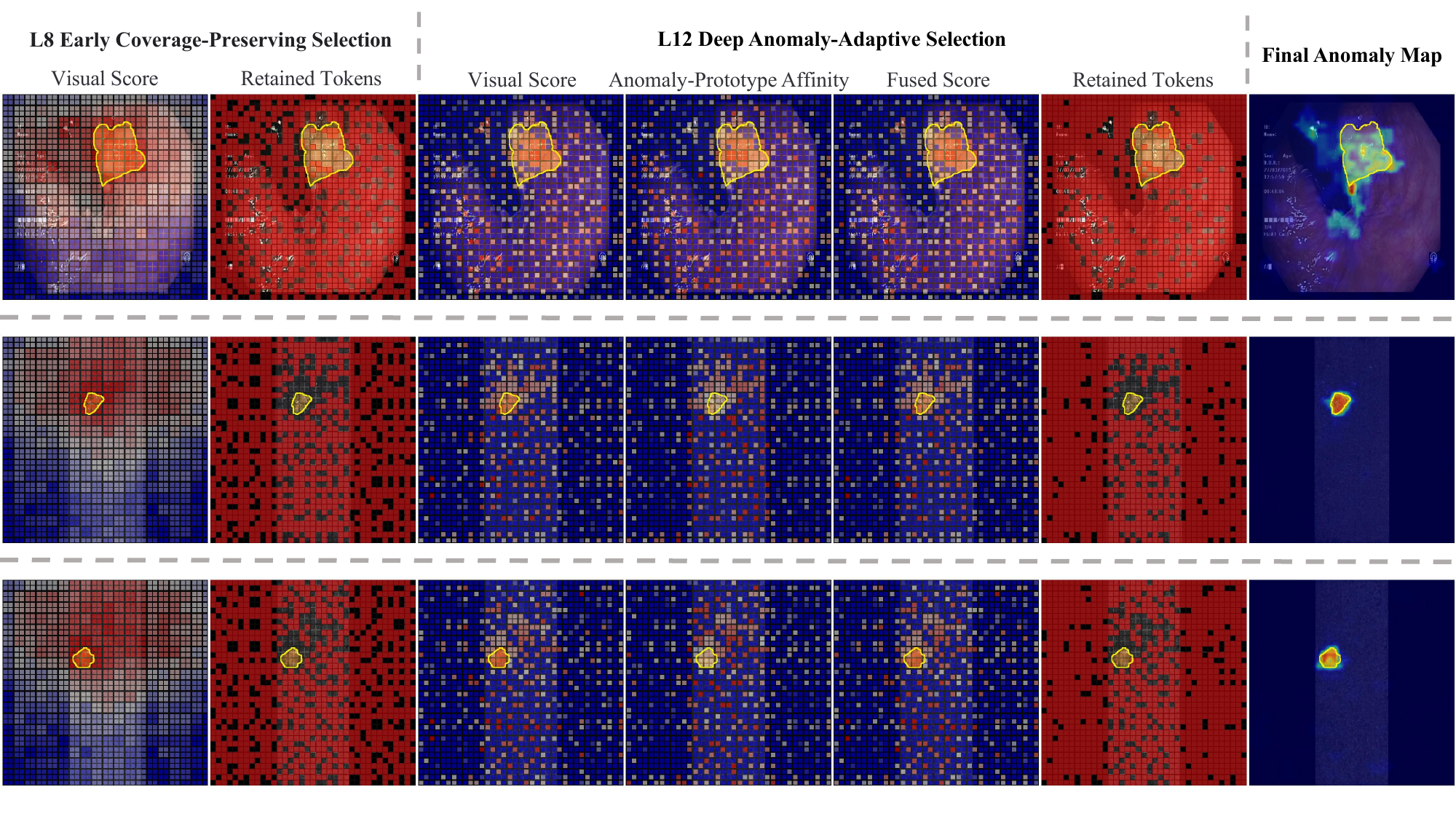}
\end{figure*}

\begin{figure*}[t]
    \centering
    \includegraphics[width=\textwidth]{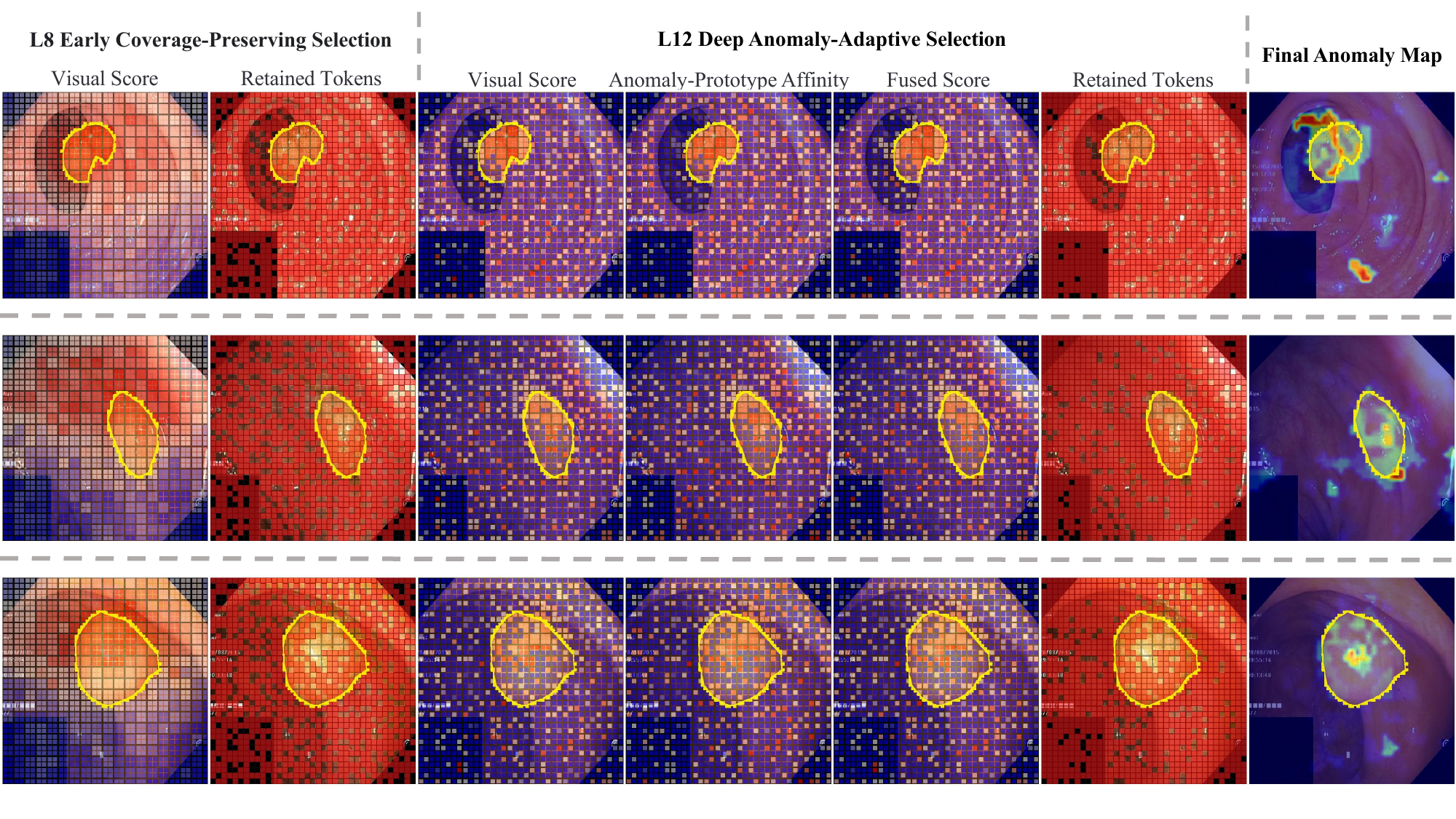}
\end{figure*}

\begin{figure*}[t]
    \centering
    \includegraphics[width=\textwidth]{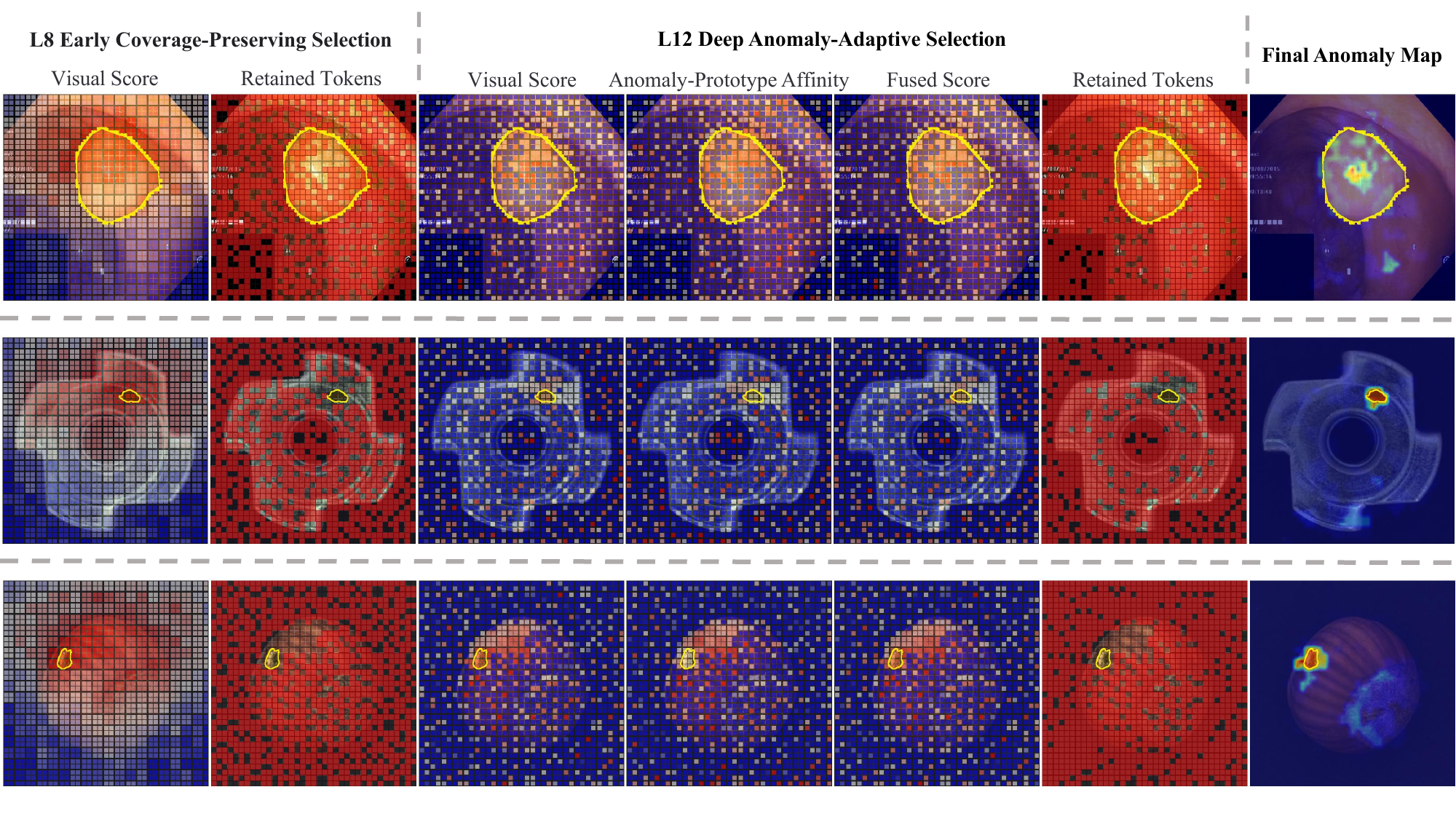}
\end{figure*}

\begin{figure*}[t]
    \centering
    \includegraphics[width=\textwidth]{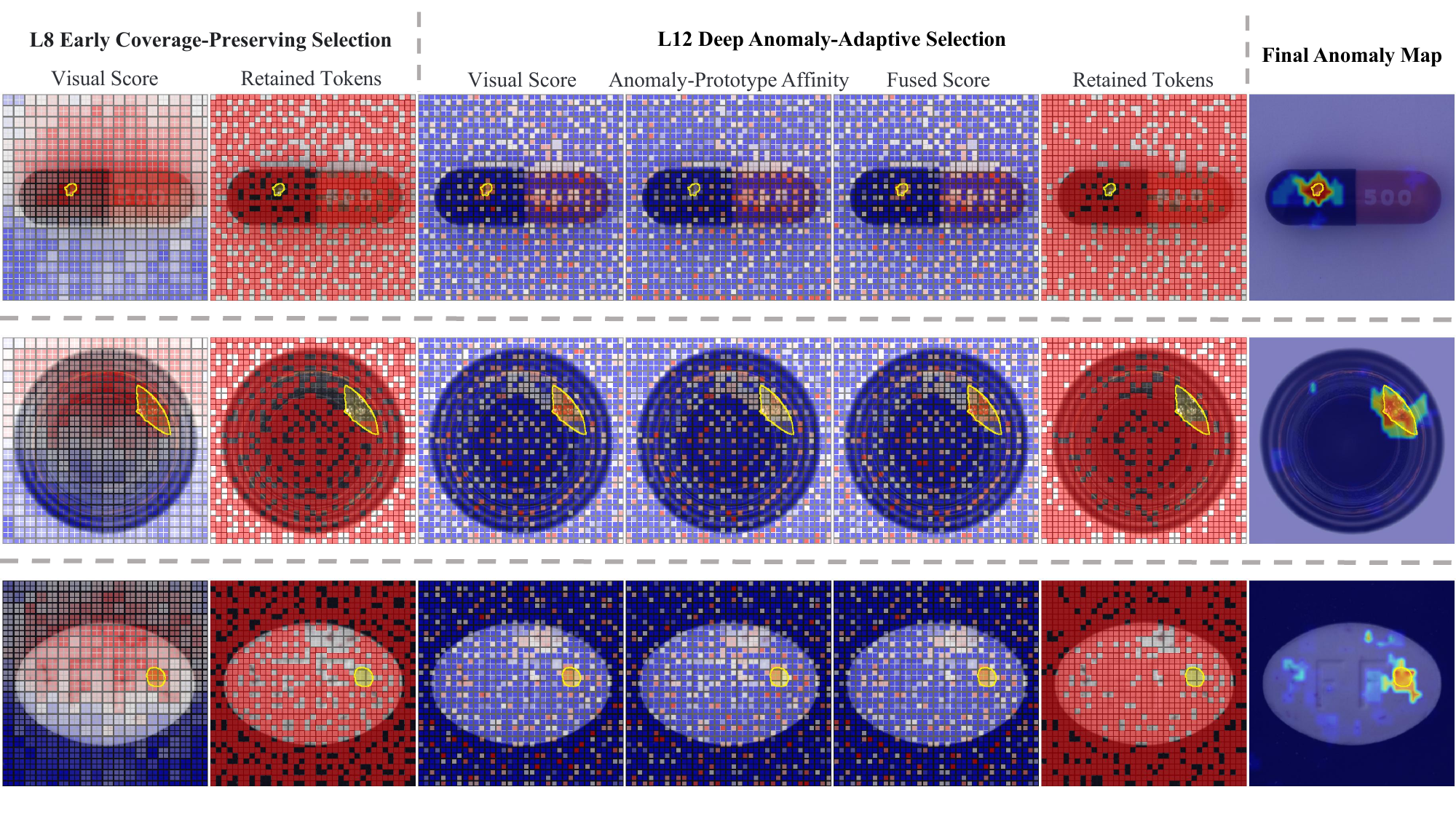}
\end{figure*}

\begin{figure*}[t]
    \centering
    \includegraphics[width=\textwidth]{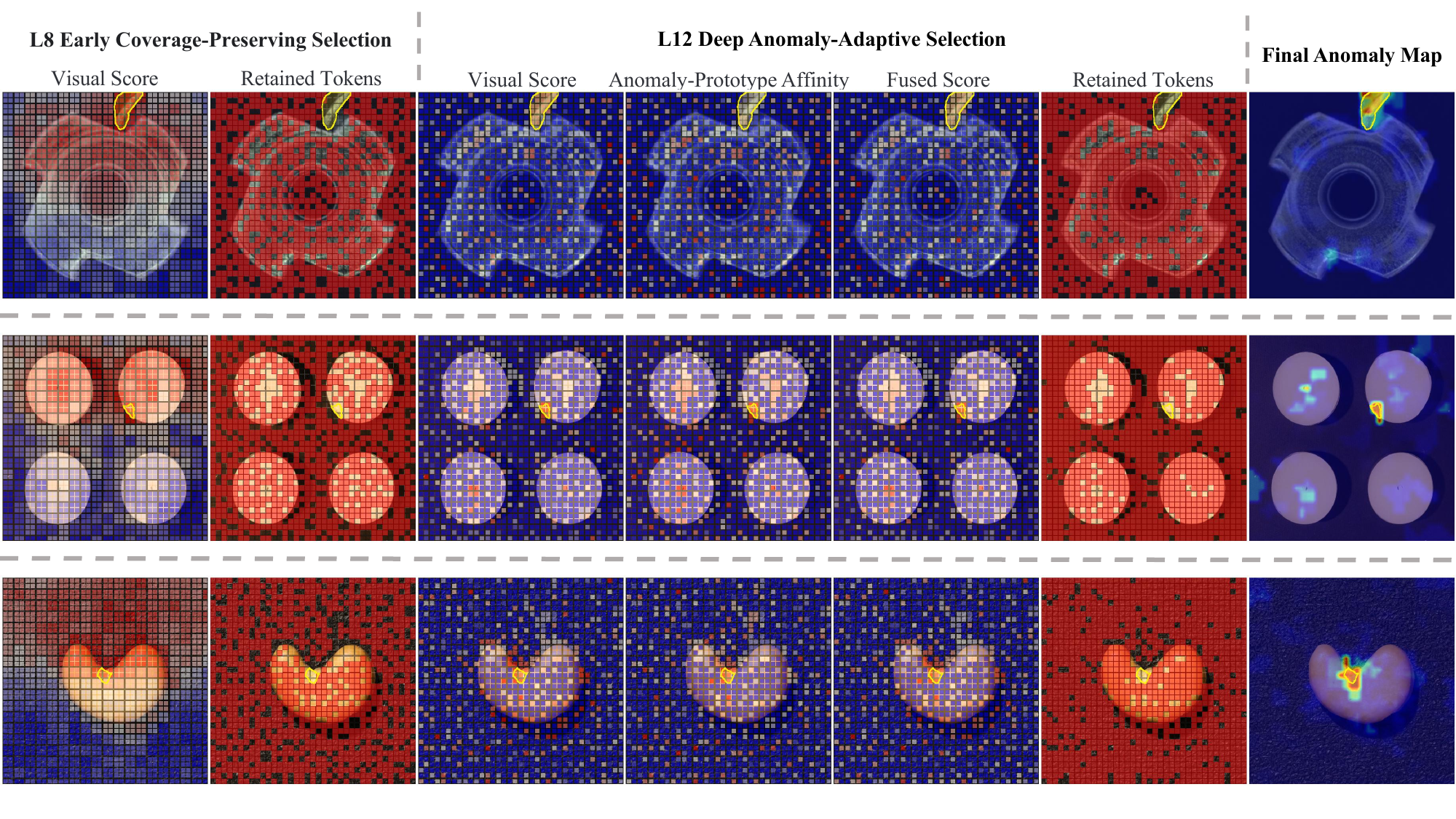}
\end{figure*}

\end{document}